\pdfoutput=1

\documentclass[11pt]{article}
\usepackage{placeins}
\usepackage[preprint]{acl}

\usepackage{enumitem}

\usepackage{times}
\usepackage{latexsym}
\usepackage{arabtex}
\usepackage{utf8}

\usepackage[most]{tcolorbox}
\usepackage{float}

\usepackage{booktabs}

\usepackage[T1]{fontenc}

\usepackage[utf8]{inputenc}

\usepackage{microtype}

\usepackage{inconsolata}

\usepackage{graphicx}
\usepackage[font=small,labelfont=bf]{caption}

\usepackage{xcolor}

\makeatletter
\renewenvironment{abstract}%
  {\centerline{\large\bf Abstract}%
    \begin{list}{}%
      {\setlength{\rightmargin}{0.6cm}%
       \setlength{\leftmargin}{0.6cm}}%
     \item[]\ignorespaces%
     \@setsize\normalsize{12pt}\xpt\@xpt
  }%
  {\unskip\end{list}}
\makeatother

\title{Beyond Understanding: Evaluating the Pragmatic and Cultural Gap in LLMs' Figurative Language Competence}

\author{
\textbf{Mena Attia\textsuperscript{1,2}}\quad 
\textbf{Aashiq Muhamed\textsuperscript{1}}\quad
\textbf{Mai Alkhamissi\textsuperscript{1}}\quad\\
\textbf{Thamar Solorio\textsuperscript{2}}\quad
\textbf{Mona Diab\textsuperscript{1}}\\
\\
\textsuperscript{1}Carnegie Mellon University \quad
\textsuperscript{2}MBZUAI\\
\texttt{mena.attia@mbzuai.ac.ae, mdiab@andrew.cmu.edu}
}

\begin{document}
\setcode{utf8}
\maketitle
\begin{abstract}
   We present a comprehensive evaluation of the ability of large language models (LLMs) to process culturally grounded language, specifically to understand and pragmatically use figurative expressions that encode local knowledge and cultural nuance. Using figurative language as a proxy for cultural competence and local knowledge, we design evaluation tasks for contextual understanding, pragmatic use, and interpretation of connotations in Arabic and English. We evaluate 22 open- and closed-source LLMs on Egyptian Arabic idioms, multidialectal Arabic proverbs, and English proverbs. Our results show a consistent hierarchy: the average accuracy for Arabic proverbs is 4.29\% lower than for English proverbs, and the performance for Egyptian idioms is 10.28\% lower than for Arabic proverbs. For the pragmatic use task, the accuracy decreases by 14.07\% relative to understanding, although providing contextual idiomatic sentences improves the accuracy by 10.66\%. Models also struggle with connotative meaning, reaching at most 85.58\% agreement with human annotators on idioms with 100\% inter-annotator agreement. These findings demonstrate that figurative language serves as an effective diagnostic for cultural reasoning: while LLMs can often interpret figurative meaning, they face challenges in using it appropriately. To support future research, we release \textit{Kinayat},\footnote{\url{https://huggingface.co/datasets/menaattia/Kinayat}} the first dataset of Egyptian Arabic idioms designed for both figurative understanding and pragmatic use evaluation. 
\end{abstract}

\section{Introduction}

\begin{table}[t]
\centering
\resizebox{0.95\columnwidth}{!}{%
\small
\setlength{\tabcolsep}{4pt}
\renewcommand{\arraystretch}{0.8}
\begin{tabular}{@{}p{1.8cm}p{5.3cm}@{}}
\toprule
\textbf{Proverb} &
\begin{arabtext}
\<ما تروحش تبيع المية في حارة السقايين>
\end{arabtext} \\
\midrule
\textbf{Literal Translation} &
Don't sell water in the village of water-sellers \\
\addlinespace[4pt]
\textbf{Figurative Meaning} &
Don't try to outsmart the experts. \\
\addlinespace[4pt]
\textbf{Historical Context} &
Historically, water-sellers in Egypt lived together in villages and brought water from the Nile to sell across neighborhoods. New sellers would mistakenly try to sell water in the water-sellers' village, failing to realize the abundance of water there made it an unsuitable market. \\
\addlinespace[4pt]
\textbf{Cultural Significance} &
Featured in the popular song ``Haret El-Sa'ayeen'' (1966) by Hussein Al-Sayed, sung by Sherifa Fadel, later by Mohammed Mounir with a performance on Arab Idol (2013) surpassing 132 million views on YouTube as of August 2025 \cite{mbc2013haret}. \\
\bottomrule
\end{tabular}
}
\caption{Sample Egyptian proverb: water-sellers village.}
\label{tab:egyptian_proverb}
\vspace{-0.2in}
\end{table}

Large language models (LLMs) have demonstrated remarkable progress in multilingual text understanding and generation. However, their ability to process \textit{culturally grounded meaning}—how language encodes local knowledge, social values, and pragmatic nuance—remains poorly understood. Figurative language offers a natural lens for studying this capability. Idioms and proverbs are among the most prevalent forms of figurative expression, deeply rooted in collective experience and cultural identity. They rely on shared world knowledge and pragmatic reasoning that extend beyond literal semantics. For example, Table~\ref{tab:egyptian_proverb} presents an Egyptian proverb. Without contextual knowledge, the reference to the village of water-sellers offers little clue to its intended message or its appropriate use. For models to truly understand language, they must internalize this cultural substrate rather than rely solely on surface statistics or translation mappings.

Most prior studies have focused on figurative comprehension—whether a model can explain an idiom or proverb—but not on pragmatic use, which requires context sensitivity, affective inference, and social appropriateness. In this work, we address these gaps by introducing a comprehensive evaluation framework that uses figurative language as a proxy for cultural competence and knowledge. We design tasks that test models’ abilities to (1) interpret figurative expressions, (2) use them appropriately in context, and (3) infer their connotative and affective dimensions. Our evaluation covers 22 open-source and closed-source models spanning multiple architectures and parameter scales. Although we leverage existing resources such as Jawaher (Arabic) and MAPS (English) for proverb interpretation, our central contribution is the introduction of \textbf{Kinayat}, a new dataset of Egyptian Arabic idioms annotated for both figurative meaning and pragmatic use.
Our work makes the following key contributions:
\begin{enumerate}[nosep]
    \item We introduce a \textit{pragmatic use task} to assess the ability of LLMs to employ figurative language appropriately in context;
    \item We present a unified cross-lingual evaluation suite that examines figurative interpretation, contextual appropriateness, and connotative inference;
    \item We present \textit{Kinayat}, a novel dataset of Egyptian Arabic idioms to evaluate figurative language understanding and pragmatic use.
\end{enumerate}

Our results demonstrate that performance consistently degrades from English proverbs to Arabic proverbs and finally to colloquial idioms, highlighting a systemic weakness in handling culturally-specific figurative language. We find that performance in Arabic proverbs is on average 4.29\% lower than in English proverbs, indicating a gap in language performance. Within Arabic, models perform 10.28\% worse on idioms from our Kinayat dataset (Egyptian dialect) compared to Modern Standard Arabic (MSA) proverbs.\footnote{MSA is the standard Arabic used in formal settings, while Egyptian dialect is a spoken vernacular, a related variant of MSA, used in Egypt.} 
In our novel pragmatic use task, we uncover a significant `Pragmatics Gap' in LLMs. Our core finding shows that model accuracy drops by an average of 14 percentage points when tasked with applying an idiom in context compared to simply explaining its meaning. That is, the average accuracy is 14.07\% lower than in the corresponding understanding task, with a maximum accuracy of 85.33\%, suggesting that the use of idioms appropriately in context remains difficult for current models. However, providing the idiom sentence as an additional context in the multiple-choice understanding prompt improves the average precision by 10.66\%. Finally, models exhibit notable limitations in grasping connotative meaning, achieving at most 85.58\% agreement with humans on samples with 100\% inter-annotator agreement.

\looseness=-1

\section{Related Work}
\begin{table*}[t]
    \centering
    \resizebox{\textwidth}{!}{%
    \begin{tabular}{l l l}
    \toprule
        \textbf{Dataset} & \textbf{Languages} & \textbf{Figurative Type} \\
    \midrule
        MAPS \cite{liu-etal-2024-multilingual}  & English, Chinese, German, Russian, Bengali, Indonesian & Proverbs \\
        MABL \cite{kabra-etal-2023-multi} & Hindi, Indonesian, Javanese, Kannada, Sundanese, Swahili, Yoruba & Metaphors \\
        Fig-QA \cite{liu-etal-2022-testing} & English & Metaphors \\
        FLUTE \cite{chakrabarty-etal-2022-flute} & English & Sarcasm, Simile, Metaphor, and Idioms \\
        PUB \cite{sravanthi-etal-2024-pub} & English & Implied answers, Presuppositions, Metonymy \\
        Jawaher \cite{magdy-etal-2025-jawaher} & Arabic (20 dialects) & Proverbs \\
        Fann or Flop \cite{alghallabi2025fannflopmultigenremultiera} & Arabic & Poetry \\
    \midrule
    Kinayat (ours) & Arabic & Idioms \\
    \bottomrule
    \end{tabular}%
    }
    \caption{Comparison of existing figurative language datasets. 
    }
    \label{tab:datasets}
    \vspace{-0.2in}
\end{table*}

Recent years have seen a surge of interest in evaluating and improving the ability of LLMs to interpret and generate figurative language. Several benchmarks have been proposed to assess various dimensions of figurative understanding, including metaphor, idiom, proverb, and poetry interpretation. Table~\ref{tab:datasets} provides a comparative summary of these datasets.

Several English-focused datasets have been developed to evaluate literal versus figurative reasoning. The Fig-QA dataset \cite{liu-etal-2022-testing} frames figurative language understanding as a multiple-choice question answering task. \citet{rocket_science} evaluate the ability of LLM to interpret idioms and similes plausibly by continuing narratives. Other efforts include FLUTE \cite{chakrabarty-etal-2022-flute} and follow-up work that explores metaphor interpretation using chain-of-thought prompting and psychologically informed reasoning \cite{he-etal-2022-pre, prystawski2023psychologicallyinformedchainofthoughtpromptsmetaphor, jang-etal-2023-figurative}, showing that while LLMs can often identify meaning at the surface level, they frequently fail to capture implicit moral or social connotations. The PUB benchmark \cite{sravanthi-etal-2024-pub} specifically evaluates pragmatic competence, testing the model’s ability to distinguish between literal and contextually appropriate figurative meanings in conversation.

In the multilingual setting, the MAPS dataset \cite{liu-etal-2024-multilingual} evaluates figurative understanding in six languages of proverbs, and the ProverbEval dataset \cite{azime-etal-2025-proverbeval} evaluates LLMs of cultural proverbs for four Ethiopian languages and English, while MABL \cite{kabra-etal-2023-multi} provides a multilingual benchmark for metaphor and simile comprehension in underrepresented languages.

Figurative language is an integral component of culture, yet Arabic language understanding and cultural benchmarks often omit it entirely. For example, widely used Arabic cultural benchmarks such as AraDiCE \cite{mousi2024aradicebenchmarksdialectalcultural}, CAMEL-Bench \cite{ghaboura2024camelbenchcomprehensivearabiclmm}, and CIDAR \cite{alyafeai-etal-2024-cidar} do not include figurative language as an explicit category. Others incorporate it only in limited capacity, such as the ArabCulture dataset \cite{sadallah2025commonsensereasoningarabculture}, which covers 13 countries and 12 topics with idioms as one topic, but includes only five samples per country, and PALM \cite{alwajih-etal-2025-palm}, which covers 20 diverse topics from 22 Arab countries with proverbs as one of the topics. 

There are two recent benchmarks that focus specifically on Arabic figurative language: the Jawaher dataset \cite{magdy-etal-2025-jawaher}, the first large-scale collection of Arabic proverbs in 20 dialects, and Fann or Flop \cite{alghallabi2025fannflopmultigenremultiera}, a benchmark for the interpretation of Arabic poetry that spans multiple genres and eras. However, the lack of datasets that focus on figurative language highlights the need for more comprehensive evaluation resources that integrate figurative language as an explicit component of cultural understanding.

Recent work has also examined the affective and social dimensions of figurative expressions. \citet{martínez2024usinglargelanguagemodels} use LLMs to estimate the valence, arousal, and concreteness of multi-word expressions, offering a route to investigate deeper semantic features. In the Arabic context, \citet{alsiyat-piao-2020-metaphorical} demonstrate that metaphorical constructions significantly affect sentiment analysis performance, underscoring the need for a connotative understanding in Arabic Natural Language Processing (NLP).

\section{Methodology}

To provide a comprehensive evaluation of the figurative language, we design a framework of tasks that probe LLM capabilities beyond simple comprehension in Arabic idioms, Arabic proverbs, and English proverbs. Although prior work has benchmarked proverb understanding, our framework introduces a suite of tasks, including negation, contextual reasoning, pragmatic use, and connotation labeling, that have not been systematically applied to the Arabic figurative language before.

\paragraph{Pragmatic Use:}
We introduce a new task to evaluate the \textit{pragmatic use} of Arabic idioms in LLMs. Using a subset of 150 idioms from the Kinayat dataset, we use a model-in-the-loop approach \cite{chakrabarty-etal-2022-flute, liu-etal-2024-multilingual}, prompting GPT-4.1 \cite{openai_gpt41_2025} with each idiom and its explanation to generate a sentence in Egyptian Arabic, using the prompt shown in Figure~\ref{fig:idiom-prompt}. 

A native Egyptian Arabic speaker then reviews the generated sentences and either accepts them, makes minor changes, or comes up with new sentences to replace them. Of the 150 samples, 77 (51.3\%) are modified, either by minor dialectal adjustments, rephrasing, the addition of feminine examples, or complete rewriting of the sentence. The unmodified generated samples appear in first person, first person plural, second person, or third person masculine. Many examples are related to work, with the word \<شغل> "work" appearing 29 times and \<مدير> "manager" appearing 7 times.

\begin{figure}[ht]
    \centering
    \includegraphics[width=0.95\columnwidth]{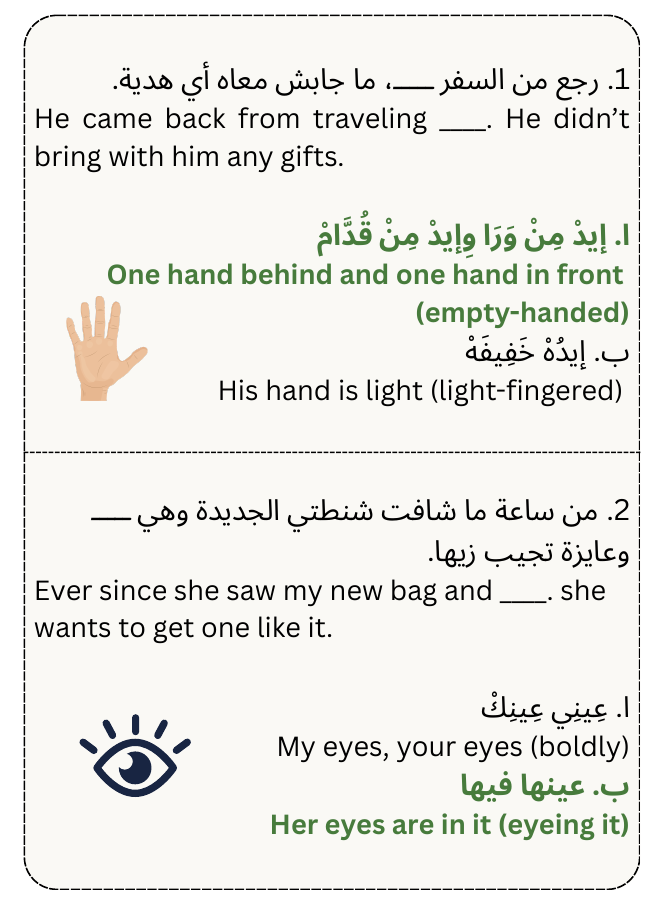}
    \caption{Sample questions from the Pragmatic Use task, along with their English translations.}
    \label{fig:prag-use-examples}
    \vspace{-0.2in}
\end{figure}

As part of the task setup, for each idiom, the annotator selects a plausible but incorrect idiom from the Kinayat set to serve as a distractor in a multiple-choice setting. We then evaluate LLMs, asking them to choose the correct idiom for a given sentence, using the template in Figure~\ref{fig:prag-use}. Representative examples of this task are shown in Figure~\ref{fig:prag-use-examples}.

We evaluate the performance of LLMs on the following tasks using zero-shot prompting, with a single evaluation run per experiment.
All prompts are included in the appendix \ref{app:prompts}. All prompts are in English to maintain methodological consistency in both English and Arabic datasets in our evaluation, allowing for direct comparison without introducing confounding factors from prompt language variation, and following previous work in multilingual LLM evaluation \cite{liu-etal-2024-multilingual, lin-etal-2022-shot, muennighoff-etal-2023-crosslingual}.

\paragraph{Multiple Choice Question (MCQ) Understanding:} 
To evaluate figurative comprehension, we construct multiple choice questions, inspired by the work of \citet{liu-etal-2024-multilingual, kabra-etal-2023-multi}, where each item presents two candidate explanations for a given idiom or proverb, one correct and one incorrect.

Because understanding is assessed through a multiple-choice format, the results are inherently influenced by the quality and difficulty of incorrect explanations. When an incorrect explanation is implausible or obviously wrong, models can often eliminate it without genuinely understanding the meaning of the idiom or proverb. 
To try to mitigate this, incorrect explanations are generated using GPT-4.1 with two distinct strategies. First, we use a general prompt template (Figure~\ref{fig:generate-prompt}) to produce a semantically plausible but incorrect explanation for each idiom or proverb. Second, we leverage semantic role labeling (SRL) with LLMs \cite{cheng2024potentiallimitationsllmscapturing} to generate more subtly incorrect explanations: we extract the semantic roles from the gold explanation and modify a single role to alter the meaning, using the SRL-based prompt in Figure~\ref{fig:generate-prompt-srl}.

Each model is prompted to choose the correct explanation in two separate evaluation settings: once with the general incorrect explanations, and once with the SRL-based alternatives.

\paragraph{Human Verification of Incorrect Distractors}
To evaluate the quality of the generated explanations, a native Arabic speaker manually assesses a random sample of 50 incorrect explanations for idioms from Kinayat and another 50 for proverbs from Jawaher. Each explanation is rated on a three-point scale from 0 to 2, where higher scores indicate explanations that are incorrect, but linguistically and culturally plausible, while lower scores correspond to explanations that are implausible or irrelevant. Detailed annotation guidelines are provided in the appendix~\ref{app:verify-incorrect}. The average verification scores are 1.72 for idioms and 1.68 for proverbs, confirming that all the generations sampled were indeed incorrect, although they vary in clarity and plausibility. This slight difference suggests that it is slightly easier to produce plausible but incorrect explanations for idioms than for proverbs, although this likely reflects differences in the gold explanations provided as input rather than in the idioms or proverbs themselves.

\paragraph{Contextual Understanding MCQ:}  
We extend the MCQ Understanding Task by providing idiomatic sentences for a subset of Arabic idioms from Kinayat as additional context. Models are then prompted to select the correct explanation given both the idiom and its usage in context, using the prompt shown in Figure~\ref{fig:contextual-mcq-prompt}.

\paragraph{Negation:}  
Building on the work of \citet{liu-etal-2024-multilingual}, we adapt the MCQ understanding task in Arabic datasets by requiring the models to select the \textit{incorrect} explanation rather than the correct one. This inversion introduces a negation component to the task, as models must reject the correct option, which prior work in natural language inference has shown to degrade model performance \cite{truong-etal-2023-language, she-etal-2023-scone}.
\paragraph{Explanation Generation:}  
Beyond evaluating models' ability to identify the correct explanation, we also assess their capability to \textit{generate} explanations for Arabic idioms and proverbs. Models are prompted to produce explanations using the prompt template shown in Figure~\ref{fig:generation-prompt}.

\paragraph{Completing the Proverb:}  
We evaluate the memorization of cultural knowledge of the models by masking the final word of each English and Arabic proverb, following \citet{liu-etal-2024-multilingual}, and prompting the model to complete it using the prompt template shown in Figure~\ref{fig:complete-proverb-prompt}.

\paragraph{Connotation Understanding:} Connotations of each Arabic proverb and idiom are labeled by three native Egyptian Arabic speakers as \textit{positive}, \textit{negative}, or \textit{neutral}. Human annotators are provided with the same prompt template as the models to standardize task instructions. We then evaluate models only on samples with 100\% agreement to minimize the impact of connotation subjectivity on model performance. We use two variants of the task employing the prompt template in Figure \ref{fig:connotation-prompt}: (1) predicting the connotation given the proverb or idiom, and (2) predicting the connotation given its explanation. Model predictions are compared with human annotations to calculate accuracy.

\section{Experimental Setup}
\subsection{Datasets}
\paragraph{Jawaher} \cite{magdy-etal-2025-jawaher}: We use 198 test samples of multidialectal proverbs across 20 varieties and their Arabic explanations. 
\vspace{-5pt}

\paragraph{MulticulturAl Proverbs and Sayings (MAPS)} \cite{liu-etal-2024-multilingual}: A multilingual dataset of proverbs in 6 languages. We only use the English test set which consists of 394 proverbs. 
\vspace{-5pt}

\paragraph{Kinayat: A Dataset of Egyptian Arabic Colloquial Idioms}
We introduce the Kinayat dataset, which consists of 325 Egyptian idioms along with their MSA explanations. We extracted them from the book \textbf{Al-Kinayat Al-'Amiyya} \cite{Taymour1949}.\footnote{The book is publicly available by the publisher.} %
Some examples are shown in Table \ref{tab:egyptian-idioms} in Appendix~\ref{app:idioms}.

\vspace{-5pt}
\paragraph{Data Preprocessing:} During preprocessing, we removed inappropriate or incomplete idioms, resulting in the exclusion of 10 entries from the dataset. Citations appearing in parentheses within the text were deleted to ensure clarity and consistency. We also removed phonetic explanations, as well as similar idioms and sayings that were mentioned with the explanations. Examples and excerpts of poetry were excluded to maintain a focus on the core figurative expressions. Finally, for idioms whose explanations merely referred to an equivalent idiom without further clarification, we supplemented the data by adding in full explanatory text.

\subsection{Models}

Our experiments consider a broad selection of LLMs, encompassing both open-source and closed-source options, as well as Arabic and non-Arabic models of varying size.
We evaluate a total of 22 models.

The open-source multilingual models evaluated include LLaMA 3.1 (8B, 70B-Instruct) \cite{grattafiori2024llama3herdmodels}, Gemma 2 (9B, 27B-IT) \cite{gemmateam2024gemma2improvingopen}, Qwen-2.5 (7B, 14B, 32B-Instruct) \cite{qwen2025qwen25technicalreport}, Aya-Expanse (8B, 32B) \cite{dang2024}, and Mistral-7B-Instruct-v0.3 \cite{jiang2023mistral7b}. We further evaluate a diverse set of open-source Arabic models, such as Jais-family-6p7b-chat, Jais-family-13b-chat \cite{sengupta2023}, ALLaM-7B-Instruct-preview \cite{bari2025allam}, SILMA-9B-Instruct-v1.0 \cite{silma-9b-2024}, Fanar-1-9B-Instruct \cite{fanarteam2025fanararabiccentricmultimodalgenerative}, and AceGPT-v2-8B-Chat \cite{acegpt}.

The closed-source models included in our evaluation are GPT-4o and 4o-mini \cite{openai2024gpt4technicalreport}, Gemini 1.5 Flash \cite{geminiteam2024gemini15unlockingmultimodal}, Gemini 2.5 Flash Lite Preview \cite{comanici2025gemini25pushingfrontier}, as well as Claude 3.5 Sonnet \cite{anthropic-claude35sonnet-announcement-2024} and Sonnet 4 \cite{anthropic-claude4-announcement-2025}.

\subsection{Evaluation}
We evaluate LLM performance across the different tasks using the \texttt{lm-eval} framework \cite{eval-harness}, which computes accuracy using log-likelihood for open-source models and model-generated outputs for closed-source APIs. For open-source models, \texttt{lm-eval} was configured with \texttt{vLLM} \cite{vllm} and tensor parallelism to distribute compute across 4 GPUs. Our code and evaluation scripts are publicly available on GitHub.\footnote{\url{https://github.com/menaattia/llm-figurative-understanding}}

For multiple-choice tasks, we report the mean \textit{accuracy $\pm$ standard error} as the primary evaluation metrics. For generation-based tasks, we employ two complementary evaluation methods:

\begin{enumerate}[nosep]
    \item \textbf{BERTScore} \cite{bertscore}: Measures semantic similarity between model-generated explanations and gold references using contextual embeddings from \texttt{bert-base-multilingual-cased}, with the language set to Arabic (\texttt{lang="ar"}). We report the F1 score as the main evaluation metric.
    \item \textbf{LLM-as-a-Judge} 
 \cite{zheng2023judgingllmasajudgemtbenchchatbot}: Uses a LLM to assess how well the generated explanation aligns with the intended meaning of the gold explanation. We use GPT-4.1 as the default system to score all models except for the evaluation of the GPT models, which are instead judged using Claude-3.5-Sonnet to avoid self-evaluation bias \cite{wataoka2024selfpreference}. The scoring prompt is shown in Figure~\ref{fig:judge-proverb-prompt}. 
 \item \textbf{Human Evaluation:} We also conduct human evaluation on a subset of the generations to validate the reliability of the LLM-as-a-Judge scores.
\end{enumerate}

\begin{figure*}[htbp]
    \centering
    \includegraphics[width=0.9\textwidth]{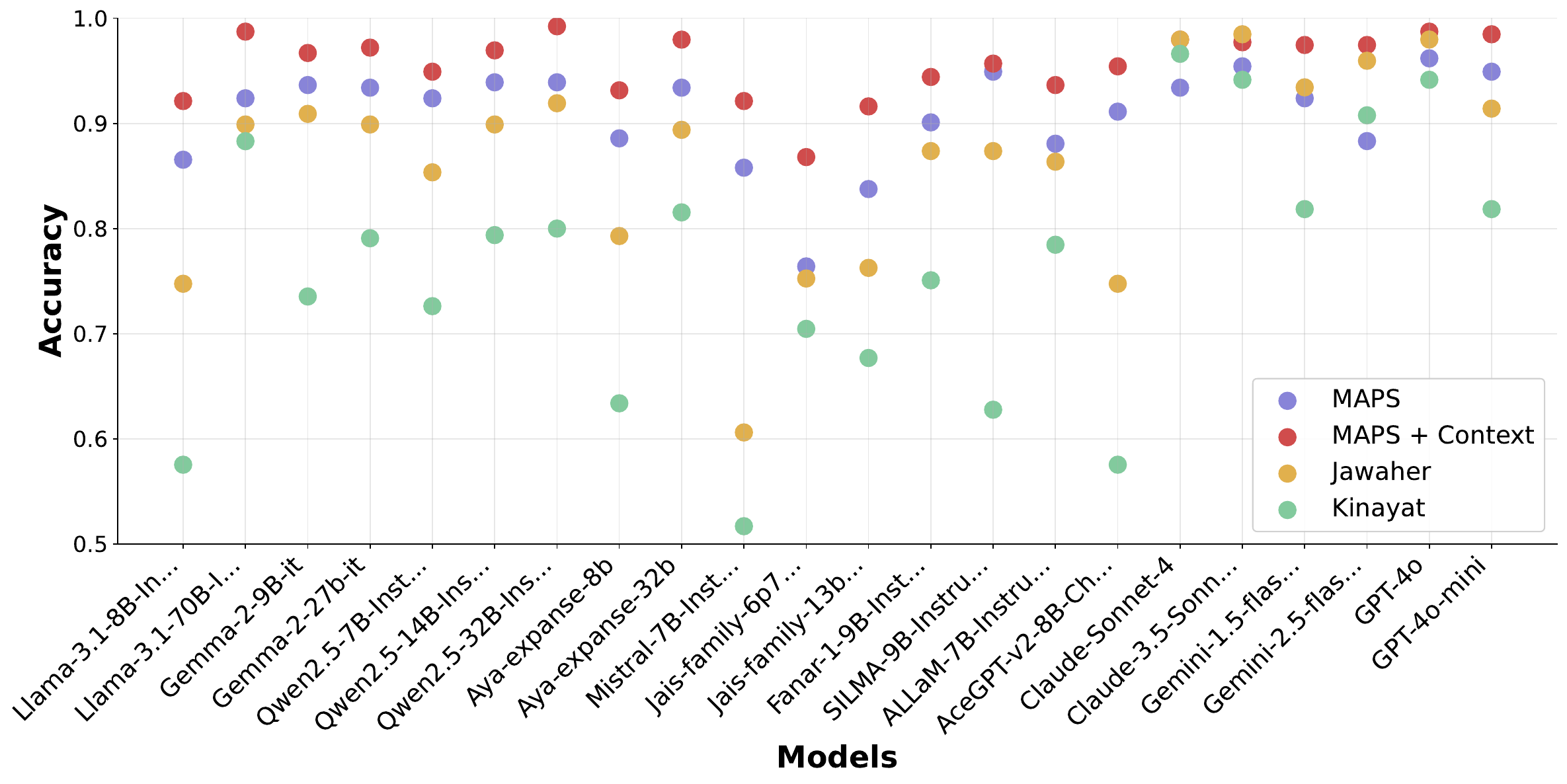}
    \vspace{-0.1in}
    \caption{Accuracy on MCQ Understanding task across different test sets.}
    \label{fig:mcq-results-graph}
    \vspace{-0.2in}
\end{figure*}

\section{Results}
\subsection{Knowledge and Understanding}

\paragraph{Distinguishing Between Correct and Incorrect explanations}
In Figure~\ref{fig:mcq-results-graph}, we observe a clear difficulty hierarchy in understanding MCQ, with the highest model performance on MAPS with context (95.66\%), followed by MAPS (90.86\%), Jawaher (86.57\%) and lowest on Kinayat (76.29\%). Arabic proverbs consistently yield lower performance than English proverbs, both with context (95.66\% vs.\ 86.57\%) and without context (90.86\% vs.\ 86.57\%), indicating a persistent gap in cross-lingual and cultural understanding. One contributing factor may be differences in context lengths; English proverbs exhibit a higher mean sentence-length distribution than Arabic proverbs (6.52 vs. 5.21), providing richer lexical and syntactic context that can help models. Furthermore, Arabic idioms are more challenging than Arabic proverbs (average accuracy of 76.29\% vs. 86.57\%), possibly because proverbs tend to be almost frozen making them easier to memorize as they could have been seen in the training data as opposed to idioms. Additionally, idioms occur in shorter and less informative contexts, as shown by their length distributions in Figure~\ref{fig:length-dist} (Appendix~\ref{app:analysis}). Detailed results for the MCQ task and subsequent tasks are presented in Appendix~\ref{app:more-results}.

Table~\ref{tab:negation-kinayat-jawaher} presents the results for the negation variant of the MCQ task. When models were asked to identify the \textit{incorrect} explanation rather than the correct one. This inversion led to a notable performance drop: for idioms, the average accuracy decreased from 76.29\% to 70.97\%, and for proverbs from 86.57\% to 82.71\%. Interestingly, GPT-4o outperformed the Claude models in this more challenging negation task, despite the Claude models achieving higher accuracy in the original MCQ understanding task. 

\vspace{-5pt}

\paragraph{Knowledge and Memorization of Proverbs} Results for the proverb completion task are presented in Table~\ref{tab:completion-model-results}. Claude 3.5 achieved the highest accuracy for both English and Arabic proverbs, with scores of 93.91\% and 36.36\%, respectively. On average, the performance of the model was substantially higher for English proverbs (75.43\%) compared to Arabic proverbs (10.65\%). The low accuracy of Arabic completions suggests limited memorization, which may reflect a lower representation of Arabic proverbs in the models’ training data. This gap in completion accuracy mirrors the performance differences observed in the proverb understanding tasks, suggesting that greater exposure to English proverbs during training may provide models with stronger lexical and phrasal cues that facilitate both memorization and understanding. Despite low memorization, models perform well in the understanding task, suggesting that they can still reason effectively without relying on memorization~\citep{liu-etal-2024-multilingual}.

\paragraph{Model Ability to Generate Explanations} Claude 3.5 Sonnet achieved the highest scores in both evaluation metrics, BERTScore-F1 and LLM-as-a-Judge, for both proverbs and idioms. For proverbs, it scored 0.70 on BERTScore-F1 and 3.89 on the LLM-as-a-Judge scale (1–5). For idioms, it scored 0.68 and 2.93, respectively. Although higher BERTScore-F1 values did not always correspond to higher LLM-as-a-Judge scores, the highest BERTScore-F1 outputs were consistently aligned with the highest human-aligned ratings. Overall, idioms proved to be more challenging for models to generate explanations for, with lower average scores across both metrics. Specifically, the average BERTScore-F1 and LLM-as-a-Judge scores were 0.65 and 2.19 for idioms, compared to 0.68 and 3.06 for proverbs. 

\paragraph{LLM-as-a-Judge Verification} To validate the reliability of the LLM-as-a-Judge scoring procedure, we include human evaluation on a representative subset of 220 generated explanations, corresponding to 10 explanations per model across the 22 evaluated models. This subset was sampled to cover a range of scores and idioms, enabling assessment of whether the automated judgments align with human judgments of correctness and plausibility. The average LLM-as-a-Judge score on this subset was 2.15, compared to an average human score of 1.94, indicating slight overestimation by the automated evaluator (0.21 points). In terms of agreement, human and LLM-as-a-Judge scores matched exactly in 62.27\% of cases; in 29.09\% of cases, the human score was lower than the LLM score, while in 8.64\% of cases, the human score was higher. Nevertheless, both evaluations consistently reflect low explanation quality overall. These results demonstrate that the LLM-as-a-Judge scores are broadly consistent with human evaluation, supporting their use at scale in our analysis.

\vspace{-5pt}

\subsection{Pragmatic Use}\label{sec:prag}
As shown in Table~\ref{tab:pragmatic-use}, the highest accuracy of 85.33\% on the pragmatic use task was achieved by Claude 3.5 Sonnet. The average accuracy on pragmatic use across all models was 64.45\%, which is 14.07\% lower than the average accuracy on the understanding task for the same set of 150 idioms. When the sentence containing the idiom was added as context to the MCQ understanding prompt, the average accuracy increased by 10.66\% to 89.18\%. These results demonstrate a performance gap between understanding and pragmatic use even for frontier models, indicating that using idioms pragmatically in context is more challenging than choosing the correct explanation from multiple choice options.

\begin{figure}[htbp]
    \centering
    \includegraphics[width=\columnwidth]{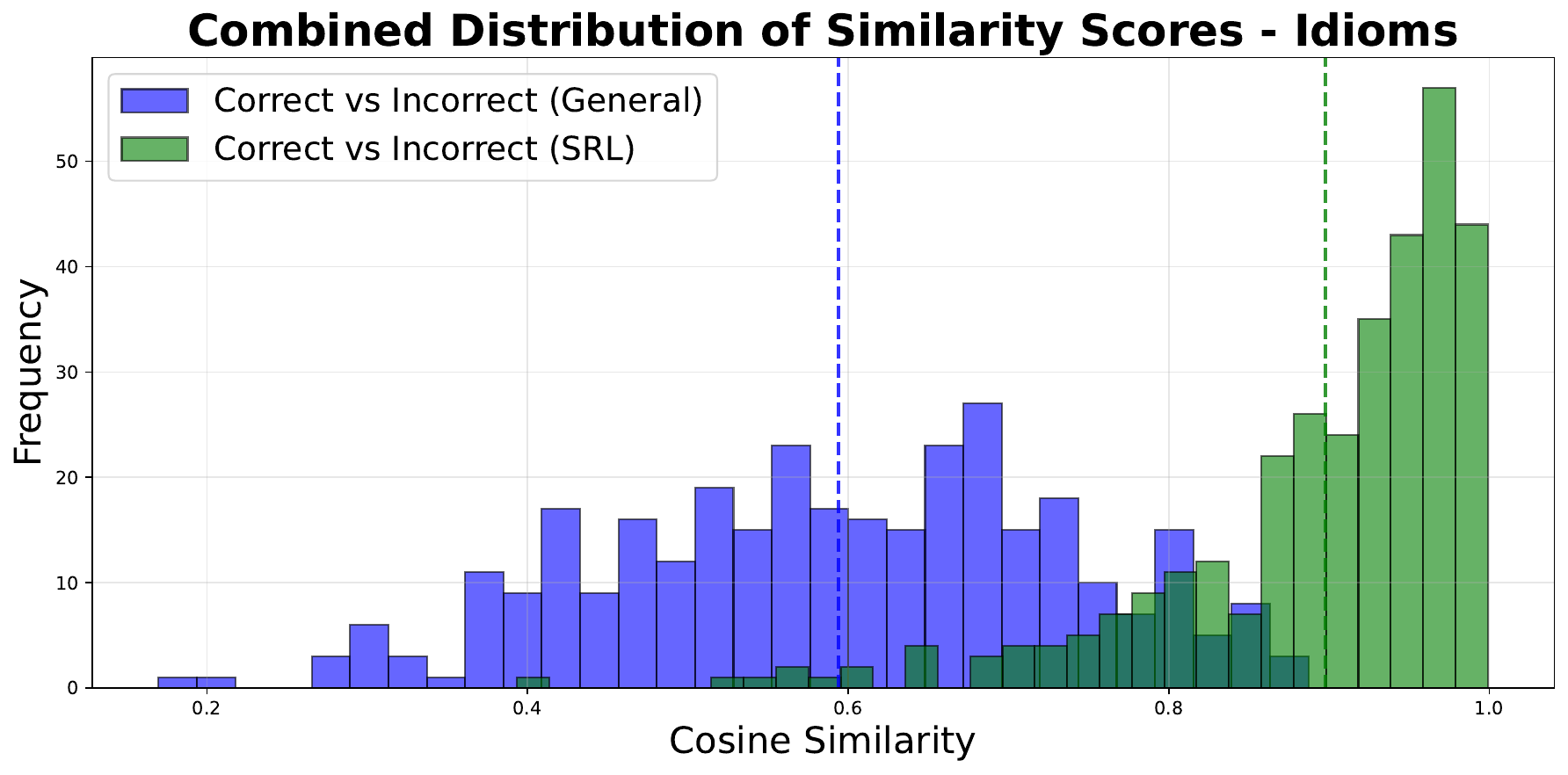}
    \vspace{-0.3in}
    \caption{Cosine similarity ($\uparrow$) between the correct and incorrect explanation choices for idioms (Kinayat).}
    \label{fig:cos-sim-idioms}
    \vspace{-0.2in}
\end{figure}

\subsection{Connotations}
Annotating connotation in Arabic figurative expressions reveals inherent subjectivities in cultural expressions. Assigning clear sentiment values is often reductive, as interpretation varies widely across individuals and groups \cite{sap-etal-2022-annotators, truthisalie}.
Out of the 198 multidialectal proverbs, 105 (53.03\%) achieved full inter-annotator agreement on their connotations, while 104 out of 150 Egyptian idioms (69.3\%) reached full agreement, underscoring the inherent subjectivity of connotation judgments. To reduce the confounding effect of subjectivity of connotations, models were evaluated only on the samples where all three annotators agreed, ensuring a focus on stronger connotations. Claude-3.5-Sonnet achieved the highest accuracy in labeling the connotations of idioms (85.58\% accuracy) and proverbs (74.04\% accuracy), while Claude-Sonnet-4 achieved the highest accuracy in labeling the connotations of explanations, with accuracy scores of 89.42\% for idioms and 86.54\% for proverbs. On average, accuracy of explanations was higher by 22.11\% for idioms and 18.46\% for proverbs, which is expected given that the purpose of the explanations is to clarify figurative expressions. These results suggest that models struggle with connotations, which could be related to how they perform less on the pragmatic use task. All connotation results are presented in Table~\ref{tab:sentiment-match} (Appendix~\ref{app:more-results}).

\begin{table*}[t]
\centering
\resizebox{0.8\textwidth}{!}{%
\small
\setlength{\tabcolsep}{7pt}
\renewcommand{\arraystretch}{0.6}
\begin{tabular}{lccc}
\toprule
\textbf{Model} &
\textbf{Pragmatic Use} &
\textbf{Understanding} &
\textbf{Contextual Understanding} \\
\midrule
Llama-3.1-8B-Instruct     & 0.5400$\,^{\pm 0.0408}$ & 0.6000$\,^{\pm 0.0401}$ & 0.7867$\,^{\pm 0.0336}$ \\
Llama-3.1-70B-Instruct    & 0.6333$\,^{\pm 0.0395}$ & 0.8867$\,^{\pm 0.0260}$ & 0.9667$\,^{\pm 0.0147}$ \\
Gemma-2-9B-it             & 0.6000$\,^{\pm 0.0401}$ & 0.7867$\,^{\pm 0.0336}$ & 0.8933$\,^{\pm 0.0253}$ \\
Gemma-2-27b-it            & 0.6933$\,^{\pm 0.0378}$ & 0.8133$\,^{\pm 0.0319}$ & 0.9533$\,^{\pm 0.0173}$ \\
Qwen2.5-7B-Instruct       & 0.5267$\,^{\pm 0.0409}$ & 0.7733$\,^{\pm 0.0343}$ & 0.8667$\,^{\pm 0.0278}$ \\
Qwen2.5-14B-Instruct      & 0.6467$\,^{\pm 0.0392}$ & 0.8400$\,^{\pm 0.0300}$ & 0.9667$\,^{\pm 0.0147}$ \\
Qwen2.5-32B-Instruct      & 0.7133$\,^{\pm 0.0370}$ & 0.8200$\,^{\pm 0.0315}$ & 0.9267$\,^{\pm 0.0214}$ \\
Aya-expanse-8b            & 0.5600$\,^{\pm 0.0407}$ & 0.6533$\,^{\pm 0.0390}$ & 0.8467$\,^{\pm 0.0295}$ \\
Aya-expanse-32b           & 0.7533$\,^{\pm 0.0353}$ & 0.8467$\,^{\pm 0.0295}$ & 0.9400$\,^{\pm 0.0195}$ \\
Mistral-7B-Instruct-v0.3  & 0.4400$\,^{\pm 0.0407}$ & 0.4867$\,^{\pm 0.0409}$ & 0.6267$\,^{\pm 0.0396}$ \\
\midrule
Jais-family-6p7b-chat     & 0.5667$\,^{\pm 0.0406}$ & 0.7067$\,^{\pm 0.0373}$ & 0.8333$\,^{\pm 0.0305}$ \\
Jais-family-13b-chat      & 0.6067$\,^{\pm 0.0400}$ & 0.7867$\,^{\pm 0.0336}$ & 0.8133$\,^{\pm 0.0319}$ \\
Fanar-1-9B-Instruct       & 0.5800$\,^{\pm 0.0404}$ & 0.7733$\,^{\pm 0.0343}$ & 0.9000$\,^{\pm 0.0246}$ \\
SILMA-9B-Instruct-v1.0    & 0.5600$\,^{\pm 0.0407}$ & 0.5933$\,^{\pm 0.0402}$ & 0.8533$\,^{\pm 0.0290}$ \\
ALLaM-7B-Instruct-preview & 0.6867$\,^{\pm 0.0380}$ & 0.8533$\,^{\pm 0.0290}$ & 0.9333$\,^{\pm 0.0204}$ \\
AceGPT-v2-8B-Chat         & 0.4667$\,^{\pm 0.0409}$ & 0.5667$\,^{\pm 0.0406}$ & 0.7400$\,^{\pm 0.0359}$ \\
\midrule
Claude-Sonnet-4           & 0.8333$\,^{\pm 0.0305}$ & \textbf{0.9533}$\,^{\pm 0.0173}$ & 0.9733$\,^{\pm 0.0132}$ \\
Claude-3.5-Sonnet         & \textbf{0.8533}$\,^{\pm 0.0290}$ & \textbf{0.9533}$\,^{\pm 0.0173}$ & 0.9667$\,^{\pm 0.0147}$ \\
Gemini-1.5-flash          & 0.6533$\,^{\pm 0.0390}$ & 0.8400$\,^{\pm 0.0300}$ & 0.9333$\,^{\pm 0.0204}$ \\
Gemini-2.5-flash-lite & 0.7800$\,^{\pm 0.0339}$ & 0.9333$\,^{\pm 0.0204}$ & \textbf{0.9867}$\,^{\pm 0.0094}$ \\
GPT-4o                    & 0.8400$\,^{\pm 0.0300}$ & \textbf{0.9533}$\,^{\pm 0.0173}$ & 0.9667$\,^{\pm 0.0147}$ \\
GPT-4o-mini               & 0.6467$\,^{\pm 0.0392}$ & 0.8533$\,^{\pm 0.0290}$ & 0.9467$\,^{\pm 0.0184}$ \\
\midrule
\textbf{Average}          & 0.6445 & 0.7852 & 0.8918 \\
\bottomrule
\end{tabular}
}
\caption{Evaluation results for Pragmatic Use, Understanding, and Understanding with Context on a subset of 150 sample idioms from the Kinayat dataset.}
\label{tab:pragmatic-use}
\vspace{-0.2in}
\end{table*}

\section{Ablation and Linguistic Variation Analysis}

\paragraph{Variants of Incorrect Distractors} To better understand the semantic relationship between correct and incorrect explanations, we compute the cosine similarity between their sentence embeddings (using the \texttt{paraphrase-multilingual-mpnet-base-v2} model \cite{reimers-2019-sentence-bert})—comparing correct vs. incorrect explanations generated using the general prompt (Figure~\ref{fig:generate-prompt}), and correct vs. incorrect explanations using the SRL prompt (Figure~\ref{fig:generate-prompt-srl}), separately for proverbs and idioms. The resulting similarity distributions, shown in Figure~\ref{fig:cos-sim-idioms} for idioms (and Figure \ref{fig:cos-sim-proverbs} for proverbs in Appendix~\ref{app:analysis}), reveal noticeable shifts in distribution, but the impact on overall performance was minor. For proverbs, average accuracy slightly decreased with SRL-based distractors (86.57\% → 86.36\%), while for idioms, performance slightly improved (76.29\% → 78.52\%). This suggests that the semantic closeness of distractors can vary depending on the generation strategy, but does not largely affect model performance. For the generation of incorrect distractors in the multiple-choice understanding task, the LLM (GPT-4.1) was prompted only with the correct explanation, not with the idiom or proverb itself. This prevents the model from using knowledge of the idiom/proverb in generating incorrect explanations. Moreover, GPT-4.1 was not among the models evaluated in the experiments. Importantly, despite distractors being generated by a GPT-family model, Claude models consistently outperformed GPT models on MCQ understanding for both idioms and proverbs, indicating that the use of GPT-4.1 did not advantage GPT models or introduce evaluation bias.

\begin{figure}[htbp]
    \centering
    \includegraphics[width=\columnwidth]{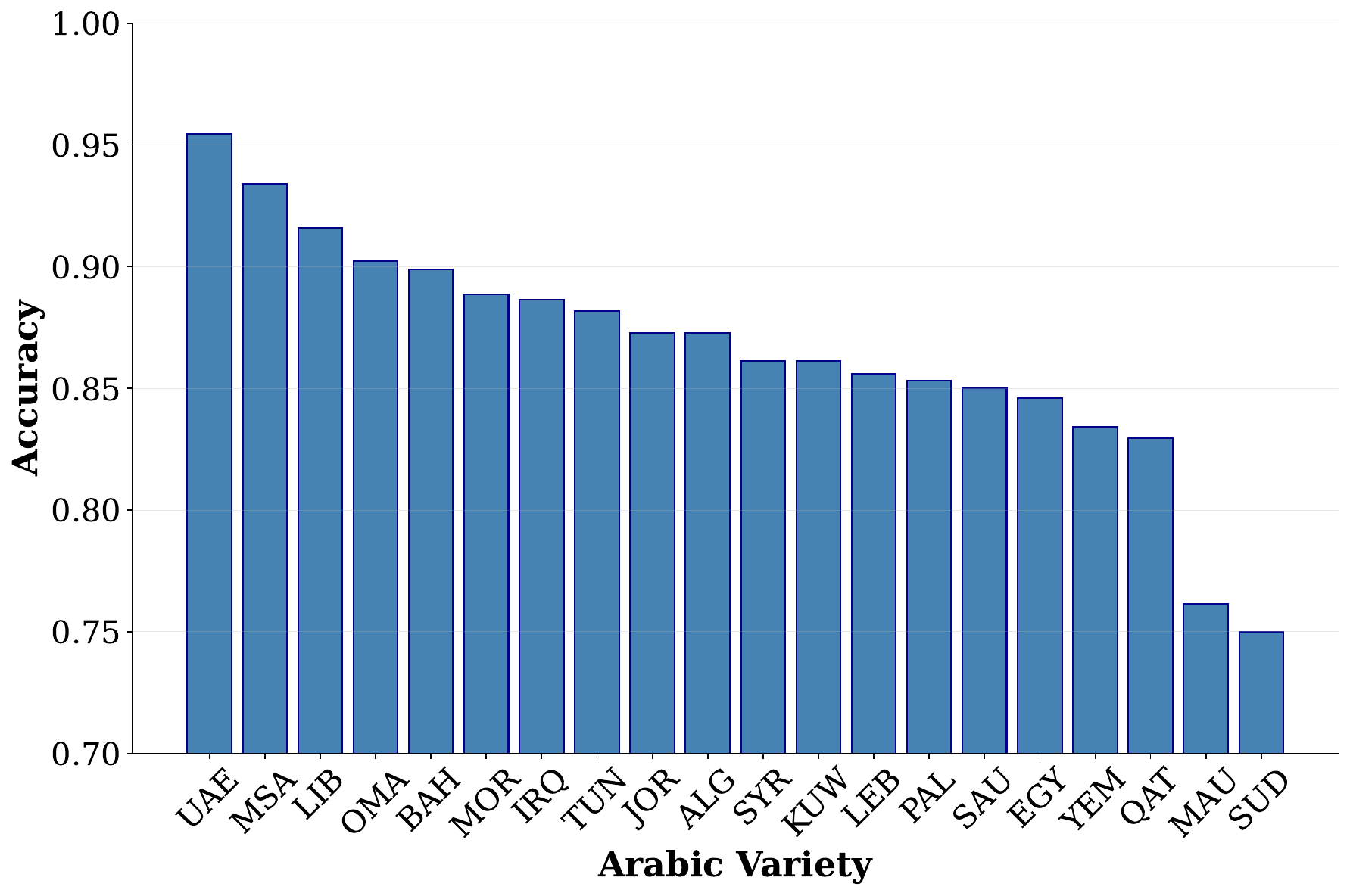}
    \caption{Country-level breakdown of MCQ Understanding Average Accuracy.}
    \label{fig:dialect-avg}
    \vspace{-0.2in}
\end{figure}

\paragraph{Dialectal Breakdown} The dialectal breakdown for the MCQ understanding results in Table~\ref{tab:idioms-srl-comparison} for the Jawaher dataset is shown in Tables~\ref{tab:arabic-dialect-accuracy} and~\ref{tab:arabic-dialect-accuracy-srl} (as well as Figures \ref{fig:dialect-general} and \ref{fig:dialect-srl} in Appendix \ref{app:more-results}. The average of the two tables is presented in Figure \ref{fig:dialect-avg} and Table~\ref{tab:arabic-dialect-accuracy-avg}. The top performing dialects on average were: UAE, MSA, Libya, and Oman. The lowest performing dialects were: Sudan, Mauritania, Qatar, and Yemen. Notably, UAE, Libya, and Oman performed unexpectedly well despite being low-resource dialects, while Egypt, a high-resource dialect, achieved only a mid-range score. One possible explanation is that certain proverbs in the Jawaher dataset are culturally shared across multiple Arab countries, particularly Gulf and Levantine varieties, whereas others are highly localized or unique. Country-level performance varied depending on whether the incorrect distractor was generated with a general or SRL-based prompt, but UAE and MSA consistently ranked in the top four, while Sudan and Mauritania consistently ranked among the lowest three.

\vspace{-5pt}

\paragraph{Arabic vs. Multilingual Models} Table \ref{tab:averages} in Appendix \ref{app:more-results} summarizes the average performance of open source multilingual, open source Arabic, and closed source models in all tasks. Multilingual models generally outperform Arabic models in English tasks by an average of 4.72\% and Arabic tasks by a smaller margin of 2.19\%. However, this trend is not consistent across all multilingual models; for example, Mistral-7B-Instruct performed worse than the Arabic models on most tasks. Closed-source models achieve the highest performance overall, surpassing both Arabic and multilingual models, which may in part be due to their larger size, as their exact parameter counts are not disclosed but they are estimated to be larger than the open-source models evaluated, where the latter models range from 6.7B to 32B parameters. However, Arabic models show notable strengths, outperforming in completion and pragmatic use tasks and achieving slightly higher average scores in generation. The higher completion performance of Arabic models could be partly explained by their pretraining on Arabic text, which likely exposed them to more Arabic proverbs during training.

\vspace{-5pt}

\paragraph{Model Size} A weak positive correlation between model size and performance was observed for understanding tasks, with values of $R^2$ ranging from 0.189 to 0.424, as shown in Figure~\ref{fig:model-size-mcq} and summarized in Table~\ref{tab:mcq_model_size} (Appendix \ref{app:ablation}). The most efficient model for the Arabic task, measured as performance per billion parameters, was ALLaM-7B-Instruct-preview, while the most efficient model for the English task was Qwen2.5-7B-Instruct. The pragmatic use task exhibited a stronger correlation with model size ($R^2 = 0.6$) compared to the MCQ understanding task ($R^2 = 0.265$), with ALLaM-7B-Instruct-preview again being the most efficient model.

\subsection{Error Analysis}

We conducted error analysis on the top eight most frequently incorrectly answered idioms across three tasks—\textit{MCQ Understanding}, \textit{MCQ Understanding with Context}, and \textit{Pragmatic Use}—using a 150-sample subset of Kinayat (corresponding to Table~3). We report the number of models (out of 22) that failed on each idiom in Tables~\ref{tab:mcq_errors}, \ref{tab:mcq_context_errors}, and \ref{tab:pragmatic_errors} in Appendix~\ref{app:error}. Our analysis reveals that providing idiomatic context sentences substantially improves performance for certain idioms in the \textit{MCQ Understanding} task. For instance, \RL{وَلَّعْ} shows dramatic improvement (error frequency: 9$\rightarrow$1), and \RL{عَنْدُه الدُّنْيَا بِالْخُلْخَالْ} improves notably (9$\rightarrow$4). However, contextualization does not consistently benefit all idioms, as performance remains poor for \RL{خَبَرَ أبْيَضْ}, \RL{سَكْرِةْ يَنِّي}, and \RL{الصَّبَاحْ رَبَاحْ} despite the addition of context. In contrast, error patterns for the \textit{Pragmatic Use} task differ markedly from those observed in \textit{MCQ Understanding}, indicating a dissociation between comprehension and pragmatic use capabilities. Models that correctly identify idiom meanings often fail to appropriately use those same idioms in context, highlighting the limitations in pragmatic competence discussed in Section~\ref{sec:prag}.

\section{Conclusion}

We presented a comprehensive evaluation of large language models' ability to understand and pragmatically use figurative expressions that encode local knowledge and social nuance. We evaluated diverse LLMs on Egyptian Arabic idioms, multidialectal Arabic proverbs, and English proverbs across tasks assessing contextual understanding, pragmatic use, and connotation interpretation.
Results revealed a consistent performance hierarchy, with English proverbs outperforming Arabic proverbs, and both outperforming Egyptian idioms. Pragmatic use emerged as significantly more challenging than understanding, and models struggled to capture connotative meaning appropriately.
Figurative language thus serves as an effective diagnostic for cultural reasoning, revealing that while LLMs often interpret figurative meaning, they face major challenges in using it appropriately. To support future research, we released \textit{Kinayat}, the first dataset of Egyptian Arabic idioms designed for both figurative understanding and pragmatic use evaluation. Future work should extend pragmatic evaluation to proverbs and idioms in multiple dialects, and assess pragmatic use in free-form generation to better capture the ability of models to produce culturally and contextually appropriate figurative expressions.

\clearpage
\section*{Limitations}
\paragraph{Scope} Although our study provides new insights into the ability of large language models to understand and use Egyptian Arabic idioms and proverbs, it is important to acknowledge several limitations. First, our benchmark focuses exclusively on Egyptian idioms. Although Egyptian Arabic is a high-resource dialect with significant representation online and in available datasets, the observed challenges that models face in the pragmatic use of figurative language underscore the necessity for similar resources and benchmarks in other Arabic dialects, many of which are lower-resource and may present even greater challenges. Additionally, since the source used for our idiom collection was published in 1949, some idioms may be outdated as language evolves over time, and idiom usage varies across generations and social groups. However, this does not necessarily render the idioms obsolete: many remain in active use among particular speaker communities. That said, we took steps to mitigate this concern in our experiments. Specifically, our pragmatic evaluation was conducted on a subset of 150 more familiar idioms out of 325, ensuring that the core findings are not driven by highly unfamiliar expressions.

Second, our evaluation is limited to idioms and proverbs, which represent only a subset of figurative language phenomena. Figurative language encompasses a broader range of expressions, including metaphors, similes, hyperboles, irony, sarcasm, and other culturally specific figures of speech. Future work should extend the evaluation to these additional forms to provide a more comprehensive assessment of model capabilities in the understanding and generation of Arabic figurative language.

\paragraph{Connotations Limitations} The connotations task also carries assumptions and limitations: the human labeling of idioms and proverbs can be somewhat subjective, particularly when the connotation is weak or context-dependent.

This ambiguity reflects broader challenges in NLP annotation, where guidelines may fail to account for the multiplicity of cultural perspectives, leading to mismatches between annotator intent and dataset construction \cite{liu-etal-2025-culturally}. Research has repeatedly shown that annotator bias—shaped by personal and cultural background—can skew labeled data, embedding specific worldviews into machine learning systems and perpetuating representational inequalities \cite{sap-etal-2022-annotators, stochasticparrots, truthisalie}. Attempts to enforce consistency through guidelines and inter-annotator agreement often overlook that language is deeply contextual and values are not universally shared \cite{stochasticparrots}.

Anthropological critiques urge moving beyond the pursuit of an “objective” gold standard in annotation, advocating instead for multi-layered or perspectivist models that document the diversity of meanings and cultural nuances present in a corpus \cite{stochasticparrots,truthisalie}. Struggling with annotation from an anthropological standpoint underscores the importance of recognizing cultural context, annotator subjectivity, and the limits of value assignment in NLP, serving as a reminder that data is more than numbers: it is a reflection of lived sociocultural realities \cite{liu-etal-2025-culturally, truthisalie}.

\section*{Ethical Considerations}
\paragraph{Offensive Content Elimination} The complete list of idioms in the Al-Kinayat Al-'Amiyya \cite{Taymour1949} book was manually reviewed and seven samples were removed that we considered inappropriate or offensive.

\paragraph{Licenses} The Kinayat dataset is released under a CC-BY 4.0 license\footnote{\url{https://creativecommons.org/licenses/by/4.0/}}, permitting use, distribution, and adaptation with attribution.

\paragraph{Annotations} Two of the annotators are coauthors of this paper, while the third annotator was compensated for their time at a rate of 50 USD for an estimated three hours of work, which is slightly above the minimum wage in the US. No personal information from annotators is included in the Kinayat dataset.

\bibliography{custom}

\begin{thebibliography}{54}
\providecommand{\natexlab}[1]{#1}

\bibitem[{Alghallabi et~al.(2025)Alghallabi, Thawkar, Ghaboura, More, Thawakar, Cholakkal, Khan, and Anwer}]{alghallabi2025fannflopmultigenremultiera}
Wafa Alghallabi, Ritesh Thawkar, Sara Ghaboura, Ketan More, Omkar Thawakar, Hisham Cholakkal, Salman Khan, and Rao~Muhammad Anwer. 2025.
\newblock \href {https://arxiv.org/abs/2505.18152} {Fann or flop: A multigenre, multiera benchmark for arabic poetry understanding in llms}.
\newblock \emph{Preprint}, arXiv:2505.18152.

\bibitem[{Alsiyat and Piao(2020)}]{alsiyat-piao-2020-metaphorical}
Israa Alsiyat and Scott Piao. 2020.
\newblock \href {https://aclanthology.org/2020.lrec-1.604/} {Metaphorical expressions in automatic {A}rabic sentiment analysis}.
\newblock In \emph{Proceedings of the Twelfth Language Resources and Evaluation Conference}, pages 4911--4916, Marseille, France. European Language Resources Association.

\bibitem[{Alwajih et~al.(2025)Alwajih, El~Mekki, Magdy, Elmadany, Nacar, Nagoudi, Abdel-Salam, Atwany, Nafea, Yahya, Alhamouri, Alsayadi, Zayed, Shatnawi, Sibaee, Ech-chammakhy, Al-Dhabyani, Ali, Jarraya, El-Shangiti, Alraeesi, AL-Ghrawi, Al-Batati, Mohamed, Elgindi, Saeed, Atou, Yahia, Bouayad, Machrouh, Makouar, Alkawi, Mohamed, Abdelfadil, Ounnoughene, Rouabhia, Assi, Sorkatti, Tourad, Koubaa, Berrada, Jarrar, Shehata, and Abdul-Mageed}]{alwajih-etal-2025-palm}
Fakhraddin Alwajih, Abdellah El~Mekki, Samar~Mohamed Magdy, AbdelRahim~A. Elmadany, Omer Nacar, El~Moatez~Billah Nagoudi, Reem Abdel-Salam, Hanin Atwany, Youssef Nafea, Abdulfattah~Mohammed Yahya, Rahaf Alhamouri, Hamzah~A. Alsayadi, Hiba Zayed, Sara Shatnawi, Serry Sibaee, Yasir Ech-chammakhy, Walid Al-Dhabyani, Marwa~Mohamed Ali, Imen Jarraya, and 25 others. 2025.
\newblock \href {https://doi.org/10.18653/v1/2025.acl-long.1579} {Palm: A culturally inclusive and linguistically diverse dataset for {A}rabic {LLM}s}.
\newblock In \emph{Proceedings of the 63rd Annual Meeting of the Association for Computational Linguistics (Volume 1: Long Papers)}, pages 32871--32894, Vienna, Austria. Association for Computational Linguistics.

\bibitem[{Alyafeai et~al.(2024)Alyafeai, Almubarak, Ashraf, Alnuhait, Alshahrani, Abdulrahman, Ahmed, Gawah, Saleh, Ghaleb, Ali, and Al-shaibani}]{alyafeai-etal-2024-cidar}
Zaid Alyafeai, Khalid Almubarak, Ahmed Ashraf, Deema Alnuhait, Saied Alshahrani, Gubran Abdulrahman, Gamil Ahmed, Qais Gawah, Zead Saleh, Mustafa Ghaleb, Yousef Ali, and Maged Al-shaibani. 2024.
\newblock \href {https://doi.org/10.18653/v1/2024.findings-acl.764} {{CIDAR}: Culturally relevant instruction dataset for {A}rabic}.
\newblock In \emph{Findings of the Association for Computational Linguistics: ACL 2024}, pages 12878--12901, Bangkok, Thailand. Association for Computational Linguistics.

\bibitem[{{Anthropic}(2024)}]{anthropic-claude35sonnet-announcement-2024}
{Anthropic}. 2024.
\newblock \href {https://www.anthropic.com/news/claude-3-5-sonnet} {Introducing claude 3.5 sonnet}.

\bibitem[{{Anthropic}(2025)}]{anthropic-claude4-announcement-2025}
{Anthropic}. 2025.
\newblock \href {https://www.anthropic.com/news/claude-4} {Introducing claude 4}.

\bibitem[{Aroyo and Welty(2015)}]{truthisalie}
Lora Aroyo and Chris Welty. 2015.
\newblock \href {https://doi.org/10.1609/aimag.v36i1.2564} {Truth is a lie: Crowd truth and the seven myths of human annotation}.
\newblock \emph{AI Mag.}, 36(1):15–24.

\bibitem[{Azime et~al.(2025)Azime, Tonja, Belay, Chanie, Balcha, Abadi, Ademtew, Nerea, Yadeta, Geremew, Tesfu, Slusallek, Solorio, and Klakow}]{azime-etal-2025-proverbeval}
Israel~Abebe Azime, Atnafu~Lambebo Tonja, Tadesse~Destaw Belay, Yonas Chanie, Bontu~Fufa Balcha, Negasi~Haile Abadi, Henok~Biadglign Ademtew, Mulubrhan~Abebe Nerea, Debela~Desalegn Yadeta, Derartu~Dagne Geremew, Assefa~Atsbiha Tesfu, Philipp Slusallek, Thamar Solorio, and Dietrich Klakow. 2025.
\newblock \href {https://doi.org/10.18653/v1/2025.findings-naacl.350} {{P}roverb{E}val: Exploring {LLM} evaluation challenges for low-resource language understanding}.
\newblock In \emph{Findings of the Association for Computational Linguistics: NAACL 2025}, pages 6250--6266, Albuquerque, New Mexico. Association for Computational Linguistics.

\bibitem[{Bari et~al.(2025)Bari, Alnumay, Alzahrani, Alotaibi, Alyahya, AlRashed, Mirza, Alsubaie, Alahmed, Alabduljabbar, Alkhathran, Almushayqih, Alnajim, Alsubaihi, Mansour, Hassan, Alrubaian, Alammari, Alawami, Al-Thubaity, Abdelali, Kuriakose, Abujabal, Al-Twairesh, Alowisheq, and Khan}]{bari2025allam}
M~Saiful Bari, Yazeed Alnumay, Norah~A. Alzahrani, Nouf~M. Alotaibi, Hisham~Abdullah Alyahya, Sultan AlRashed, Faisal~Abdulrahman Mirza, Shaykhah~Z. Alsubaie, Hassan~A. Alahmed, Ghadah Alabduljabbar, Raghad Alkhathran, Yousef Almushayqih, Raneem Alnajim, Salman Alsubaihi, Maryam~Al Mansour, Saad~Amin Hassan, Dr.~Majed Alrubaian, Ali Alammari, Zaki Alawami, and 7 others. 2025.
\newblock \href {https://openreview.net/forum?id=MscdsFVZrN} {{ALL}am: Large language models for arabic and english}.
\newblock In \emph{The Thirteenth International Conference on Learning Representations}.

\bibitem[{Bender et~al.(2021)Bender, Gebru, McMillan-Major, and Shmitchell}]{stochasticparrots}
Emily~M. Bender, Timnit Gebru, Angelina McMillan-Major, and Shmargaret Shmitchell. 2021.
\newblock \href {https://doi.org/10.1145/3442188.3445922} {On the dangers of stochastic parrots: Can language models be too big?}
\newblock In \emph{Proceedings of the 2021 ACM Conference on Fairness, Accountability, and Transparency}, FAccT '21, page 610–623, New York, NY, USA. Association for Computing Machinery.

\bibitem[{Chakrabarty et~al.(2022{\natexlab{a}})Chakrabarty, Choi, and Shwartz}]{rocket_science}
Tuhin Chakrabarty, Yejin Choi, and Vered Shwartz. 2022{\natexlab{a}}.
\newblock \href {https://doi.org/10.1162/tacl_a_00478} {It’s not rocket science: Interpreting figurative language in narratives}.
\newblock \emph{Transactions of the Association for Computational Linguistics}, 10:589--606.

\bibitem[{Chakrabarty et~al.(2022{\natexlab{b}})Chakrabarty, Saakyan, Ghosh, and Muresan}]{chakrabarty-etal-2022-flute}
Tuhin Chakrabarty, Arkadiy Saakyan, Debanjan Ghosh, and Smaranda Muresan. 2022{\natexlab{b}}.
\newblock \href {https://doi.org/10.18653/v1/2022.emnlp-main.481} {{FLUTE}: Figurative language understanding through textual explanations}.
\newblock In \emph{Proceedings of the 2022 Conference on Empirical Methods in Natural Language Processing}, pages 7139--7159, Abu Dhabi, United Arab Emirates. Association for Computational Linguistics.

\bibitem[{Cheng et~al.(2024)Cheng, Yan, Wang, Li, Yu, Zheng, Tu, Xu, and Han}]{cheng2024potentiallimitationsllmscapturing}
Ning Cheng, Zhaohui Yan, Ziming Wang, Zhijie Li, Jiaming Yu, Zilong Zheng, Kewei Tu, Jinan Xu, and Wenjuan Han. 2024.
\newblock \href {https://arxiv.org/abs/2405.06410} {Potential and limitations of llms in capturing structured semantics: A case study on srl}.
\newblock \emph{Preprint}, arXiv:2405.06410.

\bibitem[{Comanici et~al.(2025)Comanici, Bieber, Schaekermann, Pasupat, Sachdeva, Dhillon, Blistein, Ram, Zhang, Rosen, Marris, Petulla, Gaffney, Aharoni, Lintz, Pais, Jacobsson, Szpektor, Jiang, Haridasan, Omran, Saunshi, Bahri, Mishra, Chu, Boyd, Hekman, Parisi, Zhang, Kawintiranon, Bedrax-Weiss, Wang, Xu, Purkiss, Mendlovic, Deutel, Nguyen, Langley, Korn, Rossazza, Ramé, Waghmare, Miller, Byrd, Sheshan, Bhardwaj, Janus, Rissa, Horgan, Silver, Wahid, Brin, Raimond, Kloboves, Wang, Gundavarapu, Shumailov, Wang, Pajarskas, Heyward, Nikoltchev, Kula, Zhou, Garrett, Kafle, Arik, Goel, Yang, Park, Kojima, Mahmoudieh, Kavukcuoglu, Chen, Fritz, Bulyenov, Roy, Paparas, Shemtov, Chen, Strudel, Reitter, Roy, Vlasov, Ryu, Leichner, Yang, Mariet, Vnukov, Sohn, Stuart, Liang, Chen, Rawlani, Koh, Co-Reyes, Lai, Banzal, Vytiniotis, Mei, Cai, Badawi, Fry, Hartman, Zheng, Jia, Keeling, Louis, Chen, Robles, Hung, Zhou, Saxena, Goenka, Ma, Fisher, Taege, Graves, Steiner, Li, Nguyen, Sukthankar, Stanton, Eslami, Shen, Akin,
  Guseynov, Zhou, Alayrac, Joulin, Farkash, Thapliyal, Roller, Shazeer, Davchev, Koo, Forbes-Pollard, Audhkhasi, Farquhar, Gilady, Song, Aslanides, Mendolicchio, Parrish, Blitzer, Gupta, Ju, Yang, Datta, Tacchetti, Mehta, Dibb, Gupta, Piccinini, Hadsell, Rajayogam, Jiang, Griffin, Sundberg, Hayes, Frolov, Xie, Zhang, Dasgupta, Kalra, Shani, Macherey, Huang, MacDermed, Duddu, Zacchello, Yang, Lo, Hui, Kastelic, Gasaway, Tan, Yue, Barrio, Wieting, Yang, Nystrom, Demmessie, Levskaya, Viola, Tekur, Billock, Necula, Joshi, Schaeffer, Lokhande, Sorokin, Shenoy, Chen, Collier, Li, Bos, Wichers, Lee, Pouget, Thangaraj, Axiotis, Crone, Sterneck, Chinaev, Krakovna, Ferludin, Gemp, Winkler, Goldberg, Korotkov, Xiao, Mehrotra, Mariserla, Piratla, Thurk, Pham, Ma, Senges, Kumar, Meyer, Talius, Pierse, Sandhu, Toma, Lin, Nath, Stone, Sadigh, Gupta, Guez, Singh, Thomas, Duerig, Gong, Tanburn, Zhang, Dao, Hammad, Xie, Rijhwani, Murdoch, Kim, Thompson, Cheng, Sohn, Sprechmann, Xu, Tadepalli, Young, Zhang, Srinivasan,
  Aperghis, Ayyar, Fitoussi, Burnell, Madras, Dusenberry, Xiong, Oguntebi, Albrecht, Bornschein, Mitrović, Dimarco, Shamanna, Shah, Sezener, Upadhyay, Lacey, Schiff, Baur, Ganapathy, Schnider, Wirth, Schenck, Simanovsky, Tan, Fränken, Duan, Mankalale, Dhawan, Sequeira, Wei, Goel, Unlu, Zhu, Sun, Balashankar, Shuster, Umekar, Alnahlawi, van~den Oord, Chen, Zhai, Dai, Lee, Doi, Zilka, Vallu, Shrivastava, Lee, Husain, Zhuang, Cohen-Addad, Barber, Atwood, Sadovsky, Wellens, Hand, Rajendran, Turker, Carey, Xu, Soltau, Li, Song, Li, Kemaev, Brown, Burns, Patraucean, Stanczyk, Aravamudhan, Blondel, Noga, Blanco, Song, Isard, Sharma, Hayes, Badawy, Lamp, Laish, Kozlova, Chan, Singla, Sunkara, Upadhyay, Liu, Bai, Wilkiewicz, Zlocha, Liu, Li, Li, Barak, Raboshchuk, Choi, Liu, Jue, Sharma, Marzoca, Busa-Fekete, Korsun, Elisseeff, Shen, Carthy, Lamerigts, Hosseini, Lin, Chen, Yang, Chauhan, Omernick, Jia, Zainullina, Hassabis, Vainstein, Amid, Zhou, Votel, Vértes, Li, Zhou, Lazaridou, McMahan, Narayanan, Soyer, Basu,
  Lee, Perozzi, Cao, Berrada, Arya, Chen, Katrina, Xu, Lochbrunner, Hofer, Sharifzadeh, Wu, Goldman, Awasthi, Wang, Wu, Sha, Zhang, Mikuła, Graziano, Mcloughlin, Giannoumis, Namiki, Malik, Radebaugh, Hall, Leal-Cavazos, Chen, Sindhwani, Kao, Greene, Griffith, Welty, Montgomery, Yoshino, Yuan, Goodman, Michaely, Lee, Sawhney, Chen, Zheng, Shum, Savinov, Pot, Pak, Zadimoghaddam, Bhatnagar, Lewenberg, Kutzman, Liu, Katzen, Selier, Djolonga, Lepikhin, Xu, Liang, Tan, Schillings, Ersoy, Blois, Bandemer, Singh, Lebedev, Joshi, Brown, Palmer, Pathak, Jalan, Zubach, Lall, Parker, Gunjan, Rogulenko, Sanghai, Leng, Egyed, Li, Ivanova, Andriopoulos, Xie, Rosenfeld, Wright, Sharma, Geng, Wang, Kwei, Pan, Zhang, Wang, Liu, Yeung, Cole, Rosenberg, Yang, Chen, Polovets, Nair, Saxena, Smith, yiin Chang, Mahendru, Grant, Iyer, Cai, McGiffin, Shen, Walton, Girgis, Woodman, Ke, Kwong, Rouillard, Rao, Li, Xu, Prost, Zou, Ji, Magni, Liechty, Calian, Ramachandran, Krivokon, Huang, Chen, Hauth, Ilić, Xi, Lim, Ion, Moradi,
  Toksoz-Exley, Bullard, Allamanis, Yang, Wang, Hong, Gergely, Li, Mittal, Kovalev, Ungureanu, Labanowski, Wassenberg, Lacasse, Cideron, Dević, Marsden, Nguyen, Fink, Zhong, Kiyono, Ivanov, Ma, Bain, Yalasangi, She, Petrushkina, Lunayach, Bromberg, Hodkinson, Meshram, Vlasic, Kyker, Xu, Stanway, Yang, Zhao, Tung, Odoom, Fujii, Gilmer, Kim, Halim, Le, Bohnet, El-Sayed, Neyshabur, Reynolds, Reich, Xu, Moreira, Sharma, Liu, Hosseini, Raisinghani, Su, Lao, Formoso, Gelmi, Gueta, Dey, Gribovskaya, Ćevid, Mudgal, Bingham, Wang, Kumar, Cullum, Han, Bousmalis, Cedillo, Chu, Magay, Michel, Hlavnova, Calandriello, Ariafar, Yao, Sehwag, Vezer, Lago, Zhu, Rubenstein, Porter, Baddepudi, Riva, Istin, Yeh, Li, Howard, Jha, Chen, de~Liedekerke, Ahmed, Rodriguez, Bhatia, Wang, Elqursh, Klinghoffer, Chen, Kohli, I, Zhang, Nado, Chen, Chen, Zhang, Singh, Hillier, Lebron, Tao, Liu, Dulac-Arnold, Zhang, Narayan, Liu, Firat, Bhowmick, Liu, Zhang, Zhang, Rotival, Howard, Sinha, Grushetsky, Beyret, Gopalakrishnan, Zhao, He,
  Payrits, Nabulsi, Zhang, Chen, Lee, Fallen, Gollapudi, Zhou, Pavetić, Köppe, Huang, Pasumarthi, Fernando, Fischer, Ćurko, Gao, Svensson, Stone, Qureshi, Sinha, Kulshreshtha, Matysiak, Mao, Saroufim, Faust, Duan, Fidel, Katircioglu, Kaufman, Shah, Kong, Bapna, Weisz, Dunleavy, Dutta, Liu, Chaabouni, Parada, Wu, Belias, Bissacco, Fort, Xiao, Huot, Knutsen, Blau, Li, Prendki, Love, Chow, Charoenpanit, Shimokawa, Coriou, Gregor, Izo, Akula, Pinto, Hahn, Paulus, Guo, Sharma, Hsieh, Chukwuka, Hashimoto, Rauschmayr, Wu, Angermueller, Wang, Gerlach, Pliskin, Mirylenka, Ma, Baugher, Gale, Bijwadia, Rakićević, Wood, Park, Chang, Seal, Tar, Krasowiak, Song, Stephanov, Wang, Maggioni, Lin, Wu, Paul, Jiang, Agrawal, Piot, Feng, Kim, Doshi, Lai, Chuqiao, Xu, Vikram, Chelba, Krause, Zhuang, Rae, Denk, Collister, Weerts, Luo, Lu, Garnes, Gupta, Spitz, Hassidim, Liang, Shafran, Humphreys, Vassigh, Wallis, Shejwalkar, Perez-Nieves, Hornung, Tan, Westberg, Ly, Zhang, Farris, Park, Kosik, Cankara, Maksai, Xu, Cassirer,
  Caelles, Abdolmaleki, Chiang, Fabrikant, Shetty, He, Giménez, Hashemi, Panthaplackel, Kulizhskaya, Deshmukh, Pighin, Alazard, Jindal, Noury, S, Qin, Dotiwalla, Spencer, Babaeizadeh, Chen, Mehta, Lees, Leach, Koanantakool, Akolzin, Comanescu, Ahn, Svyatkovskiy, Mustafa, D'Ambrosio, Garlapati, Lamblin, Agarwal, Song, Sessa, Coquinot, Maggs, Masoom, Pitta, Wang, Morris-Suzuki, Porter, Jia, Dudek, R, Paduraru, Ansell, Bolukbasi, Lu, Ganeshan, Wang, Griffiths, Benenson, He, Swirhun, Papamakarios, Chawla, Sengupta, Wang, Milutinovic, Mordatch, Jia, Smith, Ng, Nigam, Young, Vušak, Hechtman, Goenka, Zipori, Ayoub, Popat, Acharya, Yu, Bloxwich, Song, Roit, Li, Boag, Nayakanti, Chandra, Ding, Mehta, Hope, Zhang, Shtacher, Badola, Nakashima, Sozanschi, Comşa, Žužul, Caveness, Odell, Watson, de~Cesare, Lippe, Lockhart, Verma, Chen, Sun, Zhuo, Shah, Gupta, Muzio, Niu, Zait, Singh, Gaba, Ye, Ramachandran, Saleh, Popa, Dubey, Liu, Javanmardi, Epstein, Hemsley, Green, Ranka, Cohen, Fu, Ghemawat, Borovik, Martens,
  Chen, Shyam, Pinto, Yang, Ţifrea, Du, Gong, Agarwal, Kim, Frank, Shah, Song, Deng, Mikhalap, Chatziprimou, Chung, Creswell, Zhang, Jun, Lebsack, Truong, Andačić, Yona, Fornoni, Rong, Toropov, Soudagar, Audibert, Zaiem, Abbas, Rusu, Potluri, Weng, Kementsietsidis, Tsitsulin, Peng, Ha, Jain, Latkar, Ivanov, McLean, GP, Venkataraman, Liu, Krishnan, D'sa, Yogev, Collins, Lee, Ho, Doersch, Yona, Gao, Ferreira, Ozturel, Muckenhirn, Zheng, Balasubramaniam, Bansal, van~den Driessche, Eiger, Haykal, Misra, Goyal, Martins, Leung, Valfridsson, Flynn, Bishop, Pang, Halpern, Yu, Moore, Yuvein, Zhu, Thiagarajan, Drori, Xiao, Dery, Jagerman, Lu, Ge, Aggarwal, Khare, Tran, Elyada, Alet, Rubin, Chou, Tian, Bai, Chan, Lew, Misiunas, Bilal, Ray, Raghuram, Castro-Ros, Carpenter, Zheng, Kilgore, Broder, Xue, Kallakuri, Dua, Yuen, Chien, Schultz, Agrawal, Tsarfaty, Hu, Kannan, Marcus, Kothari, Sun, Horn, Bošnjak, Naeem, Hirsch, Chiang, Fang, Han, Wang, Hora, He, Lučić, Changpinyo, Tripathi, Youssef, Kwak, Schlattner,
  Graves, Leblond, Zeng, Andreassen, Rasskin, Song, Cao, Oh, Hoffman, Skut, Zhang, Stritar, Cai, Khanna, Wang, Sharma, Reisswig, Jun, Prasad, Sholokhova, Singh, Rosenthal, Ruoss, Beaufays, Kirmani, Chen, Schalkwyk, Herzig, Kim, Jacob, Vincent, Reyes, Balazevic, Hussenot, Schneider, Barnes, Castro, Babbula, Green, Cabi, Duduta, Driess, Galt, Velan, Wang, Jiao, Mauger, Phan, Patel, Galić, Chang, Marcus, Harvey, Salazar, Dabir, Sheth, Mandhane, Sedghi, Willcock, Zandieh, Prabhakara, Amini, Miech, Stone, Nicosia, Niemczyk, Xiao, Kim, Kwasiborski, Verma, Oflazer, Hirnschall, Sung, Liu, Everett, Bakker, Ágoston Weisz, Wang, Sampathkumar, Shaham, Xu, Altun, Wang, Saeki, Chen, Taropa, Vasanth, Austin, Huang, Petrovic, Dou, Golovin, Rozhdestvenskiy, Culp, Wu, Sano, Jain, Proskurnia, Cevey, Ruiz, Patil, Mirzazadeh, Ni, Snaider, Fan, Fréchette, Pierigiovanni, Iqbal, Lee, Fantacci, Xing, Wang, Irpan, Raposo, Luan, Chen, Ganapathy, Hui, Nie, Guyon, Ge, Vij, Zheng, Lee, Castaño, Baatarsukh, Ibagon, Chronopoulou,
  FitzGerald, Viswanadha, Huda, Moroshko, Stoyanov, Kolhar, Vaucher, Watts, Kuncoro, Michalewski, Kambala, Batsaikhan, Andreev, Jurenka, Le, Chen, Jishi, Chakera, Chen, Kini, Yadav, Siddhant, Labzovsky, Lakshminarayanan, Bostock, Botadra, Anand, Bishop, Conway-Rahman, Agarwal, Donchev, Singhal, de~Chaumont~Quitry, Ponomareva, Agrawal, Ni, Krishna, Samsikova, Karro, Du, von Glehn, Lu, Choquette-Choo, Qin, Zhang, Li, Tyam, Mishra, Lowe, Ji, Wang, Faruqui, Slone, Dalibard, Narayanaswamy, Lambert, Manzagol, Karliner, Bolt, Lobov, Kusupati, Ye, Yang, Zen, George, Bhutani, Lacombe, Riachi, Bansal, Soh, Gao, Yu, Yu, Nottage, Rojas-Esponda, Noraky, Gupta, Kotikalapudi, Chang, Deur, Graur, Mossin, Farnese, Figueira, Moufarek, Huang, Zochbauer, Ingram, Chen, Wu, Puigdomènech, Rechis, Yu, Padmanabhan, Zhu, ling Ko, Banino, Daruki, Selvan, Bhaswar, Diaz, Su, Scellato, Brennan, Han, Chung, Agrawal, Khandelwal, Sim, Lustman, Ritter, Guu, Xia, Jain, Wang, Hill, Rossini, Kostelac, Misiunas, Sabne, Kim, Iscen, Wang, Leal,
  Sreevatsa, Evci, Warmuth, Joshi, Suo, Lottes, Honke, Jou, Karp, Hu, Sahni, Taïga, Kong, Ghosh, Wang, Pavagadhi, Axelsson, Grigorev, Siegler, Lin, Wang, Parisotto, Maddineni, Subudhi, Ben-David, Pochernina, Keller, Avrahami, Yuan, Mehta, Liu, Yang, Kan, Lee, Funkhouser, Cheng, Shi, Sharma, Kelley, Eyal, Malkov, Tallec, Bahat, Yan, Xintian, Wu, Lindner, Wu, Caciularu, Luo, Jenatton, Zaman, Bi, Kornakov, Mallya, Ikeda, Karo, Singh, Evans, Netrapalli, Nallatamby, Tian, Assael, Raunak, Carbune, Bica, Madmoni, Cattle, Grover, Somandepalli, Lall, Vázquez-Reina, Patana, Mu, Talluri, Tran, Aggarwal, Skerry-Ryan, Xu, Burrows, Pan, Yvinec, Lu, Zhang, Nguyen, Mu, Barcik, Ran, Beltrone, Choromanski, Kharrat, Albanie, Purser-haskell, Bieber, Zhang, Wang, Hudson, Zhang, Fu, Mauerer, Bateni, Maschinot, Wang, Zhu, Pillai, Weyand, Liu, Akerlund, Bertsch, Premachandran, Jin, Roulet, de~Boursac, Mittal, Ndebele, Karadzhov, Ghalebikesabi, Liang, Wu, Cong, Ghelani, Singh, Fatemi, Warren, Chen, Kwong, Kolganov, Li, Song, Kuang,
  Miryoosefi, Webster, Wendt, Socala, Su, Mendonça, Gupta, Li, Tsai, Qiong, Hu, Kang, Chen, Girgin, Xian, Lee, Ramsden, Baker, Elish, Krayvanova, Joshi, Simsa, Yang, Ambroszczyk, Ghosh, Kar, Shangguan, Yamamori, Akulov, Brock, Tang, Vashishtha, Munoz, Steiner, Andra, Eppens, Feng, Kobayashi, Goldshtein, Mahdy, Wang, Jilei, Wang, Killam, Kwiatkowski, Kopparapu, Zhan, Jia, Bendebury, Luo, Recasens, Knight, Chen, Patel, Li, Withbroe, Weesner, Bhatia, Ren, Eisenbud, Songhori, Sun, Choma, Kementsietsidis, Manning, Roark, Farhan, Feng, Tatineni, Cobon-Kerr, Li, Hendricks, Noble, Breaux, Kushman, Peng, Xue, Tobin, Rogers, Lipschultz, Alberti, Vlaskin, Dehghani, Sharma, Warkentin, Lee, Uria, Juan, Chandorkar, Sheftel, Liu, Davoodi, Pigem, Dhamdhere, Ross, Hoech, Mahdieh, Liu, Li, McCafferty, Liu, Mircea, Song, Savant, Saade, Cherry, Hellendoorn, Goyal, Pucciarelli, Torres, Yahav, Lee, Sjoesund, Kirov, Chang, Ghoshal, Li, Baechler, Pereira, Sainath, Boral, Grewe, Halumi, Phu, Shen, Ribeiro, Varma, Kaskasoli,
  Feinberg, Potti, Kahn, Wisniewski, Mohamed, Hrafnkelsson, Shahriari, Lespiau, Patel, Yeung, Paine, Mei, Ramirez, Shivanna, Zhong, Woodward, Tubone, Khan, Chen, Nielsen, Ionescu, Prabhu, Gao, Wang, Augenstein, Subramaniam, Chang, Iliopoulos, Luo, Khan, Kuo, Teplyashin, Perot, Kilpatrick, Globerson, Yu, Siddiqui, Sukhanov, Kandoor, Gupta, Andreetto, Ambar, Kim, Wesołowski, Perrin, Limonchik, Fan, Stephan, Stewart-Binks, Kappedal, He, Cogan, Datta, Zhou, Ye, Kieliger, Ramalho, Kastner, Mentzer, Ko, Suggala, Zhou, Butt, Strejček, Belenki, Venugopalan, Ling, Eltyshev, Deng, Kovacs, Raghavachari, Dai, Schuster, Schwarcz, Nguyen, Nguyen, Buttimore, Mallick, Gandhe, Benjamin, Jastrzebski, Yan, Basu, Apps, Edkins, Allingham, Odisho, Kocisky, Zhao, Xue, Reddy, Anastasiou, Atias, Redmond, Milan, Heess, Schmit, Dafoe, Andor, Gangwani, Dragan, Zhang, Kachra, Wu, Xue, Aydin, Liu, Zhou, Malihi, Wu, Gopal, Schumann, Stys, Wang, Olšák, Liu, Schallhart, Mao, Brady, Xu, Mery, Sitawarin, Velusamy, Cobley, Zhai, Walder,
  Katz, Jawahar, Kulkarni, Yang, Paszke, Wang, Damoc, Borsos, Smith, Li, Gupta, Kapishnikov, Prakash, Luisier, Agarwal, Grathwohl, Chen, Han, Mehta, Over, Azizi, Meng, Santo, Zheng, Shapiro, Petrovski, Hui, Ghafouri, Snoek, Qin, Jordan, Sikora, Malmaud, Kuang, Świetlik, Sang, Shi, Li, Rosenberg, Zhao, Crawford, Peter, Lei, Garcia, Le, Wang, Amelot, Orr, Kacham, Alon, Tyen, Arora, Lyon, Kurakin, Ly, Guidroz, Yan, Panigrahy, Xu, Kagohara, Cheng, Noland, Lee, Lee, Yip, Wang, Nehoran, Bykovsky, Shan, Bhagatwala, Yan, Tan, Garrido, Ethier, Hurley, Vesom, Chen, Qiao, Nayyar, Walker, Sandhu, Rosca, Swisher, Dektiarev, Dillon, Muraru, Tragut, Myaskovsky, Reid, Velic, Xiao, George, Brand, Li, Yu, Gu, Deng, Aubet, Yeganeh, Alcober, Smith, Cohn, McKinney, Tschannen, Sampath, Cheon, Luo, Liu, Orbay, Peng, Botea, Zhang, Yoon, Magalhaes, Stradomski, Mackinnon, Hemingray, Venkatesan, May, Kim, Druinsky, Ye, Xu, Huang, Abdallah, Dostmohamed, Fellinger, Munkhdalai, Maurya, Garst, Zhang, Krikun, Bucher, Veerubhotla, Liu, Li,
  Gupta, Adamek, Chen, Orlando, Zaks, van Amersfoort, Camp, Wan, Choe, Wu, Olszewska, Yu, Vadali, Scholz, Freitas, Lin, Hua, Liu, Ding, Zhou, Severson, Tsihlas, Yang, Spalink, Yerram, Pankov, Blevins, Vargas, Jauhari, Miecnikowski, Zhang, Kumar, Farabet, Lan, Flennerhag, Bitton, Ma, Bražinskas, Collins, Ahuja, Kudugunta, Bortsova, Giang, Zhu, Chi, Lundberg, Stern, Puttagunta, Xiong, Wu, Pande, Jhindal, Murphy, Clark, Brockschmidt, Deines, McKee, Bahir, Shen, Truong, McDuff, Gesmundo, Rosseel, Liang, Caluwaerts, Hamrick, Kready, Cassin, Ingale, Lao, Pollom, Ding, He, Bellot, Iljazi, Boppana, Han, Thompson, Khalifa, Bulanova, Mitrevski, Pang, Cooney, Shi, Coaguila, Yakar, Ranzato, Momchev, Rawles, Charles, Maeng, Zhang, Bansal, Zhao, Albert, Yuan, Vijayanarasimhan, Hirsch, Ramasesh, Vodrahalli, Wang, Gupta, Strouse, Ni, Patel, Taubman, Huo, Gharibian, Monteiro, Lam, Vasudevan, Chaudhary, Albuquerque, Gupta, Riedel, Hegde, Ruderman, György, Wainwright, Chaugule, Ayan, Levinboim, Shleifer, Kalley, Mirrokni,
  Rao, Radhakrishnan, Hartford, Wu, Zhu, Bertolini, Xiong, Serrano, Tomlinson, Ott, Chang, Graham, Li, Liang, Long, Borgeaud, Ahmad, Grills, Mincu, Izzard, Liu, Xie, O'Bryan, Ponda, Tong, Liu, Malkin, Salama, Chen, Anil, Rao, Swavely, Bilenko, Anderson, Tan, Xie, Wu, Yu, Vinyals, Ryabtsev, Dangovski, Baumli, Keysers, Wright, Ashwood, Chan, Shtefan, Guo, Bapna, Soricut, Pecht, Ramos, Wang, Cai, Trinh, Barham, Friso, Stickgold, Ding, Shakeri, Ardila, Briakou, Culliton, Raveret, Cui, Saxton, Roy, Azizi, Yin, Loher, Bunner, Choi, Ahmed, Li, Li, Dai, Elabd, Ganapathy, Agrawal, Hua, Kunkle, Rajayogam, Ahuja, Conmy, Vasiloff, Beak, Yew, Mudigonda, Wydrowski, Blanton, Wang, Dauphin, Xu, Polacek, Chen, Hu, Sho, Kunesch, Manshadi, Rutherford, Li, Hsiao, Barr, Tudor, Kecman, Nagrani, Pchelin, Sundermeyer, S, Karmarkar, Gao, Chole, Bachem, Gao, BC, Dibb, Verzetti, Hernandez-Campos, Lunts, Johnson, Trapani, Koster, Brusilovsky, Xiong, Mohabey, Ke, Zou, Sabolić, Campos, Palowitch, Morris, Qiu, Ponnuramu, Li, Sharma,
  Sodhia, Tekelioglu, Chuklin, Yenugula, Gemzer, Strinopoulos, El-Husseini, Wang, Zhong, Leurent, Natsev, Wang, Mahaarachchi, Zhu, Peng, Alabed, Lee, Brohan, Szlam, Oh, Kovsharov, Lee, Wong, Barnes, Thornton, Gimeno, Levy, Sevenich, Johnson, Mallinson, Dadashi, Wang, Ren, Lahoti, Dhar, Feldman, Zheng, Ulrich, Panait, Blokzijl, Baetu, Matak, Harlalka, Shah, Marian, von Dincklage, Du, Ley-Wild, Brownfield, Schumacher, Stuken, Noghabi, Gupta, Ren, Malmi, Weissenberger, Huergo, Bauza, Lampe, Douillard, Seyedhosseini, Frostig, Ghahramani, Nguyen, Krishnakumar, Ye, Gupta, Nazari, Geirhos, Shaw, Eleryan, Damen, Palomaki, Xiao, Wu, Yuan, Meadowlark, Bilotti, Lin, Sridhar, Schroecker, Chung, Luo, Strohman, Liu, Zheng, Emond, Wang, Lampinen, Fukuzawa, Campbell-Ajala, Roy, Lee-Thorp, Wang, Naim, Tony, ên, Bensky, Gupta, Rogozińska, Fu, Pillai, Veličković, Drath, Neubeck, Tulsyan, Klimovskiy, Metzler, Stevens, Yeh, Yuan, Yu, Zhang, Go, Tsang, Xu, Wan, Galatzer-Levy, Sobell, Toki, Salesky, Zhou, Antognini, Douglas,
  Wu, Lelkes, Kim, Cavallaro, Salazar, Liu, Besley, Refice, Jia, Li, Sokolik, Kannan, Simon, Chick, Aharon, Gandhi, Daswani, Amiri, Birodkar, Ittycheriah, Grabowski, Chang, Sutton, Zhixin, Lai, Telang, Sargsyan, Jiang, Hoffmann, Brichtova, Hessel, Halcrow, Jerome, Brown, Tomala, Buchatskaya, Yu, Menon, Moreno, Liao, Zayats, Tang, Mah, Shenoy, Siegman, Hadian, Kwon, Tu, Khajehnouri, Foley, Haghani, Wu, Keshava, Gupta, Bruguier, Yao, Karmon, Zintgraf, Wang, Piqueras, Jung, Brennan, Machado, Giustina, Tessler, Lee, Zhang, Moore, Daugaard, Frömmgen, Beattie, Zhang, Kasenberg, Geri, Qin, Tomar, Ouyang, Yu, Zhou, Mathews, Davis, Li, Gupta, Yates, Deng, Kemp, Joung, Vassilvitskii, Guo, LV, Dopson, Lachgar, McConnaughey, Choudhury, Dena, Cohen, Ainslie, Levi, Gopavarapu, Zablotskaia, Vallet, Bahargam, Tang, Tomasev, Dyer, Balle, Lee, Bono, Mendez, Zubov, Yang, Rendulic, Zheng, Hogue, Pundak, Leith, Bhoopchand, Han, Žanić, Schaul, Delakis, Iyer, Wang, Singh, Abdelhamed, Thomas, Brahma, Dib, Kumar, Zhou, Bai,
  Mishra, Sun, Anklin, Sukkerd, Agubuzu, Briukhov, Gulati, Sieb, Pardo, Nasso, Chen, Zhu, Sosea, Goldin, Rush, Hombaiah, Noever, Zhou, Haves, Phuong, Ades, ting Chen, Yang, Pagadora, Bileschi, Cotruta, Saputro, Pramanik, Ammirati, Garrette, Villela, Blyth, Akbulut, Jha, Rrustemi, Wongpanich, Nagpal, Wu, Rivière, Kishchenko, Srinivasan, Chen, Sinha, Pham, Jia, Hennigan, Bakalov, Attaluri, Garmon, Rodriguez, Wegner, Jia, Senter, Fiedel, Petek, Liu, Hardin, Lehri, Carreira, Smoot, Prasetya, Akazawa, Stefanoiu, Ho, Angelova, Lin, Kim, Chen, Sieniek, Li, Guo, Baltateanu, Tafti, Wunder, Olmert, Shukla, Shen, Kovelamudi, Venkatraman, Neel, Thoppilan, Connor, Benzing, Stjerngren, Ghiasi, Polozov, Howland, Weber, Chiu, Girirajan, Terzis, Wang, Li, Shalom, Tewari, Denton, Aharoni, Kalb, Zhao, Zhang, Filos, Rahtz, Jain, Fan, Rodrigues, Wang, Shin, Austin, Ring, Sanchez-Vargas, Hassen, Kessler, Alon, Zhang, Chen, Ma, Si, Hou, Mirhoseini, Wilson, Bacon, Roelofs, Shu, Vasudevan, Adler, Dwornik, Terzi, Lawlor, Askham,
  Bernico, Dong, Hidey, Kilgour, Liu, Bhupatiraju, Leonhard, Zuo, Talukdar, Wei, Severyn, Listík, Lee, Tripathi, Park, Matias, Liu, Ruiz, Jayaram, Tolins, Marcenac, Wang, Seybold, Prior, Sharma, Weber, Sirotenko, Sung, Du, Pavlick, Zinke, Freitag, Dylla, Arenas, Potikha, Goldman, Tao, Chhaparia, Voitovich, Dogra, Ražnatović, Tsai, You, Johnson, Tucker, Gu, Yoo, Majzoubi, Gabeur, Raad, Rhodes, Kolipaka, Howard, Sampemane, Li, Asawaroengchai, Nguyen, Zhang, Cour, Yu, Fu, Jiang, Huang, Surita, Iturrate, Karov, Collins, Baeuml, Fuchs, Shetty, Ramaswamy, Ebrahimi, Guo, Shar, Barth-Maron, Addepalli, Richter, Cheng, Rives, Zheng, Griesser, Dikkala, Zeldes, Safarli, Das, Srivastava, Khan, Li, Pandey, Markeeva, Belov, Yan, Rybiński, Chen, Nawhal, Quinn, Govindaraj, York, Roberts, Garg, Godbole, Abernethy, Das, Thiet, Tompson, Nham, Vats, Caine, Helmholz, Pongetti, Ko, An, Hu, Ling, Pawar, Leland, Kinoshita, Khawaja, Selvi, Ie, Sinopalnikov, Proleev, Tripuraneni, Bevilacqua, Lee, Sanford, Suh, Tran, Dean,
  Baumgartner, Heitkaemper, Gubbi, Toutanova, Xu, Thekkath, Rong, Jain, Xie, Virin, Li, Litchev, Powell, Bharti, Kraft, Hua, Ikonomidis, Hitron, Kumar, Matthey, Bridgers, Lax, Malhi, Skopek, Gupta, Cao, Rasquinha, Põder, Stokowiec, Roth, Li, Sander, Kessinger, Jain, Loper, Park, Yarom, Cheng, Guruganesh, Rao, Li, Barros, Sushkov, Ferng, Shah, Aharoni, Kumar, McConnell, Li, Wang, Pereira, Swanson, Jamil, Xiong, Vijayakumar, Shroff, Soparkar, Gu, Soares, Wang, Majmundar, Wei, Bailey, Kassner, Kawamoto, Žužić, Gomes, Gupta, Guzman, Dasgupta, Bai, Pan, Piccinno, Vogel, Ponce, Hutter, Chang, Jiang, Gog, Ionescu, Manyika, Pedregosa, Ragan, Behrman, Mullins, Devin, Pyne, Gawde, Chadwick, Gu, Tavakkol, Twigg, Goyal, Elue, Goldie, Venkatachary, Fei, Feng, Ritter, Leal, Dasari, Sun, Rochman, O'Donoghue, Liu, Sproch, Chen, Clay, Petrov, Sidhwani, Mihailescu, Panagopoulos, Piergiovanni, Bai, Powell, Karkhanis, Yacovone, Mitrichev, Kovac, Uthus, Yazdanbakhsh, Amos, Zheng, Zhang, Miao, Ramabhadran, Radpour, Thakoor,
  Newlan, Lang, Jankowski, Bharadwaj, Sarr, Ashraf, Mondal, Yan, Rawat, Velury, Kochanski, Eccles, Och, Sharma, Mahintorabi, Gurney, Muir, Cohen, Thakur, Bloniarz, Mujika, Pritzel, Caron, Rahman, Lang, Onoe, Sirkovic, Hoover, Jian, Duque, Narayanan, Soergel, Haig, Maggiore, Buch, Dean, Figotin, Karpov, Gupta, Zhou, Huang, Vaswani, Semturs, Shivakumar, Watanabe, Rajendran, Lu, Hou, Ye, Vashishth, Nti, Sakenas, Ni, DeCarlo, Bendersky, Bagri, Cano, Peake, Tokumine, Godbole, Guía, Lando, Selo, Ellis, Tarlow, Gillick, Epasto, Jonnalagadda, Wei, Xie, Taly, Paganini, Sundararajan, Toyama, Yu, Petrova, Pappu, Agrawal, Buthpitiya, Frye, Buschmann, Crocker, Tagliasacchi, Wang, Huang, Perel, Wieder, Kazawa, Wang, Cole, Gupta, Golan, Bang, Kulkarni, Franko, Liu, Reid, Dalmia, Whang, Cen, Sundaram, Ferret, Isik, Ionita, Sun, Shekhawat, Mohammad, Pham, Huang, Raman, Zhou, Mcilroy, Myers, Peng, Scott, Covington, Erell, Joshi, Oliveira, Noy, Nasir, Walker, Axelrod, Dozat, Han, Chu, Weinstein, Shukla, Chandrakaladharan,
  Poklukar, Li, Jin, Eruvbetine, Hansen, Dabush, Jacovi, Phatale, Zhu, Baker, Shomrat, Xiao, Pouget-Abadie, Zhang, Wei, Song, King, Huang, Zhu, Sun, Franco, Lin, Arora, Hui, Li, Xia, Vilnis, Schain, Alarakyia, Prince, Phillips, Habtegebriel, Xu, Gui, Ontanon, Aroyo, Gill, Lu, Katariya, Madeka, Krishnan, Raghvendra, Freedman, Tay, Menghani, Choy, Shetty, Abolafia, Kukliansky, Chou, Lichtarge, Burke, Coleman, Guo, Jin, Bhattacharya, Langston, Li, Kotecha, Yakubovich, Chen, Petrov, Powell, He, Quick, Garg, Hwang, Lu, Bhojanapalli, Kjems, Mehran, Archer, van Hasselt, Balakrishna, Kearns, Guo, Riesa, Sazanovich, Gao, Sauer, Yang, Sheng, Jimma, Gansbeke, Nikolaev, Wei, Millican, Zhao, Snyder, Bolelli, O'Brien, Xu, Xia, Yuan, Neelakantan, Barker, Yadav, Kirkwood, Ahmad, Wee, Grimstad, Wang, Wiethoff, Settle, Wang, Blundell, Chen, Duvarney, Hu, Ronneberger, Lee, Li, Chakladar, Butryna, Evangelopoulos, Desjardins, Kanerva, Wang, Nowak, Li, Loo, Khurshudov, Shafey, Baddi, Lenc, Razeghi, Lieber, Sinha, Ma, Su, Huang,
  Ushio, Klimczak-Plucińska, Mohamed, Chen, Osindero, Ginzburg, Lamprou, Bashlovkina, Tran, Khodaei, Anand, Di, Eskander, Vuyyuru, Liu, Kamath, Goldenberg, Bellaiche, Pluto, Rosgen, Mansoor, Wong, Ganesh, Bailey, Baird, Deutsch, Baek, Jia, Lee, Friesen, Braun, Lee, Panda, Hernandez, Williams, Liu, Liang, Autef, Pitler, Jain, Kirk, Bunyan, Elias, Yin, Reid, Pope, Putikhin, Samanta, Guadarrama, Kim, Rowe, Valentine, Yan, Salcianu, Silver, Song, Singh, Ye, DeBalsi, Merey, Ofek, Webson, Mourad, Kakarla, Lattanzi, Roy, Sluzhaev, Butterfield, Tonioni, Waters, Kopalle, Chase, Cohan, Rao, Berry, Voznesensky, Hu, Chiafullo, Chikkerur, Scrivener, Zheng, Wiesner, Macherey, Lillicrap, Liu, Walker, Welling, Davies, Huang, Ren, Shabat, Agostini, Iinuma, Zelle, Sathyanarayana, D'olimpio, Redshaw, Ginsberg, Murthy, Geller, Matejovicova, Chakrabarti, Julian, Chan, Hu, Jarrett, Agarwal, Challagundla, Li, Tata, Ding, Meng, Dai, Vezzani, Garg, Bulian, Jasarevic, Cai, Rajamani, Santoro, Hartmann, Liang, Perz, Jindal, Bu, Seo,
  Poplin, Goedeckemeyer, Ghazi, Khadke, Liu, Mather, Zhang, Shah, Chen, Wei, Shivam, Cao, Cho, Scarpati, Moffitt, Barbu, Jurin, Chang, Liu, Zheng, Dave, Kaeser-Chen, Yu, Abdagic, Gonzalez, Huang, Zhong, Schmid, Petrini, Wertheim, Zhu, Nguyen, Ji, Zhou, Zhou, Feng, Cohen, Rim, Phal, Georgiev, Brand, Ma, Li, Gupta, Wang, Dubov, Tarbouriech, Majumder, Li, Rink, Suman, Guo, Sun, Nair, Xu, Elhawaty, Cabrera, Han, Eisenschlos, Bai, Li, Bansal, Sellam, Khan, Nguyen, Mao-Jones, Parotsidis, Marcus, Fan, Zimmermann, Kochinski, Graesser, Behbahani, Caceres, Riley, Kane, Lefdal, Willoughby, Vicol, Wang, Zhang, Gill, Liang, Prasad, Mariooryad, Kazemi, Wang, Muralidharan, Voigtlaender, Zhao, Zhou, D'Souza, Mavalankar, Arnold, Young, Sarvana, Lee, Nasr, Zou, Kim, Haas, Patel, Bulut, Parkinson, Biles, Kalashnikov, To, Kumar, Austin, Greve, Zhang, Goel, Li, Yaroshenko, Chang, Jindal, Clark, Taitelbaum, Johnson, Roval, Ko, Mohananey, Schuler, Dodhia, Li, Osawa, Cui, Xu, Shah, Huang, Gruzewska, Clement, Verma, Sercinoglu, Qian,
  Shah, Yamaguchi, Modi, Kosakai, Strohmann, Zeng, Gunel, Qian, Tarango, Jastrzębski, David, Shan, Schuh, Lad, Gierke, Madhavan, Chen, Kurzeja, Santamaria-Fernandez, Chen, Cordell, Chervonyi, Garcia, Kannen, Perot, Ding, Cohen-Ganor, Lavrenko, Wu, Evans, dos Santos, Sewak, Brown, Hard, Puigcerver, Zheng, Liang, Gladchenko, Ingle, First, Sermanet, Magister, Velimirović, Reddi, Ricco, Agustsson, Adam, Levine, Gaddy, Holtmann-Rice, Wang, Sathe, Roy, Bratanič, Carin, Mehta, Bonacina, Cao, Finkelstein, Rieser, Wu, Altché, Scandinaro, Li, Vieillard, Sethi, Tanzer, Xing, Wang, Bhatia, Citovsky, Anthony, Lin, Shi, Jakobovits, Gibson, Apte, Lee, Chen, Byravan, Maniatis, Webster, Dai, Chen, Pan, Fadeeva, Gleicher, Luong, and Bhumihar}]{comanici2025gemini25pushingfrontier}
Gheorghe Comanici, Eric Bieber, Mike Schaekermann, Ice Pasupat, Noveen Sachdeva, Inderjit Dhillon, Marcel Blistein, Ori Ram, Dan Zhang, Evan Rosen, Luke Marris, Sam Petulla, Colin Gaffney, Asaf Aharoni, Nathan Lintz, Tiago~Cardal Pais, Henrik Jacobsson, Idan Szpektor, Nan-Jiang Jiang, and 3290 others. 2025.
\newblock \href {https://arxiv.org/abs/2507.06261} {Gemini 2.5: Pushing the frontier with advanced reasoning, multimodality, long context, and next generation agentic capabilities}.
\newblock \emph{Preprint}, arXiv:2507.06261.

\bibitem[{Dang et~al.(2024)Dang, Singh, D'souza, Ahmadian, Salamanca, Smith, Peppin, Hong, Govindassamy, Zhao, Kublik, Amer, Aryabumi, Campos, Tan, Kocmi, Strub, Grinsztajn, Flet-Berliac, Locatelli, Lin, Talupuru, Venkitesh, Cairuz, Yang, Chung, Ko, Shi, Shukayev, Bae, Piktus, Castagné, Cruz-Salinas, Kim, Crawhall-Stein, Morisot, Roy, Blunsom, Zhang, Gomez, Frosst, Fadaee, Ermis, Üstün, and Hooker}]{dang2024}
John Dang, Shivalika Singh, Daniel D'souza, Arash Ahmadian, Alejandro Salamanca, Madeline Smith, Aidan Peppin, Sungjin Hong, Manoj Govindassamy, Terrence Zhao, Sandra Kublik, Meor Amer, Viraat Aryabumi, Jon~Ander Campos, Yi-Chern Tan, Tom Kocmi, Florian Strub, Nathan Grinsztajn, Yannis Flet-Berliac, and 26 others. 2024.
\newblock \href {https://arxiv.org/abs/2412.04261} {Aya expanse: Combining research breakthroughs for a new multilingual frontier}.
\newblock \emph{Preprint}, arXiv:2412.04261.

\bibitem[{Gao et~al.(2024)Gao, Tow, Abbasi, Biderman, Black, DiPofi, Foster, Golding, Hsu, Le~Noac'h, Li, McDonell, Muennighoff, Ociepa, Phang, Reynolds, Schoelkopf, Skowron, Sutawika, Tang, Thite, Wang, Wang, and Zou}]{eval-harness}
Leo Gao, Jonathan Tow, Baber Abbasi, Stella Biderman, Sid Black, Anthony DiPofi, Charles Foster, Laurence Golding, Jeffrey Hsu, Alain Le~Noac'h, Haonan Li, Kyle McDonell, Niklas Muennighoff, Chris Ociepa, Jason Phang, Laria Reynolds, Hailey Schoelkopf, Aviya Skowron, Lintang Sutawika, and 5 others. 2024.
\newblock \href {https://doi.org/10.5281/zenodo.12608602} {The language model evaluation harness}.

\bibitem[{Ghaboura et~al.(2025)Ghaboura, Heakl, Thawakar, Alharthi, Riahi, Radman, Laaksonen, Khan, Khan, and Anwer}]{ghaboura2024camelbenchcomprehensivearabiclmm}
Sara Ghaboura, Ahmed Heakl, Omkar Thawakar, Ali Husain Salem~Abdulla Alharthi, Ines Riahi, Abduljalil Radman, Jorma Laaksonen, Fahad~Shahbaz Khan, Salman Khan, and Rao~Muhammad Anwer. 2025.
\newblock \href {https://doi.org/10.18653/v1/2025.findings-naacl.105} {{CAMEL}-bench: A comprehensive {A}rabic {LMM} benchmark}.
\newblock In \emph{Findings of the Association for Computational Linguistics: NAACL 2025}, pages 1970--1980, Albuquerque, New Mexico. Association for Computational Linguistics.

\bibitem[{Grattafiori et~al.(2024)Grattafiori, Dubey, Jauhri, Pandey, Kadian, Al-Dahle, Letman, Mathur, Schelten, Vaughan, Yang, Fan, Goyal, Hartshorn, Yang, Mitra, Sravankumar, Korenev, Hinsvark, Rao, Zhang, Rodriguez, Gregerson, Spataru, Roziere, Biron, Tang, Chern, Caucheteux, Nayak, Bi, Marra, McConnell, Keller, Touret, Wu, Wong, Ferrer, Nikolaidis, Allonsius, Song, Pintz, Livshits, Wyatt, Esiobu, Choudhary, Mahajan, Garcia-Olano, Perino, Hupkes, Lakomkin, AlBadawy, Lobanova, Dinan, Smith, Radenovic, Guzmán, Zhang, Synnaeve, Lee, Anderson, Thattai, Nail, Mialon, Pang, Cucurell, Nguyen, Korevaar, Xu, Touvron, Zarov, Ibarra, Kloumann, Misra, Evtimov, Zhang, Copet, Lee, Geffert, Vranes, Park, Mahadeokar, Shah, van~der Linde, Billock, Hong, Lee, Fu, Chi, Huang, Liu, Wang, Yu, Bitton, Spisak, Park, Rocca, Johnstun, Saxe, Jia, Alwala, Prasad, Upasani, Plawiak, Li, Heafield, Stone, El-Arini, Iyer, Malik, Chiu, Bhalla, Lakhotia, Rantala-Yeary, van~der Maaten, Chen, Tan, Jenkins, Martin, Madaan, Malo, Blecher,
  Landzaat, de~Oliveira, Muzzi, Pasupuleti, Singh, Paluri, Kardas, Tsimpoukelli, Oldham, Rita, Pavlova, Kambadur, Lewis, Si, Singh, Hassan, Goyal, Torabi, Bashlykov, Bogoychev, Chatterji, Zhang, Duchenne, Çelebi, Alrassy, Zhang, Li, Vasic, Weng, Bhargava, Dubal, Krishnan, Koura, Xu, He, Dong, Srinivasan, Ganapathy, Calderer, Cabral, Stojnic, Raileanu, Maheswari, Girdhar, Patel, Sauvestre, Polidoro, Sumbaly, Taylor, Silva, Hou, Wang, Hosseini, Chennabasappa, Singh, Bell, Kim, Edunov, Nie, Narang, Raparthy, Shen, Wan, Bhosale, Zhang, Vandenhende, Batra, Whitman, Sootla, Collot, Gururangan, Borodinsky, Herman, Fowler, Sheasha, Georgiou, Scialom, Speckbacher, Mihaylov, Xiao, Karn, Goswami, Gupta, Ramanathan, Kerkez, Gonguet, Do, Vogeti, Albiero, Petrovic, Chu, Xiong, Fu, Meers, Martinet, Wang, Wang, Tan, Xia, Xie, Jia, Wang, Goldschlag, Gaur, Babaei, Wen, Song, Zhang, Li, Mao, Coudert, Yan, Chen, Papakipos, Singh, Srivastava, Jain, Kelsey, Shajnfeld, Gangidi, Victoria, Goldstand, Menon, Sharma, Boesenberg,
  Baevski, Feinstein, Kallet, Sangani, Teo, Yunus, Lupu, Alvarado, Caples, Gu, Ho, Poulton, Ryan, Ramchandani, Dong, Franco, Goyal, Saraf, Chowdhury, Gabriel, Bharambe, Eisenman, Yazdan, James, Maurer, Leonhardi, Huang, Loyd, Paola, Paranjape, Liu, Wu, Ni, Hancock, Wasti, Spence, Stojkovic, Gamido, Montalvo, Parker, Burton, Mejia, Liu, Wang, Kim, Zhou, Hu, Chu, Cai, Tindal, Feichtenhofer, Gao, Civin, Beaty, Kreymer, Li, Adkins, Xu, Testuggine, David, Parikh, Liskovich, Foss, Wang, Le, Holland, Dowling, Jamil, Montgomery, Presani, Hahn, Wood, Le, Brinkman, Arcaute, Dunbar, Smothers, Sun, Kreuk, Tian, Kokkinos, Ozgenel, Caggioni, Kanayet, Seide, Florez, Schwarz, Badeer, Swee, Halpern, Herman, Sizov, Guangyi, Zhang, Lakshminarayanan, Inan, Shojanazeri, Zou, Wang, Zha, Habeeb, Rudolph, Suk, Aspegren, Goldman, Zhan, Damlaj, Molybog, Tufanov, Leontiadis, Veliche, Gat, Weissman, Geboski, Kohli, Lam, Asher, Gaya, Marcus, Tang, Chan, Zhen, Reizenstein, Teboul, Zhong, Jin, Yang, Cummings, Carvill, Shepard, McPhie,
  Torres, Ginsburg, Wang, Wu, U, Saxena, Khandelwal, Zand, Matosich, Veeraraghavan, Michelena, Li, Jagadeesh, Huang, Chawla, Huang, Chen, Garg, A, Silva, Bell, Zhang, Guo, Yu, Moshkovich, Wehrstedt, Khabsa, Avalani, Bhatt, Mankus, Hasson, Lennie, Reso, Groshev, Naumov, Lathi, Keneally, Liu, Seltzer, Valko, Restrepo, Patel, Vyatskov, Samvelyan, Clark, Macey, Wang, Hermoso, Metanat, Rastegari, Bansal, Santhanam, Parks, White, Bawa, Singhal, Egebo, Usunier, Mehta, Laptev, Dong, Cheng, Chernoguz, Hart, Salpekar, Kalinli, Kent, Parekh, Saab, Balaji, Rittner, Bontrager, Roux, Dollar, Zvyagina, Ratanchandani, Yuvraj, Liang, Alao, Rodriguez, Ayub, Murthy, Nayani, Mitra, Parthasarathy, Li, Hogan, Battey, Wang, Howes, Rinott, Mehta, Siby, Bondu, Datta, Chugh, Hunt, Dhillon, Sidorov, Pan, Mahajan, Verma, Yamamoto, Ramaswamy, Lindsay, Lindsay, Feng, Lin, Zha, Patil, Shankar, Zhang, Zhang, Wang, Agarwal, Sajuyigbe, Chintala, Max, Chen, Kehoe, Satterfield, Govindaprasad, Gupta, Deng, Cho, Virk, Subramanian, Choudhury,
  Goldman, Remez, Glaser, Best, Koehler, Robinson, Li, Zhang, Matthews, Chou, Shaked, Vontimitta, Ajayi, Montanez, Mohan, Kumar, Mangla, Ionescu, Poenaru, Mihailescu, Ivanov, Li, Wang, Jiang, Bouaziz, Constable, Tang, Wu, Wang, Wu, Gao, Kleinman, Chen, Hu, Jia, Qi, Li, Zhang, Zhang, Adi, Nam, Yu, Wang, Zhao, Hao, Qian, Li, He, Rait, DeVito, Rosnbrick, Wen, Yang, Zhao, and Ma}]{grattafiori2024llama3herdmodels}
Aaron Grattafiori, Abhimanyu Dubey, Abhinav Jauhri, Abhinav Pandey, Abhishek Kadian, Ahmad Al-Dahle, Aiesha Letman, Akhil Mathur, Alan Schelten, Alex Vaughan, Amy Yang, Angela Fan, Anirudh Goyal, Anthony Hartshorn, Aobo Yang, Archi Mitra, Archie Sravankumar, Artem Korenev, Arthur Hinsvark, and 542 others. 2024.
\newblock \href {https://arxiv.org/abs/2407.21783} {The llama 3 herd of models}.
\newblock \emph{Preprint}, arXiv:2407.21783.

\bibitem[{He et~al.(2022)He, Cheng, Li, Xie, and Xiao}]{he-etal-2022-pre}
Qianyu He, Sijie Cheng, Zhixu Li, Rui Xie, and Yanghua Xiao. 2022.
\newblock \href {https://doi.org/10.18653/v1/2022.acl-long.543} {Can pre-trained language models interpret similes as smart as human?}
\newblock In \emph{Proceedings of the 60th Annual Meeting of the Association for Computational Linguistics (Volume 1: Long Papers)}, pages 7875--7887, Dublin, Ireland. Association for Computational Linguistics.

\bibitem[{Huang et~al.(2024)Huang, Yu, Zhu, Sun, Cheng, Dingjie, Chen, Alharthi, An, He, Liu, Chen, Li, Wang, Zhang, Sun, Wan, Li, and Xu}]{acegpt}
Huang Huang, Fei Yu, Jianqing Zhu, Xuening Sun, Hao Cheng, Song Dingjie, Zhihong Chen, Mosen Alharthi, Bang An, Juncai He, Ziche Liu, Junying Chen, Jianquan Li, Benyou Wang, Lian Zhang, Ruoyu Sun, Xiang Wan, Haizhou Li, and Jinchao Xu. 2024.
\newblock \href {https://doi.org/10.18653/v1/2024.naacl-long.450} {{A}ce{GPT}, localizing large language models in {A}rabic}.
\newblock In \emph{Proceedings of the 2024 Conference of the North American Chapter of the Association for Computational Linguistics: Human Language Technologies (Volume 1: Long Papers)}, pages 8139--8163, Mexico City, Mexico. Association for Computational Linguistics.

\bibitem[{Jang et~al.(2023)Jang, Yu, and Frassinelli}]{jang-etal-2023-figurative}
Hyewon Jang, Qi~Yu, and Diego Frassinelli. 2023.
\newblock \href {https://doi.org/10.18653/v1/2023.findings-acl.622} {Figurative language processing: A linguistically informed feature analysis of the behavior of language models and humans}.
\newblock In \emph{Findings of the Association for Computational Linguistics: ACL 2023}, pages 9816--9832, Toronto, Canada. Association for Computational Linguistics.

\bibitem[{Jiang et~al.(2023)Jiang, Sablayrolles, Mensch, Bamford, Chaplot, de~las Casas, Bressand, Lengyel, Lample, Saulnier, Lavaud, Lachaux, Stock, Scao, Lavril, Wang, Lacroix, and Sayed}]{jiang2023mistral7b}
Albert~Q. Jiang, Alexandre Sablayrolles, Arthur Mensch, Chris Bamford, Devendra~Singh Chaplot, Diego de~las Casas, Florian Bressand, Gianna Lengyel, Guillaume Lample, Lucile Saulnier, Lélio~Renard Lavaud, Marie-Anne Lachaux, Pierre Stock, Teven~Le Scao, Thibaut Lavril, Thomas Wang, Timothée Lacroix, and William~El Sayed. 2023.
\newblock \href {https://arxiv.org/abs/2310.06825} {Mistral 7b}.
\newblock \emph{Preprint}, arXiv:2310.06825.

\bibitem[{Kabra et~al.(2023)Kabra, Liu, Khanuja, Aji, Winata, Cahyawijaya, Aremu, Ogayo, and Neubig}]{kabra-etal-2023-multi}
Anubha Kabra, Emmy Liu, Simran Khanuja, Alham~Fikri Aji, Genta Winata, Samuel Cahyawijaya, Anuoluwapo Aremu, Perez Ogayo, and Graham Neubig. 2023.
\newblock \href {https://doi.org/10.18653/v1/2023.findings-acl.525} {Multi-lingual and multi-cultural figurative language understanding}.
\newblock In \emph{Findings of the Association for Computational Linguistics: ACL 2023}, pages 8269--8284, Toronto, Canada. Association for Computational Linguistics.

\bibitem[{Kwon et~al.(2023)Kwon, Li, Zhuang, Sheng, Zheng, Yu, Gonzalez, Zhang, and Stoica}]{vllm}
Woosuk Kwon, Zhuohan Li, Siyuan Zhuang, Ying Sheng, Lianmin Zheng, Cody~Hao Yu, Joseph Gonzalez, Hao Zhang, and Ion Stoica. 2023.
\newblock \href {https://doi.org/10.1145/3600006.3613165} {Efficient memory management for large language model serving with pagedattention}.
\newblock In \emph{Proceedings of the 29th Symposium on Operating Systems Principles}, SOSP '23, page 611–626, New York, NY, USA. Association for Computing Machinery.

\bibitem[{Lin et~al.(2022)Lin, Mihaylov, Artetxe, Wang, Chen, Simig, Ott, Goyal, Bhosale, Du, Pasunuru, Shleifer, Koura, Chaudhary, O{'}Horo, Wang, Zettlemoyer, Kozareva, Diab, Stoyanov, and Li}]{lin-etal-2022-shot}
Xi~Victoria Lin, Todor Mihaylov, Mikel Artetxe, Tianlu Wang, Shuohui Chen, Daniel Simig, Myle Ott, Naman Goyal, Shruti Bhosale, Jingfei Du, Ramakanth Pasunuru, Sam Shleifer, Punit~Singh Koura, Vishrav Chaudhary, Brian O{'}Horo, Jeff Wang, Luke Zettlemoyer, Zornitsa Kozareva, Mona Diab, and 2 others. 2022.
\newblock \href {https://doi.org/10.18653/v1/2022.emnlp-main.616} {Few-shot learning with multilingual generative language models}.
\newblock In \emph{Proceedings of the 2022 Conference on Empirical Methods in Natural Language Processing}, pages 9019--9052, Abu Dhabi, United Arab Emirates. Association for Computational Linguistics.

\bibitem[{Liu et~al.(2024)Liu, Koto, Baldwin, and Gurevych}]{liu-etal-2024-multilingual}
Chen Liu, Fajri Koto, Timothy Baldwin, and Iryna Gurevych. 2024.
\newblock \href {https://doi.org/10.18653/v1/2024.naacl-long.112} {Are multilingual {LLM}s culturally-diverse reasoners? an investigation into multicultural proverbs and sayings}.
\newblock In \emph{Proceedings of the 2024 Conference of the North American Chapter of the Association for Computational Linguistics: Human Language Technologies (Volume 1: Long Papers)}, pages 2016--2039, Mexico City, Mexico. Association for Computational Linguistics.

\bibitem[{Liu et~al.(2025)Liu, Gurevych, and Korhonen}]{liu-etal-2025-culturally}
Chen~Cecilia Liu, Iryna Gurevych, and Anna Korhonen. 2025.
\newblock \href {https://doi.org/10.1162/tacl_a_00760} {Culturally aware and adapted {NLP}: A taxonomy and a survey of the state of the art}.
\newblock \emph{Transactions of the Association for Computational Linguistics}, 13:652--689.

\bibitem[{Liu et~al.(2022)Liu, Cui, Zheng, and Neubig}]{liu-etal-2022-testing}
Emmy Liu, Chenxuan Cui, Kenneth Zheng, and Graham Neubig. 2022.
\newblock \href {https://doi.org/10.18653/v1/2022.naacl-main.330} {Testing the ability of language models to interpret figurative language}.
\newblock In \emph{Proceedings of the 2022 Conference of the North American Chapter of the Association for Computational Linguistics: Human Language Technologies}, pages 4437--4452, Seattle, United States. Association for Computational Linguistics.

\bibitem[{Magdy et~al.(2025)Magdy, Kwon, Alwajih, Abdelfadil, Shehata, and Abdul-Mageed}]{magdy-etal-2025-jawaher}
Samar~Mohamed Magdy, Sang~Yun Kwon, Fakhraddin Alwajih, Safaa~Taher Abdelfadil, Shady Shehata, and Muhammad Abdul-Mageed. 2025.
\newblock \href {https://aclanthology.org/2025.naacl-long.613/} {{JAWAHER}: A multidialectal dataset of {A}rabic proverbs for {LLM} benchmarking}.
\newblock In \emph{Proceedings of the 2025 Conference of the Nations of the Americas Chapter of the Association for Computational Linguistics: Human Language Technologies (Volume 1: Long Papers)}, pages 12320--12341, Albuquerque, New Mexico. Association for Computational Linguistics.

\bibitem[{Martínez et~al.(2024)Martínez, Molero, González, Conde, Brysbaert, and Reviriego}]{martínez2024usinglargelanguagemodels}
Gonzalo Martínez, Juan~Diego Molero, Sandra González, Javier Conde, Marc Brysbaert, and Pedro Reviriego. 2024.
\newblock \href {https://arxiv.org/abs/2408.16012} {Using large language models to estimate features of multi-word expressions: Concreteness, valence, arousal}.
\newblock \emph{Preprint}, arXiv:2408.16012.

\bibitem[{{MBC}(2013)}]{mbc2013haret}
{MBC}. 2013.
\newblock \href {https://www.youtube.com/watch?v=72VFVRn4T7c} {Mohamed {M}ounir \& {N}ancy {A}jram - \<حارة السقايين> | {A}rab {I}dol}.
\newblock YouTube video, accessed August 8, 2025.

\bibitem[{Mousi et~al.(2025)Mousi, Durrani, Ahmad, Hasan, Hasanain, Kabbani, Dalvi, Chowdhury, and Alam}]{mousi2024aradicebenchmarksdialectalcultural}
Basel Mousi, Nadir Durrani, Fatema Ahmad, Md.~Arid Hasan, Maram Hasanain, Tameem Kabbani, Fahim Dalvi, Shammur~Absar Chowdhury, and Firoj Alam. 2025.
\newblock \href {https://aclanthology.org/2025.coling-main.283/} {{A}ra{D}i{CE}: Benchmarks for dialectal and cultural capabilities in {LLM}s}.
\newblock In \emph{Proceedings of the 31st International Conference on Computational Linguistics}, pages 4186--4218, Abu Dhabi, UAE. Association for Computational Linguistics.

\bibitem[{Muennighoff et~al.(2023)Muennighoff, Wang, Sutawika, Roberts, Biderman, Le~Scao, Bari, Shen, Yong, Schoelkopf, Tang, Radev, Aji, Almubarak, Albanie, Alyafeai, Webson, Raff, and Raffel}]{muennighoff-etal-2023-crosslingual}
Niklas Muennighoff, Thomas Wang, Lintang Sutawika, Adam Roberts, Stella Biderman, Teven Le~Scao, M~Saiful Bari, Sheng Shen, Zheng~Xin Yong, Hailey Schoelkopf, Xiangru Tang, Dragomir Radev, Alham~Fikri Aji, Khalid Almubarak, Samuel Albanie, Zaid Alyafeai, Albert Webson, Edward Raff, and Colin Raffel. 2023.
\newblock \href {https://doi.org/10.18653/v1/2023.acl-long.891} {Crosslingual generalization through multitask finetuning}.
\newblock In \emph{Proceedings of the 61st Annual Meeting of the Association for Computational Linguistics (Volume 1: Long Papers)}, pages 15991--16111, Toronto, Canada. Association for Computational Linguistics.

\bibitem[{OpenAI(2025)}]{openai_gpt41_2025}
OpenAI. 2025.
\newblock \href {https://openai.com/index/gpt-4-1/} {Introducing gpt-4.1 in the api}.
\newblock OpenAI.

\bibitem[{OpenAI et~al.(2024)OpenAI, Achiam, Adler, Agarwal, Ahmad, Akkaya, Aleman, Almeida, Altenschmidt, Altman, Anadkat, Avila, Babuschkin, Balaji, Balcom, Baltescu, Bao, Bavarian, Belgum, Bello, Berdine, Bernadett-Shapiro, Berner, Bogdonoff, Boiko, Boyd, Brakman, Brockman, Brooks, Brundage, Button, Cai, Campbell, Cann, Carey, Carlson, Carmichael, Chan, Chang, Chantzis, Chen, Chen, Chen, Chen, Chen, Chess, Cho, Chu, Chung, Cummings, Currier, Dai, Decareaux, Degry, Deutsch, Deville, Dhar, Dohan, Dowling, Dunning, Ecoffet, Eleti, Eloundou, Farhi, Fedus, Felix, Fishman, Forte, Fulford, Gao, Georges, Gibson, Goel, Gogineni, Goh, Gontijo-Lopes, Gordon, Grafstein, Gray, Greene, Gross, Gu, Guo, Hallacy, Han, Harris, He, Heaton, Heidecke, Hesse, Hickey, Hickey, Hoeschele, Houghton, Hsu, Hu, Hu, Huizinga, Jain, Jain, Jang, Jiang, Jiang, Jin, Jin, Jomoto, Jonn, Jun, Kaftan, Łukasz Kaiser, Kamali, Kanitscheider, Keskar, Khan, Kilpatrick, Kim, Kim, Kim, Kirchner, Kiros, Knight, Kokotajlo, Łukasz Kondraciuk,
  Kondrich, Konstantinidis, Kosic, Krueger, Kuo, Lampe, Lan, Lee, Leike, Leung, Levy, Li, Lim, Lin, Lin, Litwin, Lopez, Lowe, Lue, Makanju, Malfacini, Manning, Markov, Markovski, Martin, Mayer, Mayne, McGrew, McKinney, McLeavey, McMillan, McNeil, Medina, Mehta, Menick, Metz, Mishchenko, Mishkin, Monaco, Morikawa, Mossing, Mu, Murati, Murk, Mély, Nair, Nakano, Nayak, Neelakantan, Ngo, Noh, Ouyang, O'Keefe, Pachocki, Paino, Palermo, Pantuliano, Parascandolo, Parish, Parparita, Passos, Pavlov, Peng, Perelman, de~Avila Belbute~Peres, Petrov, de~Oliveira~Pinto, Michael, Pokorny, Pokrass, Pong, Powell, Power, Power, Proehl, Puri, Radford, Rae, Ramesh, Raymond, Real, Rimbach, Ross, Rotsted, Roussez, Ryder, Saltarelli, Sanders, Santurkar, Sastry, Schmidt, Schnurr, Schulman, Selsam, Sheppard, Sherbakov, Shieh, Shoker, Shyam, Sidor, Sigler, Simens, Sitkin, Slama, Sohl, Sokolowsky, Song, Staudacher, Such, Summers, Sutskever, Tang, Tezak, Thompson, Tillet, Tootoonchian, Tseng, Tuggle, Turley, Tworek, Uribe, Vallone,
  Vijayvergiya, Voss, Wainwright, Wang, Wang, Wang, Ward, Wei, Weinmann, Welihinda, Welinder, Weng, Weng, Wiethoff, Willner, Winter, Wolrich, Wong, Workman, Wu, Wu, Wu, Xiao, Xu, Yoo, Yu, Yuan, Zaremba, Zellers, Zhang, Zhang, Zhao, Zheng, Zhuang, Zhuk, and Zoph}]{openai2024gpt4technicalreport}
OpenAI, Josh Achiam, Steven Adler, Sandhini Agarwal, Lama Ahmad, Ilge Akkaya, Florencia~Leoni Aleman, Diogo Almeida, Janko Altenschmidt, Sam Altman, Shyamal Anadkat, Red Avila, Igor Babuschkin, Suchir Balaji, Valerie Balcom, Paul Baltescu, Haiming Bao, Mohammad Bavarian, Jeff Belgum, and 262 others. 2024.
\newblock \href {https://arxiv.org/abs/2303.08774} {Gpt-4 technical report}.
\newblock \emph{Preprint}, arXiv:2303.08774.

\bibitem[{Pasha(1949)}]{Taymour1949}
Ahmad~Taymour Pasha. 1949.
\newblock \href {https://www.hindawi.org/books/93047907/} {\emph{Al-Kinayat Al-'Amiyya}}, 2016 edition.
\newblock Mu'assasat Hindawi.
\newblock Available in EPUB, PDF, KFX formats.

\bibitem[{Pezeshkpour and Hruschka(2023)}]{pezeshkpour2023largelanguagemodelssensitivity}
Pouya Pezeshkpour and Estevam Hruschka. 2023.
\newblock \href {https://arxiv.org/abs/2308.11483} {Large language models sensitivity to the order of options in multiple-choice questions}.
\newblock \emph{Preprint}, arXiv:2308.11483.

\bibitem[{Prystawski et~al.(2023)Prystawski, Thibodeau, Potts, and Goodman}]{prystawski2023psychologicallyinformedchainofthoughtpromptsmetaphor}
Ben Prystawski, Paul Thibodeau, Christopher Potts, and Noah~D. Goodman. 2023.
\newblock \href {https://arxiv.org/abs/2209.08141} {Psychologically-informed chain-of-thought prompts for metaphor understanding in large language models}.
\newblock \emph{Preprint}, arXiv:2209.08141.

\bibitem[{Qwen et~al.(2025)Qwen, :, Yang, Yang, Zhang, Hui, Zheng, Yu, Li, Liu, Huang, Wei, Lin, Yang, Tu, Zhang, Yang, Yang, Zhou, Lin, Dang, Lu, Bao, Yang, Yu, Li, Xue, Zhang, Zhu, Men, Lin, Li, Tang, Xia, Ren, Ren, Fan, Su, Zhang, Wan, Liu, Cui, Zhang, and Qiu}]{qwen2025qwen25technicalreport}
Qwen, :, An~Yang, Baosong Yang, Beichen Zhang, Binyuan Hui, Bo~Zheng, Bowen Yu, Chengyuan Li, Dayiheng Liu, Fei Huang, Haoran Wei, Huan Lin, Jian Yang, Jianhong Tu, Jianwei Zhang, Jianxin Yang, Jiaxi Yang, Jingren Zhou, and 25 others. 2025.
\newblock \href {https://arxiv.org/abs/2412.15115} {Qwen2.5 technical report}.
\newblock \emph{Preprint}, arXiv:2412.15115.

\bibitem[{Reimers and Gurevych(2019)}]{reimers-2019-sentence-bert}
Nils Reimers and Iryna Gurevych. 2019.
\newblock \href {http://arxiv.org/abs/1908.10084} {Sentence-bert: Sentence embeddings using siamese bert-networks}.
\newblock In \emph{Proceedings of the 2019 Conference on Empirical Methods in Natural Language Processing}. Association for Computational Linguistics.

\bibitem[{Sadallah et~al.(2025)Sadallah, Tonga, Almubarak, Almheiri, Atif, Qwaider, Kadaoui, Shatnawi, Alesh, and Koto}]{sadallah2025commonsensereasoningarabculture}
Abdelrahman Sadallah, Junior~Cedric Tonga, Khalid Almubarak, Saeed Almheiri, Farah Atif, Chatrine Qwaider, Karima Kadaoui, Sara Shatnawi, Yaser Alesh, and Fajri Koto. 2025.
\newblock \href {https://doi.org/10.18653/v1/2025.acl-long.380} {Commonsense reasoning in {A}rab culture}.
\newblock In \emph{Proceedings of the 63rd Annual Meeting of the Association for Computational Linguistics (Volume 1: Long Papers)}, pages 7695--7710, Vienna, Austria. Association for Computational Linguistics.

\bibitem[{Sap et~al.(2022)Sap, Swayamdipta, Vianna, Zhou, Choi, and Smith}]{sap-etal-2022-annotators}
Maarten Sap, Swabha Swayamdipta, Laura Vianna, Xuhui Zhou, Yejin Choi, and Noah~A. Smith. 2022.
\newblock \href {https://doi.org/10.18653/v1/2022.naacl-main.431} {Annotators with attitudes: How annotator beliefs and identities bias toxic language detection}.
\newblock In \emph{Proceedings of the 2022 Conference of the North American Chapter of the Association for Computational Linguistics: Human Language Technologies}, pages 5884--5906, Seattle, United States. Association for Computational Linguistics.

\bibitem[{Sengupta et~al.(2023)Sengupta, Sahu, Jia, Katipomu, Li, Koto, Marshall, Gosal, Liu, Chen, Afzal, Kamboj, Pandit, Pal, Pradhan, Mujahid, Baali, Han, Bsharat, Aji, Shen, Liu, Vassilieva, Hestness, Hock, Feldman, Lee, Jackson, Ren, Nakov, Baldwin, and Xing}]{sengupta2023}
Neha Sengupta, Sunil~Kumar Sahu, Bokang Jia, Satheesh Katipomu, Haonan Li, Fajri Koto, William Marshall, Gurpreet Gosal, Cynthia Liu, Zhiming Chen, Osama~Mohammed Afzal, Samta Kamboj, Onkar Pandit, Rahul Pal, Lalit Pradhan, Zain~Muhammad Mujahid, Massa Baali, Xudong Han, Sondos~Mahmoud Bsharat, and 13 others. 2023.
\newblock \href {https://arxiv.org/abs/2308.16149} {Jais and jais-chat: Arabic-centric foundation and instruction-tuned open generative large language models}.
\newblock \emph{Preprint}, arXiv:2308.16149.

\bibitem[{She et~al.(2023)She, Potts, Bowman, and Geiger}]{she-etal-2023-scone}
Jingyuan~S. She, Christopher Potts, Samuel~R. Bowman, and Atticus Geiger. 2023.
\newblock \href {https://doi.org/10.18653/v1/2023.acl-short.154} {{S}co{N}e: Benchmarking negation reasoning in language models with fine-tuning and in-context learning}.
\newblock In \emph{Proceedings of the 61st Annual Meeting of the Association for Computational Linguistics (Volume 2: Short Papers)}, pages 1803--1821, Toronto, Canada. Association for Computational Linguistics.

\bibitem[{{silma-ai}(2024)}]{silma-9b-2024}
{silma-ai}. 2024.
\newblock Silma 9b instruct v1.0.
\newblock \url{https://huggingface.co/silma-ai/SILMA-9B-Instruct-v1.0}.

\bibitem[{Sravanthi et~al.(2024)Sravanthi, Doshi, Tankala, Murthy, Dabre, and Bhattacharyya}]{sravanthi-etal-2024-pub}
Settaluri Sravanthi, Meet Doshi, Pavan Tankala, Rudra Murthy, Raj Dabre, and Pushpak Bhattacharyya. 2024.
\newblock \href {https://doi.org/10.18653/v1/2024.findings-acl.719} {{PUB}: A pragmatics understanding benchmark for assessing {LLM}s' pragmatics capabilities}.
\newblock In \emph{Findings of the Association for Computational Linguistics: ACL 2024}, pages 12075--12097, Bangkok, Thailand. Association for Computational Linguistics.

\bibitem[{Team et~al.(2025)Team, Abbas, Ahmad, Alam, Altinisik, Asgari, Boshmaf, Boughorbel, Chawla, Chowdhury, Dalvi, Darwish, Durrani, Elfeky, Elmagarmid, Eltabakh, Fatehkia, Fragkopoulos, Hasanain, Hawasly, Husaini, Jung, Lucas, Magdy, Messaoud, Mohamed, Mohiuddin, Mousi, Mubarak, Musleh, Naeem, Ouzzani, Popovic, Sadeghi, Sencar, Shinoy, Sinan, Zhang, Ali, Kheir, Ma, and Ruan}]{fanarteam2025fanararabiccentricmultimodalgenerative}
Fanar Team, Ummar Abbas, Mohammad~Shahmeer Ahmad, Firoj Alam, Enes Altinisik, Ehsannedin Asgari, Yazan Boshmaf, Sabri Boughorbel, Sanjay Chawla, Shammur Chowdhury, Fahim Dalvi, Kareem Darwish, Nadir Durrani, Mohamed Elfeky, Ahmed Elmagarmid, Mohamed Eltabakh, Masoomali Fatehkia, Anastasios Fragkopoulos, Maram Hasanain, and 23 others. 2025.
\newblock \href {https://arxiv.org/abs/2501.13944} {Fanar: An arabic-centric multimodal generative ai platform}.
\newblock \emph{Preprint}, arXiv:2501.13944.

\bibitem[{Team et~al.(2024{\natexlab{a}})Team, Georgiev, Lei, Burnell, Bai, Gulati, Tanzer, Vincent, Pan, Wang, Mariooryad, Ding, Geng, Alcober, Frostig, Omernick, Walker, Paduraru, Sorokin, Tacchetti, Gaffney, Daruki, Sercinoglu, Gleicher, Love, Voigtlaender, Jain, Surita, Mohamed, Blevins, Ahn, Zhu, Kawintiranon, Firat, Gu, Zhang, Rahtz, Faruqui, Clay, Gilmer, Co-Reyes, Penchev, Zhu, Morioka, Hui, Haridasan, Campos, Mahdieh, Guo, Hassan, Kilgour, Vezer, Cheng, de~Liedekerke, Goyal, Barham, Strouse, Noury, Adler, Sundararajan, Vikram, Lepikhin, Paganini, Garcia, Yang, Valter, Trebacz, Vodrahalli, Asawaroengchai, Ring, Kalb, Soares, Brahma, Steiner, Yu, Mentzer, He, Gonzalez, Xu, Kaufman, Shafey, Oh, Hennigan, van~den Driessche, Odoom, Lucic, Roelofs, Lall, Marathe, Chan, Ontanon, He, Teplyashin, Lai, Crone, Damoc, Ho, Riedel, Lenc, Yeh, Chowdhery, Xu, Kazemi, Amid, Petrushkina, Swersky, Khodaei, Chen, Larkin, Pinto, Yan, Badia, Patil, Hansen, Orr, Arnold, Grimstad, Dai, Douglas, Sinha, Yadav, Chen,
  Gribovskaya, Austin, Zhao, Patel, Komarek, Austin, Borgeaud, Friso, Goyal, Caine, Cao, Chung, Lamm, Barth-Maron, Kagohara, Olszewska, Chen, Shivakumar, Agarwal, Godhia, Rajwar, Snaider, Dotiwalla, Liu, Barua, Ungureanu, Zhang, Batsaikhan, Wirth, Qin, Danihelka, Doshi, Chadwick, Chen, Jain, Le, Kar, Gurumurthy, Li, Sang, Liu, Lamprou, Munoz, Lintz, Mehta, Howard, Reynolds, Aroyo, Wang, Blanco, Cassirer, Griffith, Das, Lee, Sygnowski, Fisher, Besley, Powell, Ahmed, Paulus, Reitter, Borsos, Joshi, Pope, Hand, Selo, Jain, Sethi, Goel, Makino, May, Yang, Schalkwyk, Butterfield, Hauth, Goldin, Hawkins, Senter, Brin, Woodman, Ritter, Noland, Giang, Bolina, Lee, Blyth, Mackinnon, Reid, Sarvana, Silver, Chen, Wang, Maggiore, Chang, Attaluri, Thornton, Chiu, Bunyan, Levine, Chung, Eltyshev, Si, Lillicrap, Brady, Aggarwal, Wu, Xu, McIlroy, Badola, Sandhu, Moreira, Stokowiec, Hemsley, Li, Tudor, Shyam, Rahimtoroghi, Haykal, Sprechmann, Zhou, Mincu, Li, Addanki, Krishna, Wu, Frechette, Eyal, Dafoe, Lacey, Whang,
  Avrahami, Zhang, Taropa, Lin, Toyama, Rutherford, Sano, Choe, Tomala, Safranek-Shrader, Kassner, Pajarskas, Harvey, Sechrist, Fortunato, Lyu, Elsayed, Kuang, Lottes, Chu, Jia, Chen, Humphreys, Baumli, Tao, Samuel, dos Santos, Andreassen, Rakićević, Grewe, Kumar, Winkler, Caton, Brock, Dalmia, Sheahan, Barr, Miao, Natsev, Devlin, Behbahani, Prost, Sun, Myaskovsky, Pillai, Hurt, Lazaridou, Xiong, Zheng, Pardo, Li, Horgan, Stanton, Ambar, Xia, Lince, Wang, Mustafa, Webson, Lee, Anil, Wicke, Dozat, Sinha, Piqueras, Dabir, Upadhyay, Boral, Hendricks, Fry, Djolonga, Su, Walker, Labanowski, Huang, Misra, Chen, Skerry-Ryan, Singh, Rijhwani, Yu, Castro-Ros, Changpinyo, Datta, Bagri, Hrafnkelsson, Maggioni, Zheng, Sulsky, Hou, Paine, Yang, Riesa, Rogozinska, Marcus, Badawy, Zhang, Wang, Miller, Greer, Sjos, Nova, Zen, Chaabouni, Rosca, Jiang, Chen, Liu, Sainath, Krikun, Polozov, Lespiau, Newlan, Cankara, Kwak, Xu, Chen, Coenen, Meyer, Tsihlas, Ma, Gottweis, Xing, Gu, Miao, Frank, Cankara, Ganapathy, Dasgupta,
  Hughes-Fitt, Chen, Reid, Rong, Fan, van Amersfoort, Zhuang, Cohen, Gu, Mohananey, Ilic, Tobin, Wieting, Bortsova, Thacker, Wang, Caveness, Chiu, Sezener, Kaskasoli, Baker, Millican, Elhawaty, Aisopos, Lebsack, Byrd, Dai, Jia, Wiethoff, Davoodi, Weston, Yagati, Ahuja, Gao, Pundak, Zhang, Azzam, Sim, Caelles, Keeling, Sharma, Swing, Li, Liu, Bostock, Bansal, Nado, Anand, Lipschultz, Karmarkar, Proleev, Ittycheriah, Yeganeh, Polovets, Faust, Sun, Rrustemi, Li, Shivanna, Liu, Welty, Lebron, Baddepudi, Krause, Parisotto, Soricut, Xu, Bloxwich, Johnson, Neyshabur, Mao-Jones, Wang, Ramasesh, Abbas, Guez, Segal, Nguyen, Svensson, Hou, York, Milan, Bridgers, Gworek, Tagliasacchi, Lee-Thorp, Chang, Guseynov, Hartman, Kwong, Zhao, Kashem, Cole, Miech, Tanburn, Phuong, Pavetic, Cevey, Comanescu, Ives, Yang, Du, Li, Zhang, Iinuma, Hu, Roy, Bijwadia, Zhu, Martins, Saputro, Gergely, Zheng, Jia, Antonoglou, Sadovsky, Gu, Bi, Andreev, Samangooei, Khan, Kocisky, Filos, Kumar, Bishop, Yu, Hodkinson, Mittal, Shah, Moufarek,
  Cheng, Bloniarz, Lee, Pejman, Michel, Spencer, Feinberg, Xiong, Savinov, Smith, Shakeri, Tran, Chesus, Bohnet, Tucker, von Glehn, Muir, Mao, Kazawa, Slone, Soparkar, Shrivastava, Cobon-Kerr, Sharman, Pavagadhi, Araya, Misiunas, Ghelani, Laskin, Barker, Li, Briukhov, Houlsby, Glaese, Lakshminarayanan, Schucher, Tang, Collins, Lim, Feng, Recasens, Lai, Magni, Cao, Siddhant, Ashwood, Orbay, Dehghani, Brennan, He, Xu, Gao, Saroufim, Molloy, Wu, Arnold, Chang, Schrittwieser, Buchatskaya, Radpour, Polacek, Giordano, Bapna, Tokumine, Hellendoorn, Sottiaux, Cogan, Severyn, Saleh, Thakoor, Shefey, Qiao, Gaba, yiin Chang, Swanson, Zhang, Lee, Rubenstein, Song, Kwiatkowski, Koop, Kannan, Kao, Schuh, Stjerngren, Ghiasi, Gibson, Vilnis, Yuan, Ferreira, Kamath, Klimenko, Franko, Xiao, Bhattacharya, Patel, Wang, Morris, Strudel, Sharma, Choy, Hashemi, Landon, Finkelstein, Jhakra, Frye, Barnes, Mauger, Daun, Baatarsukh, Tung, Farhan, Michalewski, Viola, de~Chaumont~Quitry, Lan, Hudson, Wang, Fischer, Zheng, White, Dragan,
  baptiste Alayrac, Ni, Pritzel, Iwanicki, Isard, Bulanova, Zilka, Dyer, Sachan, Srinivasan, Muckenhirn, Cai, Mandhane, Tariq, Rae, Wang, Ayoub, FitzGerald, Zhao, Han, Alberti, Garrette, Krishnakumar, Gimenez, Levskaya, Sohn, Matak, Iturrate, Chang, Xiang, Cao, Ranka, Brown, Hutter, Mirrokni, Chen, Yao, Egyed, Galilee, Liechty, Kallakuri, Palmer, Ghemawat, Liu, Tao, Thornton, Green, Jasarevic, Lin, Cotruta, Tan, Fiedel, Yu, Chi, Neitz, Heitkaemper, Sinha, Zhou, Sun, Kaed, Hulse, Mishra, Georgaki, Kudugunta, Farabet, Shafran, Vlasic, Tsitsulin, Ananthanarayanan, Carin, Su, Sun, V, Carvajal, Broder, Comsa, Repina, Wong, Chen, Hawkins, Filonov, Loher, Hirnschall, Wang, Ye, Burns, Cate, Wright, Piccinini, Zhang, Lin, Gog, Kulizhskaya, Sreevatsa, Song, Cobo, Iyer, Tekur, Garrido, Xiao, Kemp, Zheng, Li, Agarwal, Ngani, Goshvadi, Santamaria-Fernandez, Fica, Chen, Gorgolewski, Sun, Garg, Ye, Eslami, Hua, Simon, Joshi, Kim, Tenney, Potluri, Thiet, Yuan, Luisier, Chronopoulou, Scellato, Srinivasan, Chen, Koverkathu,
  Dalibard, Xu, Saeta, Anderson, Sellam, Fernando, Huot, Jung, Varadarajan, Quinn, Raul, Le, Habalov, Clark, Jalan, Bullard, Singhal, Luong, Wang, Rajayogam, Eisenschlos, Jia, Finchelstein, Yakubovich, Balle, Fink, Agarwal, Li, Dvijotham, Pal, Kang, Konzelmann, Beattie, Dousse, Wu, Crocker, Elkind, Jonnalagadda, Lee, Holtmann-Rice, Kallarackal, Liu, Vnukov, Vats, Invernizzi, Jafari, Zhou, Taylor, Prendki, Wu, Eccles, Liu, Kopparapu, Beaufays, Angermueller, Marzoca, Sarcar, Dib, Stanway, Perbet, Trdin, Sterneck, Khorlin, Li, Wu, Goenka, Madras, Goldshtein, Gierke, Zhou, Liu, Liang, White, Li, Singh, Bahargam, Epstein, Basu, Lao, Ozturel, Crous, Zhai, Lu, Tung, Gaur, Walton, Dixon, Zhang, Globerson, Uy, Bolt, Wiles, Nasr, Shumailov, Selvi, Piccinno, Aguilar, McCarthy, Khalman, Shukla, Galic, Carpenter, Villela, Zhang, Richardson, Martens, Bosnjak, Belle, Seibert, Alnahlawi, McWilliams, Singh, Louis, Ding, Popovici, Simicich, Knight, Mehta, Gupta, Shi, Fatehi, Mitrovic, Grills, Pagadora, Munkhdalai, Petrova,
  Eisenbud, Zhang, Yates, Mittal, Tripuraneni, Assael, Brovelli, Jain, Velimirovic, Akbulut, Mu, Macherey, Kumar, Xu, Qureshi, Comanici, Wiesner, Gong, Ruddock, Bauer, Felt, GP, Arnab, Zelle, Rothfuss, Rosgen, Shenoy, Seybold, Li, Mudigonda, Erdogan, Xia, Simsa, Michi, Yao, Yew, Kan, Caswell, Radebaugh, Elisseeff, Valenzuela, McKinney, Paterson, Cui, Latorre-Chimoto, Kim, Zeng, Durden, Ponnapalli, Sosea, Choquette-Choo, Manyika, Robenek, Vashisht, Pereira, Lam, Velic, Owusu-Afriyie, Lee, Bolukbasi, Parrish, Lu, Park, Venkatraman, Talbert, Rosique, Cheng, Sozanschi, Paszke, Kumar, Austin, Li, Salama, Perz, Kim, Dukkipati, Baryshnikov, Kaplanis, Sheng, Chervonyi, Unlu, de~Las~Casas, Askham, Tunyasuvunakool, Gimeno, Poder, Kwak, Miecnikowski, Mirrokni, Dimitriev, Parisi, Liu, Tsai, Shevlane, Kouridi, Garmon, Goedeckemeyer, Brown, Vijayakumar, Elqursh, Jazayeri, Huang, Carthy, Hoover, Kim, Kumar, Chen, Biles, Bingham, Rosen, Wang, Tan, Engel, Pongetti, de~Cesare, Hwang, Yu, Pullman, Narayanan, Levin, Gopal, Li,
  Aharoni, Trinh, Lo, Casagrande, Vij, Matthey, Ramadhana, Matthews, Carey, Johnson, Goranova, Shah, Ashraf, Dasgupta, Larsen, Wang, Vuyyuru, Jiang, Ijazi, Osawa, Smith, Boppana, Bilal, Koizumi, Xu, Altun, Shabat, Bariach, Korchemniy, Choo, Ronneberger, Iwuanyanwu, Zhao, Soergel, Hsieh, Cai, Iqbal, Sundermeyer, Chen, Bursztein, Malaviya, Biadsy, Shroff, Dhillon, Latkar, Dyer, Forbes, Nicosia, Nikolaev, Greene, Georgiev, Wang, Martin, Sedghi, Zhang, Banzal, Fritz, Rao, Wang, Zhang, Patraucean, Du, Mordatch, Jurin, Liu, Dubey, Mohan, Nowakowski, Ion, Wei, Tojo, Raad, Hudson, Keshava, Agrawal, Ramirez, Wu, Nguyen, Liu, Sewak, Petrini, Choi, Philips, Wang, Bica, Garg, Wilkiewicz, Agrawal, Li, Guo, Xue, Shaik, Leach, Khan, Wiesinger, Jerome, Chakladar, Wang, Ornduff, Abu, Ghaffarkhah, Wainwright, Cortes, Liu, Maynez, Terzis, Samangouei, Mansour, Kępa, Aubet, Algymr, Banica, Weisz, Orban, Senges, Andrejczuk, Geller, Santo, Anklin, Merey, Baeuml, Strohman, Bai, Petrov, Wu, Hassabis, Kavukcuoglu, Dean, and
  Vinyals}]{geminiteam2024gemini15unlockingmultimodal}
Gemini Team, Petko Georgiev, Ving~Ian Lei, Ryan Burnell, Libin Bai, Anmol Gulati, Garrett Tanzer, Damien Vincent, Zhufeng Pan, Shibo Wang, Soroosh Mariooryad, Yifan Ding, Xinyang Geng, Fred Alcober, Roy Frostig, Mark Omernick, Lexi Walker, Cosmin Paduraru, Christina Sorokin, and 1118 others. 2024{\natexlab{a}}.
\newblock \href {https://arxiv.org/abs/2403.05530} {Gemini 1.5: Unlocking multimodal understanding across millions of tokens of context}.
\newblock \emph{Preprint}, arXiv:2403.05530.

\bibitem[{Team et~al.(2024{\natexlab{b}})Team, Riviere, Pathak, Sessa, Hardin, Bhupatiraju, Hussenot, Mesnard, Shahriari, Ramé, Ferret, Liu, Tafti, Friesen, Casbon, Ramos, Kumar, Lan, Jerome, Tsitsulin, Vieillard, Stanczyk, Girgin, Momchev, Hoffman, Thakoor, Grill, Neyshabur, Bachem, Walton, Severyn, Parrish, Ahmad, Hutchison, Abdagic, Carl, Shen, Brock, Coenen, Laforge, Paterson, Bastian, Piot, Wu, Royal, Chen, Kumar, Perry, Welty, Choquette-Choo, Sinopalnikov, Weinberger, Vijaykumar, Rogozińska, Herbison, Bandy, Wang, Noland, Moreira, Senter, Eltyshev, Visin, Rasskin, Wei, Cameron, Martins, Hashemi, Klimczak-Plucińska, Batra, Dhand, Nardini, Mein, Zhou, Svensson, Stanway, Chan, Zhou, Carrasqueira, Iljazi, Becker, Fernandez, van Amersfoort, Gordon, Lipschultz, Newlan, yeong Ji, Mohamed, Badola, Black, Millican, McDonell, Nguyen, Sodhia, Greene, Sjoesund, Usui, Sifre, Heuermann, Lago, McNealus, Soares, Kilpatrick, Dixon, Martins, Reid, Singh, Iverson, Görner, Velloso, Wirth, Davidow, Miller, Rahtz,
  Watson, Risdal, Kazemi, Moynihan, Zhang, Kahng, Park, Rahman, Khatwani, Dao, Bardoliwalla, Devanathan, Dumai, Chauhan, Wahltinez, Botarda, Barnes, Barham, Michel, Jin, Georgiev, Culliton, Kuppala, Comanescu, Merhej, Jana, Rokni, Agarwal, Mullins, Saadat, Carthy, Cogan, Perrin, Arnold, Krause, Dai, Garg, Sheth, Ronstrom, Chan, Jordan, Yu, Eccles, Hennigan, Kocisky, Doshi, Jain, Yadav, Meshram, Dharmadhikari, Barkley, Wei, Ye, Han, Kwon, Xu, Shen, Gong, Wei, Cotruta, Kirk, Rao, Giang, Peran, Warkentin, Collins, Barral, Ghahramani, Hadsell, Sculley, Banks, Dragan, Petrov, Vinyals, Dean, Hassabis, Kavukcuoglu, Farabet, Buchatskaya, Borgeaud, Fiedel, Joulin, Kenealy, Dadashi, and Andreev}]{gemmateam2024gemma2improvingopen}
Gemma Team, Morgane Riviere, Shreya Pathak, Pier~Giuseppe Sessa, Cassidy Hardin, Surya Bhupatiraju, Léonard Hussenot, Thomas Mesnard, Bobak Shahriari, Alexandre Ramé, Johan Ferret, Peter Liu, Pouya Tafti, Abe Friesen, Michelle Casbon, Sabela Ramos, Ravin Kumar, Charline~Le Lan, Sammy Jerome, and 179 others. 2024{\natexlab{b}}.
\newblock \href {https://arxiv.org/abs/2408.00118} {Gemma 2: Improving open language models at a practical size}.
\newblock \emph{Preprint}, arXiv:2408.00118.

\bibitem[{Truong et~al.(2023)Truong, Baldwin, Verspoor, and Cohn}]{truong-etal-2023-language}
Thinh~Hung Truong, Timothy Baldwin, Karin Verspoor, and Trevor Cohn. 2023.
\newblock \href {https://doi.org/10.18653/v1/2023.starsem-1.10} {Language models are not naysayers: an analysis of language models on negation benchmarks}.
\newblock In \emph{Proceedings of the 12th Joint Conference on Lexical and Computational Semantics (*SEM 2023)}, pages 101--114, Toronto, Canada. Association for Computational Linguistics.

\bibitem[{Wataoka et~al.(2024)Wataoka, Takahashi, and Ri}]{wataoka2024selfpreference}
Koki Wataoka, Tsubasa Takahashi, and Ryokan Ri. 2024.
\newblock \href {https://openreview.net/forum?id=tLZZZIgPJX} {Self-preference bias in {LLM}-as-a-judge}.
\newblock In \emph{Neurips Safe Generative AI Workshop 2024}.

\bibitem[{Zhang et~al.(2020)Zhang, Kishore, Wu, Weinberger, and Artzi}]{bertscore}
Tianyi Zhang, Varsha Kishore, Felix Wu, Kilian~Q. Weinberger, and Yoav Artzi. 2020.
\newblock \href {https://arxiv.org/abs/1904.09675} {Bertscore: Evaluating text generation with bert}.
\newblock In \emph{International Conference on Learning Representations}.

\bibitem[{Zheng et~al.(2024)Zheng, Zhou, Meng, Zhou, and Huang}]{zheng2024largelanguagemodelsrobust}
Chujie Zheng, Hao Zhou, Fandong Meng, Jie Zhou, and Minlie Huang. 2024.
\newblock \href {https://arxiv.org/abs/2309.03882} {Large language models are not robust multiple choice selectors}.
\newblock \emph{Preprint}, arXiv:2309.03882.

\bibitem[{Zheng et~al.(2023)Zheng, Chiang, Sheng, Zhuang, Wu, Zhuang, Lin, Li, Li, Xing, Zhang, Gonzalez, and Stoica}]{zheng2023judgingllmasajudgemtbenchchatbot}
Lianmin Zheng, Wei-Lin Chiang, Ying Sheng, Siyuan Zhuang, Zhanghao Wu, Yonghao Zhuang, Zi~Lin, Zhuohan Li, Dacheng Li, Eric~P. Xing, Hao Zhang, Joseph~E. Gonzalez, and Ion Stoica. 2023.
\newblock \href {https://arxiv.org/abs/2306.05685} {Judging llm-as-a-judge with mt-bench and chatbot arena}.
\newblock In \emph{Thirty-seventh Conference on Neural Information Processing Systems Datasets and Benchmarks Track}.

\end{thebibliography}

\appendix
\section{Prompt Templates}\label{app:prompts}

The prompts used across our evaluation tasks are illustrated in Figures~\ref{fig:mcq-prompt}--\ref{fig:connotation-prompt}. Figure~\ref{fig:mcq-prompt} presents the prompt for the MCQ understanding task, while Figure~\ref{fig:contextual-mcq-prompt} shows the version that includes the idiom’s sentence context. Figures~\ref{fig:generate-prompt} and~\ref{fig:generate-prompt-srl} display the prompts used to generate incorrect distractors for the MCQ task, the latter leveraging semantic role labeling. Figure~\ref{fig:complete-proverb-prompt} illustrates the prompt used to complete proverbs by predicting the final word, and Figure~\ref{fig:mcq-incorrect-prompt} depicts the negated version of the MCQ understanding task. The generation-based prompt used to produce incorrect explanations is shown in Figure~\ref{fig:generation-prompt}, with their evaluation guided by the judging prompt in Figure~\ref{fig:judge-proverb-prompt}. Figure~\ref{fig:idiom-prompt} displays the prompt used to generate example sentences containing idioms from Kinayat prior to human post-editing. Finally, Figures~\ref{fig:prag-use} and~\ref{fig:connotation-prompt} show the prompts for the pragmatic use and connotation labeling tasks, respectively.

In all prompt templates that include the word ``proverb'', they refer to items from the Jawaher dataset of proverbs. For the Kinayat dataset of idioms, the word ``proverb'' in the templates was replaced with ``idiom'' to match the dataset content.

\begin{figure}[!htb]
\centering
\begin{tcolorbox}[  before=\FloatBarrier, 
  after=\FloatBarrier, colback=gray!10!white, colframe=black, boxrule=0.7pt, arc=2pt, left=6pt, right=6pt, top=6pt, bottom=6pt, width=0.95\columnwidth,]
You are tasked with selecting the correct explanation for the following proverb.

Choose the correct explanation from the options provided. Only output the letter corresponding to the correct answer and nothing else.  \\

\textbf{Proverb:} \texttt{[PROVERB]} \\

\textbf{Options:} A. \texttt{[OPTION 1]} \\
\hspace*{2.7em} B. \texttt{[OPTION 2]} \\

\textbf{Answer:}
\end{tcolorbox}

\caption{Prompt used for the MCQ understanding task.}
\label{fig:mcq-prompt}
\end{figure}

\begin{figure}[!htb]
\centering
\begin{tcolorbox}[before=\FloatBarrier, after=\FloatBarrier,colback=gray!10!white, colframe=black, boxrule=0.7pt, arc=2pt, left=6pt, right=6pt, top=6pt, bottom=6pt, width=0.95\columnwidth]
You are tasked with selecting the correct explanation for the following idiom, given the idiom in a sentence for context.

Choose the correct explanation from the options provided. Only output the letter corresponding to the correct answer and nothing else. \\

\textbf{Idiom:} \texttt{[IDIOM]} \\

\textbf{Sentence:} \texttt{[SENTENCE]} \\

\textbf{Options:} A. \texttt{[OPTION 1]} \\
\hspace*{2.7em} B. \texttt{[OPTION 2]} \\

\textbf{Answer:}
\end{tcolorbox}
\caption{Prompt used for the Contextual MCQ Idiom Explanation task.}
\label{fig:contextual-mcq-prompt}
\end{figure}

\begin{figure}[!htb]
\centering
\begin{tcolorbox}[before=\FloatBarrier, 
after=\FloatBarrier,colback=gray!10!white, colframe=black, boxrule=0.7pt, arc=2pt, left=6pt, right=6pt, top=6pt, bottom=6pt, width=0.95\columnwidth]
Given the following correct explanation, generate an incorrect explanation that sounds plausible and is not trivially incorrect. Only output the incorrect explanation and nothing else \\

\textbf{Correct explanation:} [correct explanation] \\

\end{tcolorbox}
\caption{Prompt used for the generating incorrect explanantions.}
\label{fig:generate-prompt}
\end{figure}

\begin{figure}[!htb]
\centering
\begin{tcolorbox}[before=\FloatBarrier, 
after=\FloatBarrier, colback=gray!10!white, colframe=black, boxrule=0.7pt, arc=2pt, left=6pt, right=6pt, top=6pt, bottom=6pt, width=0.95\columnwidth]
Your task is to generate an incorrect explanation from the provided correct explanation by following the given steps: 

1- Find the semantic role labels for the sentence. 

2- Change one of the semantic role labels. 

3- Generate an explanation using the new semantic role labels. 

Only output the result in the following JSON format:\\ \texttt{{\{semantic\_role\_labels: ,\\ new\_labels: ,\\ new\_sentence:\}}}

\vspace{1em}
\textbf{Correct explanation:} [correct explanation]
\end{tcolorbox}
\caption{Prompt used for generating incorrect explanations using semantic role labeling.}
\label{fig:generate-prompt-srl}
\end{figure}

\begin{figure}[!htb]
\centering
\begin{tcolorbox}[before=\FloatBarrier, 
  after=\FloatBarrier, colback=gray!10!white, colframe=black, boxrule=0.7pt, arc=2pt, left=6pt, right=6pt, top=6pt, bottom=6pt, width=0.95\columnwidth,]
You are tasked with completing the proverb with the last word. Output the next word only.\\

\textbf{Incomplete Proverb:} [incomplete proverb]

\textbf{Answer:}
\end{tcolorbox}
\caption{Prompt used for last-word proverb completion.}
\label{fig:complete-proverb-prompt}
\end{figure}

\begin{figure}[!htb]
\centering
\begin{tcolorbox}[before=\FloatBarrier, 
after=\FloatBarrier,colback=gray!10!white, colframe=black, boxrule=0.7pt, arc=2pt, left=6pt, right=6pt, top=6pt, bottom=6pt, width=0.95\columnwidth]
You are tasked with selecting the incorrect explanation for the following proverb.

Choose the incorrect explanation from the options provided. Only output the letter corresponding to the incorrect answer and nothing else.  \\

\textbf{Proverb:} \texttt{[PROVERB]} \\

\textbf{Options:} A. \texttt{[OPTION 1]} \\
\hspace*{2.7em} B. \texttt{[OPTION 2]} \\

\textbf{Answer:}
\end{tcolorbox}
\caption{Prompt used for the MCQ negation task.}
\label{fig:mcq-incorrect-prompt}
\end{figure}

\begin{figure}[!htb]
\centering
\begin{tcolorbox}[colback=gray!10!white, colframe=black, boxrule=0.7pt, arc=2pt, left=6pt, right=6pt, top=6pt, bottom=6pt, width=0.95\columnwidth]

Your task is to explain the meaning of the following Arabic proverb. Provide a clear and concise explanation in Arabic, highlighting its figurative meaning and any cultural or contextual significance.\\
Only output the Arabic explanation and nothing else.\\

\textbf{Proverb:} \texttt{[PROVERB]}

\textbf{Arabic Explanation:}
\end{tcolorbox}
\caption{Prompt used for the task of generating explanations.}
\label{fig:generation-prompt}
\end{figure}

\begin{figure}[!htb]
\centering
\begin{tcolorbox}[before=\FloatBarrier, 
after=\FloatBarrier,colback=gray!10!white, colframe=black, boxrule=0.7pt, arc=2pt, left=6pt, right=6pt, top=6pt, bottom=6pt, width=0.95\columnwidth]

You are an expert in Arabic language and culture. Your task is to evaluate how well a generated explanation matches the intended meaning of the reference explanation of an Arabic idiom/proverb.
Below is a reference (gold) explanation and a generated explanation.\\

Rate the accuracy of the generated explanation based on how well it preserves the intended meaning of the idiom/proverb.\\

\textbf{Gold Explanation:} \{gold\_explanation\}\\
\textbf{Generated Explanation:} \{generated\_explanation\}\\

Use the following rating scale:

5 = \textit{Excellent:} Perfectly matches the gold explanation in meaning.

4 = \textit{Good:} Minor omissions or phrasing differences, but the meaning is well preserved.

3 = \textit{Fair:} Partial understanding, some inaccuracies or missing key aspects.

2 = \textit{Poor:} Significant misunderstanding or loss of core meaning.

1 = Very Poor: Completely incorrect or irrelevant explanation.\\

Only output a numerical rating and nothing else.

\textbf{Rating (1–5):}

\end{tcolorbox}
\caption{Prompt used for LLM-as-a-judge rating of idiom/proverb explanation quality.}
\label{fig:judge-proverb-prompt}
\end{figure}

\begin{figure}[!htb]
\centering
\begin{tcolorbox}[before=\FloatBarrier, 
after=\FloatBarrier,colback=gray!10!white, colframe=black, boxrule=0.7pt, arc=2pt, width=0.95\columnwidth, left=6pt, right=6pt, top=6pt, bottom=6pt]
Generate a sample sentence using the following idiom in the correct context in the Egyptian Arabic dialect given the following explanation of the idiom.\\

\textbf{Idiom:} \texttt{\{idiom\}}\\
\textbf{Explanation:} \texttt{\{explanation\}}\\

Only output the sentence and nothing else.
\end{tcolorbox}
\caption{Prompt for generating an Egyptian Arabic sentence using a given idiom in context.}
\label{fig:idiom-prompt}
\end{figure}

\begin{figure}[!htbp]
\centering
\begin{tcolorbox}[before=\FloatBarrier, 
after=\FloatBarrier,colback=gray!10!white, colframe=black, boxrule=0.7pt, arc=2pt, left=6pt, right=6pt, top=6pt, bottom=6pt, width=0.95\columnwidth]
Your task is to fill in the blank with the correct idiom.\\
Choose the correct idiom from the options provided. Only output the letter corresponding to the correct answer and nothing else.\\

\textbf{Sentence:} \texttt{[sentence with blank]} \\

\textbf{Options:} A. \texttt{[OPTION 1]} \\
\hspace*{2.7em} B. \texttt{[OPTION 2]} \\

\textbf{Answer:}
\end{tcolorbox}
\caption{Prompt used for the idiom pragmatic use task.}
\label{fig:prag-use}
\end{figure}

\begin{figure}[!htb]
\centering
\begin{tcolorbox}[before=\FloatBarrier, 
after=\FloatBarrier,colback=gray!10!white, colframe=black, boxrule=0.7pt, arc=2pt, left=6pt, right=6pt, top=6pt, bottom=6pt, width=0.95\columnwidth]
Determine the connotation of the following Arabic proverb or explanation. Classify the connotation as Positive, Negative, or Neutral based on the following guidelines:\\

\textbf{Positive Connotation}: It conveys optimism, hope, praise, or beneficial outcomes. It highlights virtues such as kindness, success, loyalty, or happiness. It encourages or celebrates desirable behaviors or outcomes.\\

\textbf{Negative Connotation}: It expresses pessimism, caution, loss, or undesirable consequences. It highlights flaws, mistakes, or risks and often reflects on the dangers or negative results of certain actions.\\

\textbf{Neutral Connotation}: It provides general advice or observation without invoking strong feelings or judgment.\\

\textbf{Proverb/Explanation:} \texttt{[PROVERB OR EXPLANATION]} \\

Only output the connotation and nothing else.

\textbf{Connotation:}
\end{tcolorbox}
\caption{Prompt used for connotation classification of Arabic proverbs and their explanations.}
\label{fig:connotation-prompt}
\end{figure}

\clearpage

\section{Verification of Incorrect Explanations}
\label{app:verify-incorrect}

Figure~\ref{fig:verify-explanations} shows the annotation guidelines for verifying the quality of the generated incorrect explanations used as distractors in the MCQ understanding task, while Table~\ref{tab:llm_judge_examples}  provides examples of generated explanations that received scores of 2, 1, and 0, along with the rationale for each score.

\begin{figure}[!htb]
\centering
\begin{tcolorbox}[colback=gray!10!white, colframe=black, boxrule=0.7pt, arc=2pt, left=6pt, right=6pt, top=6pt, bottom=6pt, width=\columnwidth]

\paragraph{Task Overview}
A single annotator evaluated the quality of automatically generated \textit{incorrect explanations} for idioms and proverbs. The goal was to determine whether each explanation was both \textit{plausible} and \textit{incorrect}, ensuring that it functioned as a meaningful distractor rather than a trivial or nonsensical option.\\

\paragraph{Inputs}
For each sample, the annotator was provided with: the idioms/proverbs, the correct explanations (true meaning), and the corresponding generated incorrect explanations.\\

\paragraph{Annotation Guidelines}
Each incorrect explanation was rated on a \textbf{0--2 scale} based on its plausibility, distinctness from the correct explanation, and linguistic coherence.\\

Use the following rating scale:

2 = \textit{High-Quality Incorrect (Plausible but Wrong):} The explanation is clearly incorrect yet plausible. It makes sense linguistically and culturally, could realistically confuse a reader, and differs semantically from the correct meaning.

1 = \textit{Medium-Quality Incorrect (Too Similar or Slightly Off):} The explanation is partially incorrect, implausible or too close in meaning to the correct one (e.g., paraphrase or mild variation). It shows partial understanding but fails to be a strong distractor.

0 = \textit{Low-Quality Incorrect (Implausible or Unrelated):} The explanation is either correct, or clearly irrelevant, nonsensical, or incomplete. It fails to make sense in context and would not plausibly be mistaken for the correct meaning. \\

\end{tcolorbox}
\caption{Guidelines for human verification of incorrect explanations generated by GPT-4.1.}
\label{fig:verify-explanations}
\end{figure}

\begin{table*}[t]
\centering
\small
\begin{tabular}{p{1.8cm} | p{4.2cm} | p{4.2cm} | c | p{3.4cm}}
\hline
\textbf{Idiom} & \textbf{Correct Explanation} & \textbf{Incorrect Generation} & \textbf{Score} & \textbf{Rationale} \\
\hline
\RL{جَسِّ الْمَخَاضَهْ} 
& \RL{كناية عن استطلاع الأمر، وهي في معنى سَبْرِ الغَوْر.}
& \RL{كناية عن إدراك التفاصيل الدقيقة، وهي في معنى تتبع الأثر.}
& 2 
& The generation is incorrect but remains semantically plausible. \\
\hline
\RL{سَيِّبُهْ يِرِنْ} 
& \RL{ترك الدار خالية تصفر، هذا الأصل في الكناية، ثم كنوا بها عن ترك شخص وشأنه فيما يعمل أو يتكلم به، وإهماله منفردًا يناجي نفسه في وحدته.}
& \RL{تُرِكَت الدار خالية تصفر، أي أنها امتلأت بالناس وازدحمت بالأصوات، ثم صار يُكنى بذلك عن مواكبة الآخرين للمرء في أفعاله وأقواله ومشاركته في كل شأن من شؤونه.}
& 1 
& The generation is incorrect and weakly plausible, as it contradicts the literal meaning by interpreting ``empty'' as ``full.'' \\
\hline
\RL{جَابْهَا فِي قُبِّتُهْ} 
& \RL{أي علَّقها أو ربطها في قبِّ قميص ذلك الشخص، أي طوَّق بها عنقه وألصقها به، والمراد التهمة يتحايل بعضهم حتى يلصقها بشخص.}
& \RL{أي وضعها في جيب قميصه ليحفظها من الضياع، أي أبقاها بالقرب من صدره كعلامة على البراءة والثقة، والمقصود أن التهمة يبعدها البعض عن أنفسهم كي لا تلتصق بهم.}
& 0 
& The generation is implausible and semantically incoherent with the idiom’s intended meaning. \\
\hline
\end{tabular}
\caption{Examples of model-generated explanations receiving scores of 2, 1, and 0, along with rationales. These examples illustrate the annotation guidelines used for explanation verification (see Figure~16).}
\label{tab:llm_judge_examples}
\end{table*}

\clearpage 
\section{Data Analysis}\label{app:analysis}

\begin{figure}[htbp]
    \centering
    \includegraphics[width=\columnwidth]{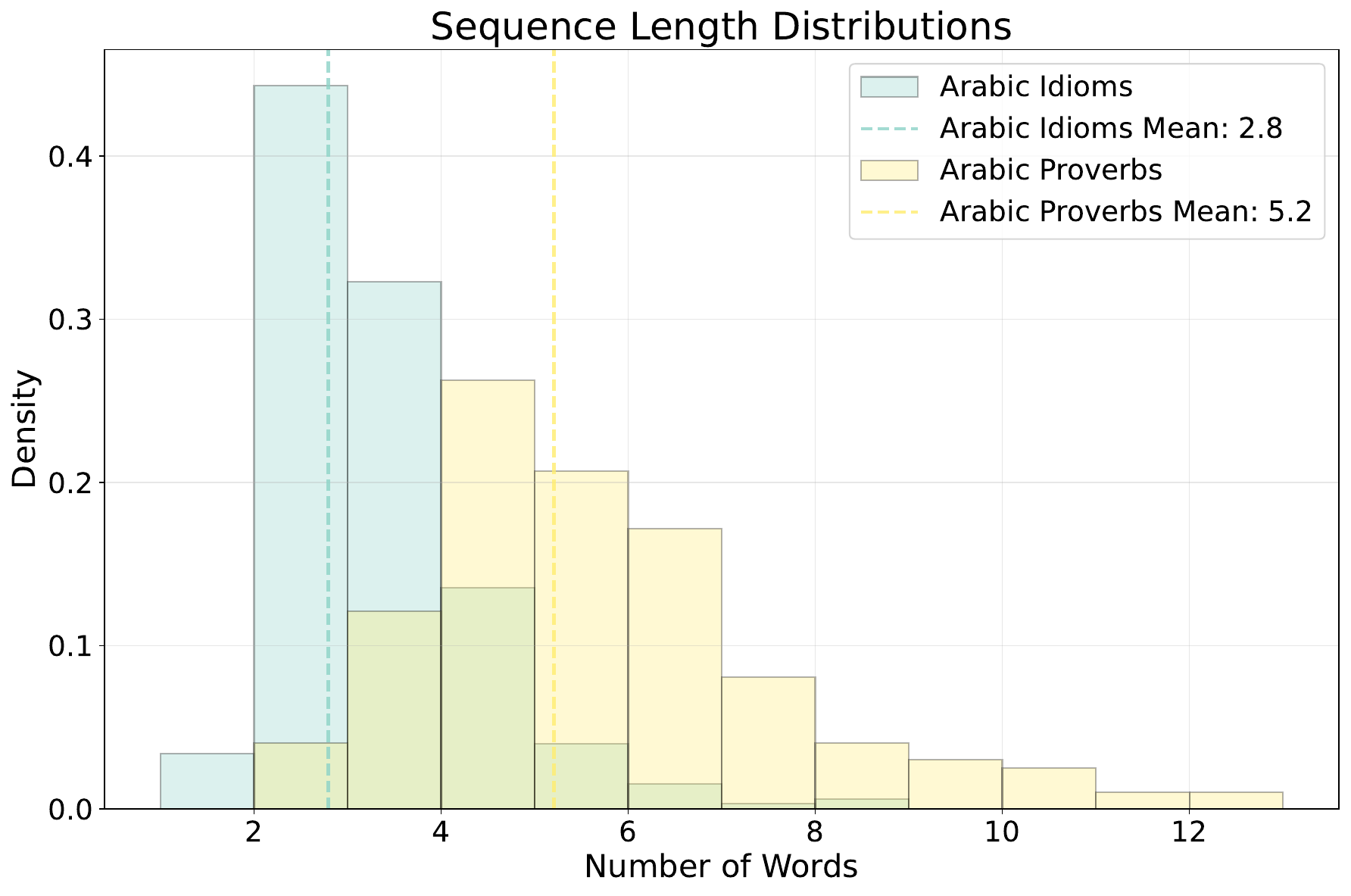}
    \caption{Sequence length distributions for Idioms and Proverbs.}
    \label{fig:length-dist}
\end{figure}

Figure~\ref{fig:length-dist} shows the sequence length distributions for Arabic idioms (Kinayat), Arabic proverbs (Jawaher), and English proverbs (MAPS), with corresponding statistics in Table~\ref{tab:dataset-stats}, highlighting that Arabic proverbs exhibit a higher mean sentence length than Arabic idioms, and English proverbs exhibit a higher mean sentence length than Arabic proverbs.

\begin{figure}[htbp]
    \centering
    \includegraphics[width=\columnwidth]{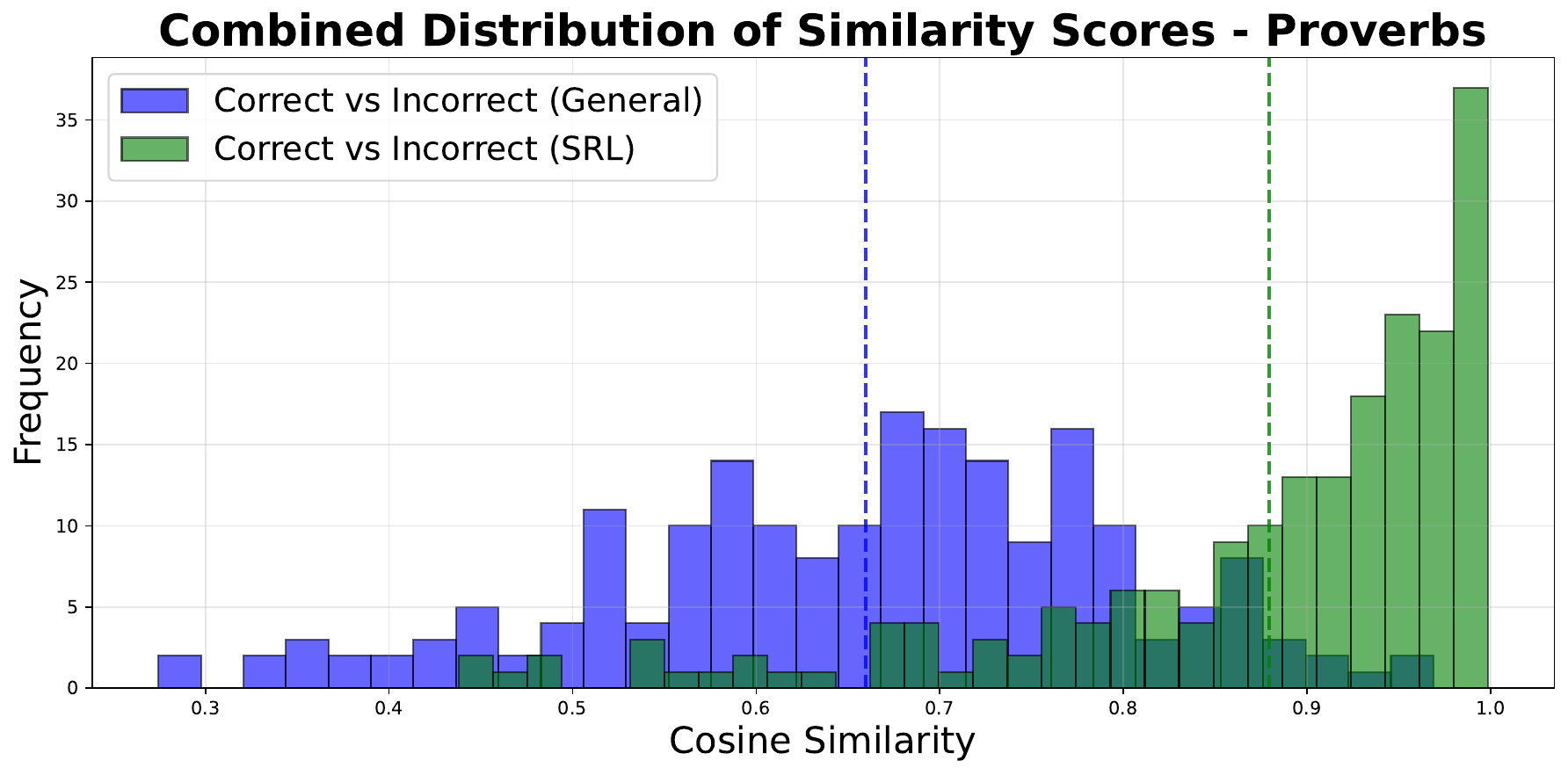}
    \caption{Cosine similarity ($\uparrow$) between the correct and incorrect explanation choices for proverbs (Jawaher).}
    \label{fig:cos-sim-proverbs}
\end{figure}

To assess semantic similarity, we compute cosine similarity between sentence embeddings using the \texttt{paraphrase-multilingual-mpnet-base-v2} model \cite{reimers-2019-sentence-bert}, comparing correct and incorrect explanations generated with the general prompt (Figure~\ref{fig:generate-prompt}) and the SRL-based prompt (Figure~\ref{fig:generate-prompt-srl}). As shown in Figure~\ref{fig:cos-sim-proverbs}, the incorrect explanations produced by the general prompt result in a roughly normal-shaped distribution (mean = 0.6596), whereas the SRL-based prompt leads to a highly right-skewed distribution (mean = 0.9239).

\begin{figure}[htbp]
    \centering
    \includegraphics[width=\columnwidth]{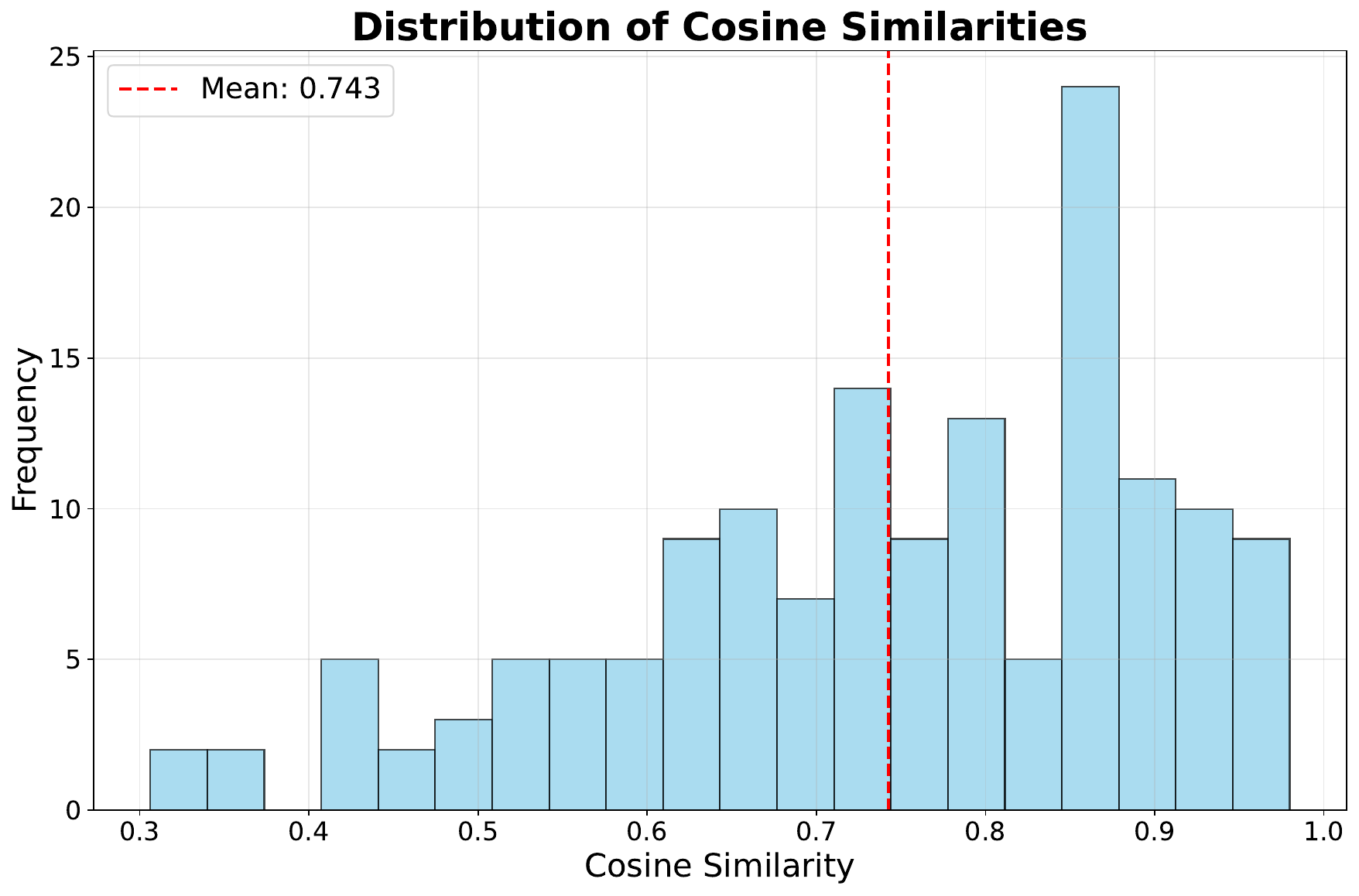}
    \caption{Cosine similarity ($\uparrow$) between the correct and incorrect idiom choices for the pragmatic use task.}
    \label{fig:prag-use-similarity}
\end{figure}

Figure~\ref{fig:prag-use-similarity} presents the distribution of cosine similarity between the two idiom options in the pragmatic use task, with a mean similarity of 0.743.

\begin{table}[ht]
\centering
\small
\begin{tabular}{lccc}
\hline
\textbf{Statistic} & \texttt{Kinayat} & \texttt{Jawaher} & \texttt{MAPS} \\
\hline
Samples   & 325  & 198 & 394\\
Mean    & 2.79 & 5.21 & 6.52\\
Median  & 3.0  & 5.0 & 6.0\\
Std     & 1.08 & 1.98 & 2.64\\
Range   & 1 -- 8 & 2 -- 12 & 3 -- 26\\
\hline
\end{tabular}
\caption{Dataset statistics for \texttt{Kinayat},  \texttt{Jawaher}, and \texttt{MAPS (English)}.}
\label{tab:dataset-stats}
\end{table}

\section{Additional Results}\label{app:more-results}
\paragraph{MCQ Positional Selection Bias}

Research has shown that LLMs can favor certain answer choices due to token-level prior probabilities \cite{zheng2024largelanguagemodelsrobust, pezeshkpour2023largelanguagemodelssensitivity}. We ran the evaluation with both permutations of the correct and incorrect answers. When the correct answer was always listed first, models performed better (91.95\% vs. 79.65\% average accuracy), as shown in Table \ref{tab:positional-bias-mcq}. For the remaining experiments, the order of the options was randomized to mitigate this bias. 
\begin{table}[htb]
\centering
\resizebox{\columnwidth}{!}{
\begin{tabular}{l | c | c | c}
\hline
\textbf{Model} 
    & \textbf{MCQ A} 
    & \textbf{MCQ B}
    & \textbf{Difference} \\
\hline
Llama-3.1-8B-Instruct     & 0.8182$\,^{\pm 0.0275}$ & 0.6667$\,^{\pm 0.0336}$ & 0.1515 \\
Gemma-2-9B-it             & 0.9343$\,^{\pm 0.0176}$ & 0.8788$\,^{\pm 0.0233}$ & 0.0556 \\
Qwen2.5-7B-Instruct       & 0.9141$\,^{\pm 0.0200}$ & 0.7828$\,^{\pm 0.0294}$ & 0.1313 \\
Qwen2.5-14B-Instruct      & 0.9192$\,^{\pm 0.0194}$ & 0.9091$\,^{\pm 0.0205}$ & 0.0101 \\
Aya-expanse-8b            & 0.8838$\,^{\pm 0.0228}$ & 0.6667$\,^{\pm 0.0336}$ & 0.2172 \\
Aya-expanse-32b           & 0.9293$\,^{\pm 0.0183}$ & 0.8838$\,^{\pm 0.0228}$ & 0.0455 \\
Mistral-7B-Instruct-v0.3  & 0.9293$\,^{\pm 0.0183}$ & 0.2475$\,^{\pm 0.0307}$ & 0.6818 \\
Jais-family-6p7b-chat     & 0.6212$\,^{\pm 0.0346}$ & 0.8182$\,^{\pm 0.0275}$ & -0.1970 \\
Fanar-1-9B-Instruct       & 0.9444$\,^{\pm 0.0163}$ & 0.8434$\,^{\pm 0.0259}$ & 0.1010 \\
SILMA-9B-Instruct-v1.0    & 0.9444$\,^{\pm 0.0163}$ & 0.7475$\,^{\pm 0.0310}$ & 0.1970 \\
ALLaM-7B-Instruct-preview & 0.9141$\,^{\pm 0.0200}$ & 0.8737$\,^{\pm 0.0237}$ & 0.0404 \\
AceGPT-v2-8B-Chat         & 0.9697$\,^{\pm 0.0122}$ & 0.5202$\,^{\pm 0.0356}$ & 0.4495 \\
Claude-Sonnet-4           & 0.9848$\,^{\pm 0.0087}$ & 0.9798$\,^{\pm 0.0100}$ & 0.0051 \\
Claude-3.5-Sonnet         & 0.9949$\,^{\pm 0.0051}$ & 0.9747$\,^{\pm 0.0112}$ & 0.0202 \\
Gemini-1.5-flash          & 0.9545$\,^{\pm 0.0148}$ & 0.9091$\,^{\pm 0.0205}$ & 0.0455 \\
GPT-4o                    & 0.9899$\,^{\pm 0.0071}$ & 0.9646$\,^{\pm 0.0132}$ & 0.0253 \\
GPT-4o-mini               & 0.9848$\,^{\pm 0.0087}$ & 0.8737$\,^{\pm 0.0237}$ & 0.1111 \\
\hline
\textbf{Average} & 0.9195 & 0.7965 & 0.1230 \\
\hline
\end{tabular}
}
\caption{Accuracy (↑)\textsuperscript{$\pm$stderr (↓)} on the positional bias MCQ understanding task for variants A and B, and their difference (A -- B) on the Jawaher dataset (general prompt used for generating incorrect choices) for a subset of models.}
\label{tab:positional-bias-mcq}
\end{table}

\paragraph{Knowledge and Understanding:}  
Table~\ref{tab:mcq-model-results} reports MCQ understanding performance across all datasets (MAPS, Kinayat, and Jawaher), while Table~\ref{tab:idioms-srl-comparison} compares results on Kinayat and Jawaher under both the general and SRL-based prompting strategies. The highest accuracy is observed on English proverbs with context (95.66\%), followed by English proverbs without context (90.86\%), multidialectal Arabic proverbs (86.57\%), and Egyptian Arabic idioms, which yield the lowest accuracy (76.29\%). Table~\ref{tab:negation-kinayat-jawaher} presents results for the task in which models are asked to identify the incorrect explanation (generated using the general prompt), showing a decline in performance compared to selecting the correct explanation. Table~\ref{tab:completion-model-results} summarizes results for the completion task, where a substantial performance gap emerges between Arabic and English proverb completion (10.64\% vs.\ 75.43\%).

\begin{figure}[htb]
    \centering
    \includegraphics[width=\columnwidth]{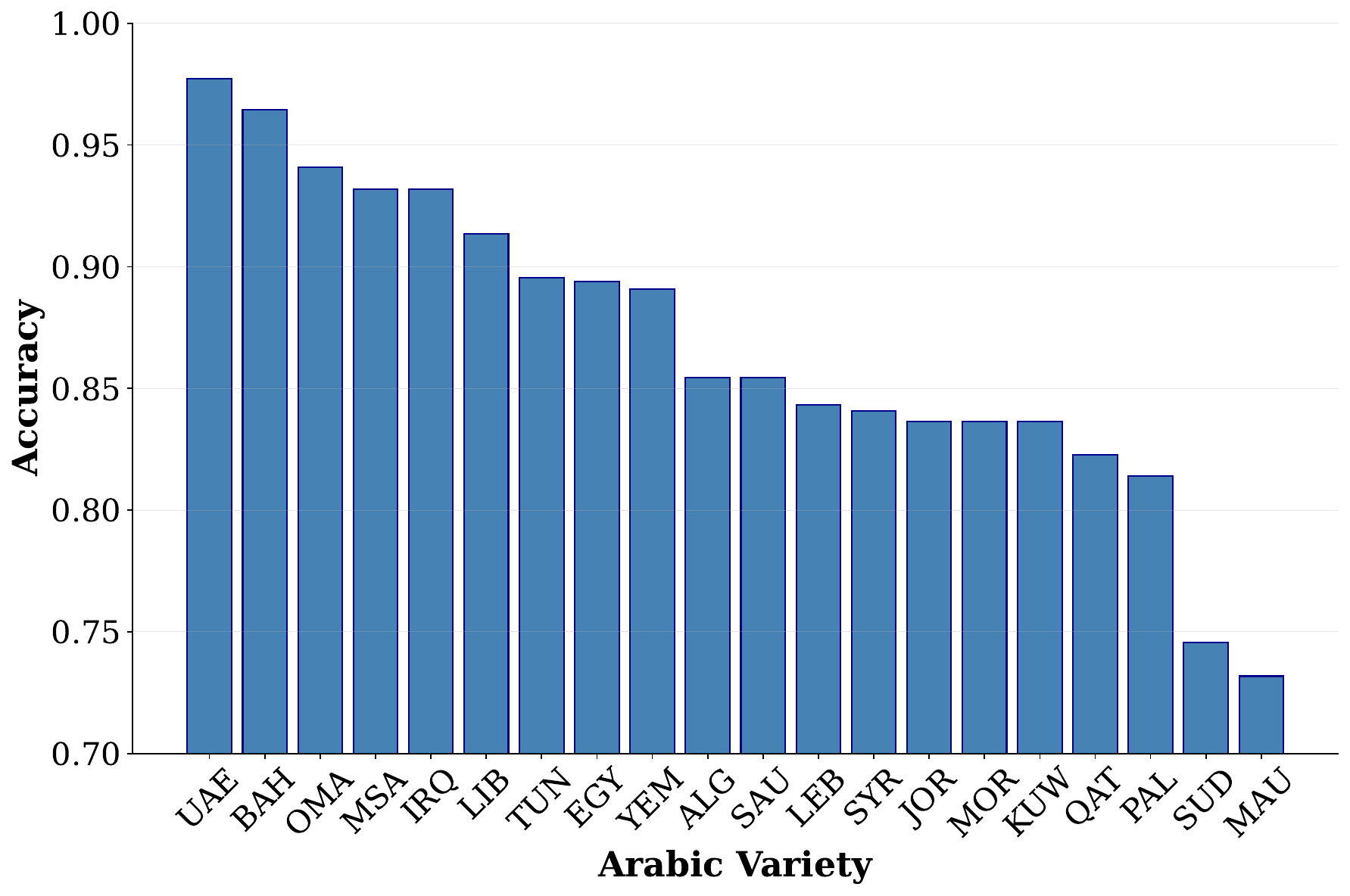}
    \caption{Country-level breakdown of MCQ Understanding Accuracy (incorrect distractor generated with general prompt).}
    \label{fig:dialect-general}
\end{figure}

\begin{figure}[htb]
    \centering
    \includegraphics[width=\columnwidth]{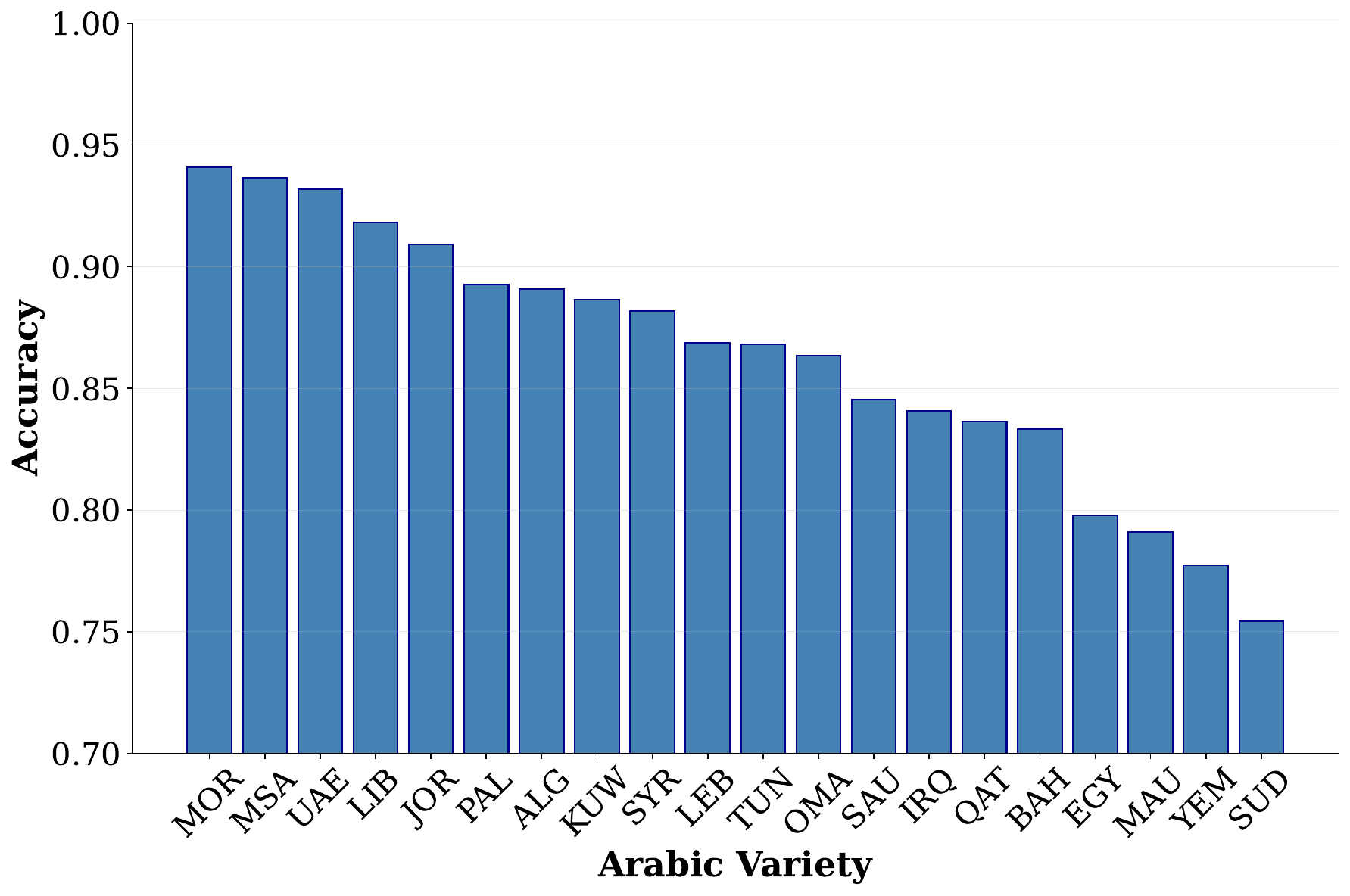}
    \caption{Country-level breakdown of MCQ Understanding Accuracy (incorrect distractor generated with SRL-based prompt).}
    \label{fig:dialect-srl}
\end{figure}

\paragraph{Country Breakdown:}  
Figures~\ref{fig:dialect-general} and \ref{fig:dialect-srl} illustrate country-level accuracies in descending order for the MCQ Understanding task using the general and SRL-based prompts, respectively. Mauritania and Sudan consistently appear among the lowest-performing dialects in both settings, whereas UAE and MSA remain within the top-performing group across both prompting strategies. Tables~\ref{tab:arabic-dialect-accuracy} and \ref{tab:arabic-dialect-accuracy-srl} provide detailed country-level results for the Jawaher dataset under each prompt condition, and Table~\ref{tab:arabic-dialect-accuracy-avg} reports the average accuracy across the two settings.

\begin{figure*}[htbp]
    \centering
    \includegraphics[width=\textwidth]{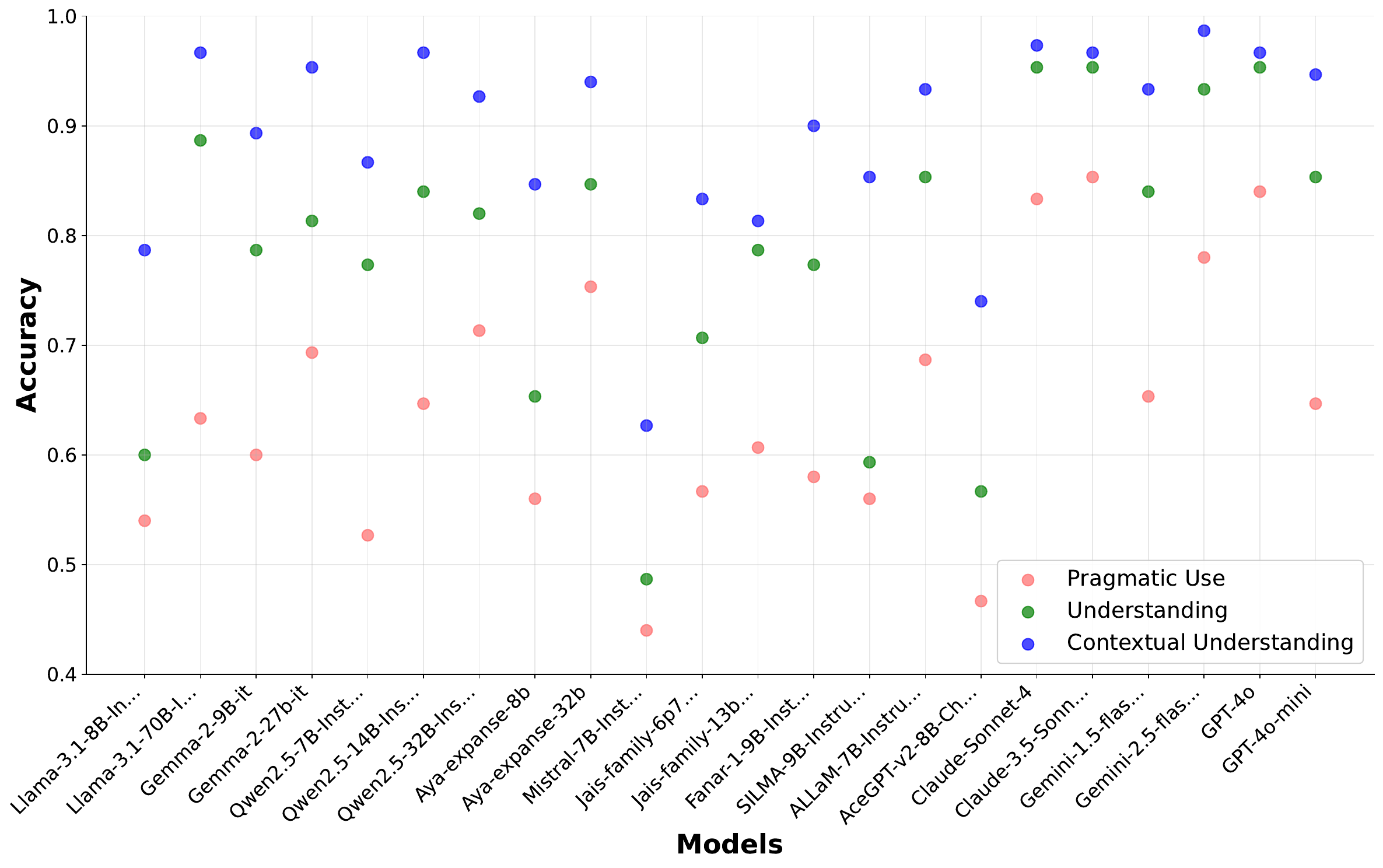}
    \caption{Accuracy ($\uparrow$) on Pragmatic Use and Understanding tasks on 150 samples from the Kinayat dataset.}
    \label{fig:prag-use-results-graph}
\end{figure*}

\paragraph{Pragmatic Use:}  
Figure~\ref{fig:prag-use-results-graph} visualizes model performance on the pragmatic use, MCQ understanding, and contextual MCQ understanding tasks using a 150-idiom subset from the Kinayat dataset, corresponding to the results reported in Table~\ref{tab:pragmatic-use}. A consistent performance gradient emerges, with the pragmatic use task yielding the lowest scores, followed by MCQ understanding, and the highest performance observed in contextual MCQ understanding.

\paragraph{Connotations:}  
Table~\ref{tab:sentiment-match} presents the connotation task results for the Jawaher and Kinayat datasets, restricted to samples with full inter-annotator agreement (105 entries for Jawaher and 104 for Kinayat). Among all models, the Claude family (Claude Sonnet 4 and Claude 3.5 Sonnet) achieved the highest performance. Overall, proverb connotations were slightly harder to identify than idiomatic ones, with lower average model agreement (49.71\% vs. 50.10\%). However, models demonstrated stronger performance when inferring connotations from the explanations, achieving average accuracies of 68.17\% for proverbs and 72.21\% for idioms. 

\paragraph{Overall Performance:}  
Table~\ref{tab:averages} presents the average performance of multilingual open-source, Arabic open-source, and closed-source models across all tasks. On average, multilingual open-source models slightly outperformed Arabic models on most tasks; however, Arabic models led on proverb completion, Kinayat MCQ (SRL-prompt), Kinayat pragmatic use, and explanation generation for both proverbs and idioms. These averages, however, mask variation within each category. For example, the multilingual model Mistral-7B-Instruct performed below most Arabic models on several tasks. In contrast, closed-source models consistently outperformed both Arabic and multilingual open-source models across all evaluations. One potential contributing factor is model scale, as closed-source systems are generally assumed to be larger than the open-source models evaluated, though their exact sizes are not publicly disclosed.

\begin{table}[htb]
\centering
\resizebox{\columnwidth}{!}{
\begin{tabular}{l | c | c}
\hline
\textbf{Model} 
    & \textbf{Kinayat} 
    & \textbf{Jawaher} \\
\hline
Llama-3.1-8B-Instruct     & 0.6492$\,^{\pm 0.0265}$ & 0.7424$\,^{\pm 0.0312}$ \\
			
Llama-3.1-70B-Instruct    & 0.5508$\,^{\pm 0.0276}$ & 0.6919$\,^{\pm 0.0329}$ \\
Gemma-2-9B-it             & 0.6031$\,^{\pm 0.0272}$ & 0.8384$\,^{\pm 0.0262}$ \\
Gemma-2-27b-it            & 0.8123$\,^{\pm 0.0217}$ & 0.9192$\,^{\pm 0.0194}$ \\
Qwen2.5-7B-Instruct       & 0.6277$\,^{\pm 0.0269}$ & 0.7677$\,^{\pm 0.0301}$ \\
Qwen2.5-14B-Instruct      & 0.7908$\,^{\pm 0.0226}$ & 0.8737$\,^{\pm 0.0237}$ \\
Qwen2.5-32B-Instruct      & 0.8000$\,^{\pm 0.0222}$ & 0.9141$\,^{\pm 0.0200}$ \\
Aya-expanse-8b            & 0.5754$\,^{\pm 0.0275}$ & 0.7020$\,^{\pm 0.0326}$ \\
Aya-expanse-32b           & 0.8092$\,^{\pm 0.0218}$ & 0.8586$\,^{\pm 0.0248}$ \\
Mistral-7B-Instruct-v0.3  & 0.4923$\,^{\pm 0.0278}$ & 0.5808$\,^{\pm 0.0352}$ \\
\hline
Jais-family-6p7b-chat     & 0.6062$\,^{\pm 0.0271}$ & 0.7374$\,^{\pm 0.0314}$ \\
Jais-family-13b-chat      & 0.5846$\,^{\pm 0.0274}$ & 0.7020$\,^{\pm 0.0326}$ \\
Fanar-1-9B-Instruct       & 0.5815$\,^{\pm 0.0274}$ & 0.7980$\,^{\pm 0.0286}$ \\
SILMA-9B-Instruct-v1.0    & 0.5354$\,^{\pm 0.0277}$ & 0.8131$\,^{\pm 0.0278}$ \\
ALLaM-7B-Instruct-preview & 0.8000$\,^{\pm 0.0222}$ & 0.8990$\,^{\pm 0.0215}$ \\
AceGPT-v2-8B-Chat         & 0.5969$\,^{\pm 0.0273}$ & 0.7374$\,^{\pm 0.0314}$ \\
\hline
Claude-Sonnet-4           & 0.9446$\,^{\pm 0.0127}$ & 0.9697$\,^{\pm 0.0122}$ \\
Claude-3.5-Sonnet         & 0.9169$\,^{\pm 0.0153}$ & 0.9747$\,^{\pm 0.0112}$ \\
Gemini-1.5-flash          & 0.7785$\,^{\pm 0.0231}$ & 0.8939$\,^{\pm 0.0219}$ \\
Gemini-2.5-flash-lite-preview-06-17 & 0.8308$\,^{\pm 0.0208}$ & 0.9242$\,^{\pm 0.0189}$ \\
GPT-4o                    & 0.9477$\,^{\pm 0.0124}$ & 0.9798$\,^{\pm 0.0100}$ \\
GPT-4o-mini               & 0.7785$\,^{\pm 0.0231}$ & 0.8788$\,^{\pm 0.0233}$ \\
\hline
\textbf{Average}          & 0.7097$\,^{\pm 0.0000}$ & 0.8271$\,^{\pm 0.0000}$ \\
\hline
\end{tabular}
}
\caption{Accuracy (↑)\textsuperscript{$\pm$stderr (↓)} on the task of selecting the incorrect explanation for Kinayat and Jawaher datasets.}
\label{tab:negation-kinayat-jawaher}
\end{table}

\begin{table}[!htb]
\centering
\resizebox{\columnwidth}{!}{
\begin{tabular}{l|c|c}
\hline
\textbf{Model} & \textbf{MAPS} & \textbf{Jawaher} \\
\hline
Llama-3.1-8B-Instruct & $0.6954^{\,\pm 0.0232}$ & $0.0253^{\,\pm 0.0112}$ \\
			
Llama-3.1-70B-Instruct & $0.8756^{\,\pm 0.0166}$ & $0.0707^{\,\pm 0.0183}$ \\
Gemma-2-9B-it & $0.8046^{\,\pm 0.0200}$ & $0.0556^{\,\pm 0.0163}$ \\
Gemma-2-27b-it & $0.6751^{\,\pm 0.0236}$ & $0.0808^{\,\pm 0.0194}$ \\
Qwen2.5-7B-Instruct & $0.6421^{\,\pm 0.0242}$ & $0.0253^{\,\pm 0.0112}$ \\
Qwen2.5-14B-Instruct & $0.7792^{\,\pm 0.0209}$ & $0.0354^{\,\pm 0.0132}$ \\
Qwen2.5-32B-Instruct & $0.8401^{\,\pm 0.0185}$ & $0.0606^{\,\pm 0.0170}$ \\
Aya-expanse-8b & $0.5051^{\,\pm 0.0252}$ & $0.0556^{\,\pm 0.0163}$ \\
Aya-expanse-32b & $0.7107^{\,\pm 0.0229}$ & $0.1364^{\,\pm 0.0245}$ \\
Mistral-7B-Instruct-v0.3 & $0.7614^{\,\pm 0.0215}$ & $0.0000^{\,\pm 0.0000}$ \\
\hline
Jais-family-6p7b-chat & $0.2995^{\,\pm 0.0231}$ & $0.0303^{\,\pm 0.0122}$ \\
Jais-family-13b-chat & $0.3604^{\,\pm 0.0242}$ & $0.0152^{\,\pm 0.0087}$ \\
Fanar-1-9B-Instruct & $0.8046^{\,\pm 0.0200}$ & $0.1111^{\,\pm 0.0224}$ \\
SILMA-9B-Instruct-v1.0 & $0.8756^{\,\pm 0.0166}$ & $0.0354^{\,\pm 0.0132}$ \\
ALLaM-7B-Instruct-preview & $0.7107^{\,\pm 0.0229}$ & $0.1313^{\,\pm 0.0241}$ \\
AceGPT-v2-8B-Chat & $0.7690^{\,\pm 0.0213}$ & $0.0808^{\,\pm 0.0194}$ \\
\hline
Claude-Sonnet-4 & $0.9340^{\,\pm 0.0125}$ & $0.2980^{\,\pm 0.0326}$ \\
Claude-3.5-Sonnet & $\textbf{0.9391}^{\,\pm 0.0121}$ & $\textbf{0.3636}^{\,\pm 0.0343}$ \\
Gemini-1.5-flash & $0.9061^{\,\pm 0.0147}$ & $0.1212^{\,\pm 0.0233}$ \\
Gemini-2.5-flash-lite-preview-06-17 & $0.8782^{\,\pm 0.0165}$ & $0.2273^{\,\pm 0.0299}$ \\
GPT-4o & $0.9340^{\,\pm 0.0125}$ & $0.2879^{\,\pm 0.0323}$ \\
GPT-4o-mini & $0.8934^{\,\pm 0.0156}$ & $0.0960^{\,\pm 0.0210}$ \\
\hline
\textbf{Average} & 0.7543 & 0.1065 \\
\hline
\end{tabular}
}
\caption{Evaluation results of completion task (accuracy (↑)\textsuperscript{$\pm$stderr (↓)}) for different models on MAPS and Jawaher datasets.}
\label{tab:completion-model-results}
\end{table}

\begin{table*}[ht]
\centering
\resizebox{\textwidth}{!}{
\begin{tabular}{l|c|c|c|c}
\hline
\textbf{Model} 
    & \textbf{MAPS} 
    & \textbf{MAPS + Context} 
    & \textbf{Jawaher} 
    & \textbf{Kinayat} \\
\hline
Llama-3.1-8B-Instruct     & 0.8655\,\textsuperscript{$\pm$0.0172} & 0.9213\,\textsuperscript{$\pm$0.0136} & 0.7475\,\textsuperscript{$\pm$0.0310} & 0.5754\,\textsuperscript{$\pm$0.0275} \\

Llama-3.1-70B-Instruct    & 0.9239\,\textsuperscript{$\pm$0.0134} & 0.9873\,\textsuperscript{$\pm$0.0056} & 
0.8990\,\textsuperscript{$\pm$0.0215} & 0.8831\,\textsuperscript{$\pm$0.0179} \\
Gemma-2-9B-it             & 0.9365\,\textsuperscript{$\pm$0.0123} & 0.9670\,\textsuperscript{$\pm$0.0090} & 0.9091\,\textsuperscript{$\pm$0.0205} & 0.7354\,\textsuperscript{$\pm$0.0245} \\
Gemma-2-27b-it            & 0.9340\,\textsuperscript{$\pm$0.0125} & 0.9721\,\textsuperscript{$\pm$0.0083} & 0.8990\,\textsuperscript{$\pm$0.0215} & 0.7908\,\textsuperscript{$\pm$0.0226} \\
Qwen2.5-7B-Instruct       & 0.9239\,\textsuperscript{$\pm$0.0134} & 0.9492\,\textsuperscript{$\pm$0.0111} & 0.8535\,\textsuperscript{$\pm$0.0252} & 0.7262\,\textsuperscript{$\pm$0.0248} \\
Qwen2.5-14B-Instruct      & 0.9391\,\textsuperscript{$\pm$0.0121} & 0.9695\,\textsuperscript{$\pm$0.0087} & 0.8990\,\textsuperscript{$\pm$0.0215} & 0.7938\,\textsuperscript{$\pm$0.0225} \\
Qwen2.5-32B-Instruct      & 0.9391\,\textsuperscript{$\pm$0.0121} & \textbf{0.9924}\,\textsuperscript{$\pm$0.0044} & 0.9192\,\textsuperscript{$\pm$0.0194} & 0.8000\,\textsuperscript{$\pm$0.0222} \\
Aya-expanse-8b            & 0.8858\,\textsuperscript{$\pm$0.0160} & 0.9315\,\textsuperscript{$\pm$0.0127} & 0.7929\,\textsuperscript{$\pm$0.0289} & 0.6338\,\textsuperscript{$\pm$0.0268} \\
Aya-expanse-32b           & 0.9340\,\textsuperscript{$\pm$0.0125} & 0.9797\,\textsuperscript{$\pm$0.0071} & 0.8939\,\textsuperscript{$\pm$0.0219} & 0.8154\,\textsuperscript{$\pm$0.0216} \\
Mistral-7B-Instruct-v0.3  & 0.8579\,\textsuperscript{$\pm$0.0176} & 0.9213\,\textsuperscript{$\pm$0.0136} & 0.6061\,\textsuperscript{$\pm$0.0348} & 0.5169\,\textsuperscript{$\pm$0.0278} \\
\hline
Jais-family-6p7b-chat     & 0.7640\,\textsuperscript{$\pm$0.0214} & 0.8680\,\textsuperscript{$\pm$0.0171} & 0.7525\,\textsuperscript{$\pm$0.0307} & 0.7046\,\textsuperscript{$\pm$0.0253} \\
Jais-family-13b-chat     & 0.8376\,\textsuperscript{$\pm$0.0186} & 0.9162\,\textsuperscript{$\pm$0.0140} & 0.7626\,\textsuperscript{$\pm$0.0303} & 0.6769\,\textsuperscript{$\pm$0.0260} \\
Fanar-1-9B-Instruct       & 0.9010\,\textsuperscript{$\pm$0.0151} & 0.9442\,\textsuperscript{$\pm$0.0116} & 0.8737\,\textsuperscript{$\pm$0.0237} & 0.7508\,\textsuperscript{$\pm$0.0240} \\
SILMA-9B-Instruct-v1.0    & 0.9492\,\textsuperscript{$\pm$0.0111} & 0.9569\,\textsuperscript{$\pm$0.0102} & 0.8737\,\textsuperscript{$\pm$0.0237} & 0.6277\,\textsuperscript{$\pm$0.0269} \\
ALLaM-7B-Instruct-preview & 0.8807\,\textsuperscript{$\pm$0.0164} & 0.9365\,\textsuperscript{$\pm$0.0123} & 0.8636\,\textsuperscript{$\pm$0.0245} & 0.7846\,\textsuperscript{$\pm$0.0228} \\
AceGPT-v2-8B-Chat         & 0.9112\,\textsuperscript{$\pm$0.0144} & 0.9543\,\textsuperscript{$\pm$0.0105} & 0.7475\,\textsuperscript{$\pm$0.0310} & 0.5754\,\textsuperscript{$\pm$0.0275} \\
\hline
Claude-Sonnet-4           & 0.9340\,\textsuperscript{$\pm$0.0125} & 0.9797\,\textsuperscript{$\pm$0.0071} & 0.9798\,\textsuperscript{$\pm$0.0100} & \textbf{0.9662}\,\textsuperscript{$\pm$0.0100} \\
Claude-3.5-Sonnet         & 0.9543\,\textsuperscript{$\pm$0.0105} & 0.9772\,\textsuperscript{$\pm$0.0075} & \textbf{0.9848}\,\textsuperscript{$\pm$0.0087} & 0.9415\,\textsuperscript{$\pm$0.0130} \\
Gemini-1.5-flash          & 0.9239\,\textsuperscript{$\pm$0.0134} & 0.9746\,\textsuperscript{$\pm$0.0079} & 0.9343\,\textsuperscript{$\pm$0.0176} & 0.8185\,\textsuperscript{$\pm$0.0214} \\
Gemini-2.5-flash-lite-preview-06-17 & 0.8832\,\textsuperscript{$\pm$0.0162} & 0.9746\,\textsuperscript{$\pm$0.0079} & 0.9596\,\textsuperscript{$\pm$0.0140} & 0.9077\,\textsuperscript{$\pm$0.0161} \\
GPT-4o                    & \textbf{0.9619}\,\textsuperscript{$\pm$0.0097} & 0.9873\,\textsuperscript{$\pm$0.0056} & 0.9798\,\textsuperscript{$\pm$0.0100} & 0.9415\,\textsuperscript{$\pm$0.0130} \\
GPT-4o-mini               & 0.9492\,\textsuperscript{$\pm$0.0111} & 0.9848\,\textsuperscript{$\pm$0.0062} & 0.9141\,\textsuperscript{$\pm$0.0200} & 0.8185\,\textsuperscript{$\pm$0.0214} \\
\hline
Average & 0.9086 & 0.9566 & 0.8657 & 0.7629 \\
\hline
\end{tabular}
}
\caption{Evaluation results (accuracy (↑)\textsuperscript{$\pm$stderr (↓)}) of multiple choice understanding task on different test sets. The incorrect explanation choices for Jawaher and Kinayat for the results shown here were generated with the general prompt.}
\label{tab:mcq-model-results}
\end{table*}

\begin{table*}[t]
\centering
\resizebox{\textwidth}{!}{
\begin{tabular}{l | c | c | c | c}
\hline
\textbf{Model} 
    & \textbf{Jawaher (General)} 
    & \textbf{Jawaher (SRL)}
    & \textbf{Kinayat (General)}
    & \textbf{Kinayat (SRL)} \\
\hline
Llama-3.1-8B-Instruct     & 0.7475$\,^{\pm 0.0310}$ & 0.7475$\,^{\pm 0.0310}$ & 0.5754$\,^{\pm 0.0275}$ & 0.6400$\,^{\pm 0.0267}$ \\
							
Llama-3.1-70B-Instruct    & 0.8990$\,^{\pm 0.0215}$ & 0.9141$\,^{\pm 0.0200}$ & 0.8990\,\textsuperscript{$\pm$0.0215} & 0.8492$\,^{\pm 0.0199}$ \\
Gemma-2-9B-it             & 0.9091$\,^{\pm 0.0205}$ & 0.8838$\,^{\pm 0.0228}$ & 0.7354$\,^{\pm 0.0245}$ & 0.7692$\,^{\pm 0.0234}$ \\
Gemma-2-27b-it            & 0.8990$\,^{\pm 0.0215}$ & 0.8939$\,^{\pm 0.0219}$ & 0.7908$\,^{\pm 0.0226}$ & 0.8123$\,^{\pm 0.0217}$ \\
Qwen2.5-7B-Instruct       & 0.8535$\,^{\pm 0.0252}$ & 0.8687$\,^{\pm 0.0241}$ & 0.7262$\,^{\pm 0.0248}$ & 0.7815$\,^{\pm 0.0230}$ \\
Qwen2.5-14B-Instruct      & 0.8990$\,^{\pm 0.0215}$ & 0.9293$\,^{\pm 0.0183}$ & 0.7938$\,^{\pm 0.0225}$ & 0.8431$\,^{\pm 0.0202}$ \\
Qwen2.5-32B-Instruct      & 0.9192$\,^{\pm 0.0194}$ & 0.9545$\,^{\pm 0.0148}$ & 0.8000$\,^{\pm 0.0222}$ & 0.8308$\,^{\pm 0.0208}$ \\
Aya-expanse-8b            & 0.7929$\,^{\pm 0.0289}$ & 0.7677$\,^{\pm 0.0301}$ & 0.6338$\,^{\pm 0.0268}$ & 0.7323$\,^{\pm 0.0246}$ \\
Aya-expanse-32b           & 0.8939$\,^{\pm 0.0219}$ & 0.9242$\,^{\pm 0.0189}$ & 0.8154$\,^{\pm 0.0216}$ & 0.8185$\,^{\pm 0.0214}$ \\
Mistral-7B-Instruct-v0.3  & 0.6061$\,^{\pm 0.0348}$ & 0.5960$\,^{\pm 0.0350}$ & 0.5169$\,^{\pm 0.0278}$ & 0.5262$\,^{\pm 0.0277}$ \\
\hline
Jais-family-6p7b-chat     & 0.7525$\,^{\pm 0.0307}$ & 0.7929$\,^{\pm 0.0289}$ & 0.7046$\,^{\pm 0.0253}$ & 0.7262$\,^{\pm 0.0248}$ \\
Jais-family-13b-chat      & 0.7626$\,^{\pm 0.0303}$ & 0.7879$\,^{\pm 0.0291}$ & 0.6769$\,^{\pm 0.0260}$ & 0.7446$\,^{\pm 0.0242}$ \\
Fanar-1-9B-Instruct       & 0.8737$\,^{\pm 0.0237}$ & 0.8737$\,^{\pm 0.0237}$ & 0.7508$\,^{\pm 0.0240}$ & 0.7446$\,^{\pm 0.0242}$ \\
SILMA-9B-Instruct-v1.0    & 0.8737$\,^{\pm 0.0237}$ & 0.8485$\,^{\pm 0.0255}$ & 0.6277$\,^{\pm 0.0269}$ & 0.7415$\,^{\pm 0.0243}$ \\
ALLaM-7B-Instruct-preview & 0.8636$\,^{\pm 0.0245}$ & 0.8586$\,^{\pm 0.0248}$ & 0.7846$\,^{\pm 0.0228}$ & 0.8462$\,^{\pm 0.0200}$ \\
AceGPT-v2-8B-Chat         & 0.7475$\,^{\pm 0.0310}$ & 0.7576$\,^{\pm 0.0305}$ & 0.5754$\,^{\pm 0.0275}$ & 0.6585$\,^{\pm 0.0263}$ \\
\hline
Claude-Sonnet-4           & 0.9798$\,^{\pm 0.0100}$ & \textbf{0.9848}$\,^{\pm 0.0087}$ & \textbf{0.9662}$\,^{\pm 0.0100}$ & 0.9077$\,^{\pm 0.0161}$ \\
Claude-3.5-Sonnet         & \textbf{0.9848}$\,^{\pm 0.0087}$ & 0.9646$\,^{\pm 0.0132}$ & 0.9415$\,^{\pm 0.0130}$ & 0.9046$\,^{\pm 0.0163}$ \\
Gemini-1.5-flash          & 0.9343$\,^{\pm 0.0176}$ & 0.8990$\,^{\pm 0.0215}$ & 0.8185$\,^{\pm 0.0214}$ & 0.8738$\,^{\pm 0.0184}$ \\
Gemini-2.5-flash-lite-preview-06-17 & 0.9596$\,^{\pm 0.0140}$ & 0.9343$\,^{\pm 0.0176}$ & 0.9077$\,^{\pm 0.0161}$ & 0.8431$\,^{\pm 0.0202}$ \\
GPT-4o                    & 0.9798$\,^{\pm 0.0100}$ & 0.9596$\,^{\pm 0.0140}$ & 0.9415$\,^{\pm 0.0130}$ & \textbf{0.9138}$\,^{\pm 0.0156}$ \\
GPT-4o-mini               & 0.9141$\,^{\pm 0.0200}$ & 0.8586$\,^{\pm 0.0248}$ & 0.8185$\,^{\pm 0.0214}$ & 0.7662$\,^{\pm 0.0235}$ \\
\hline
						
\textbf{Average}          & 0.8657 & 0.8636 & 0.7629 & 0.7852 \\
\hline
\end{tabular}
}
\caption{Accuracy (↑)\textsuperscript{$\pm$stderr (↓)}) on Jawaher and Kinayat datasets with incorrect explanations generated by a general vs. SRL variant prompt.}
\label{tab:idioms-srl-comparison}
\end{table*}

\begin{table*}[t]
\centering
\small
\begin{tabular}{l | cc | cc}
\hline
\textbf{Model} & \multicolumn{2}{c|}{\textbf{Jawaher}} & \multicolumn{2}{c}{\textbf{Kinayat}} \\
               & \textbf{BERT-F1} & \textbf{LLM-Judge}           & \textbf{BERT-F1} & \textbf{LLM-Judge} \\
\hline
Llama-3.1-8B-Instruct     & 0.5988$\,^{\pm 0.0022}$ & 2.1111$\,^{\pm 0.0701}$ & 0.5714$\,^{\pm 0.0015}$ & 1.5292$\,^{\pm 0.0350}$ \\				
Llama-3.1-70B-Instruct    & 0.6816$\,^{\pm 0.0021}$ & 3.1869$\,^{\pm 0.0934}$ & 0.6486$\,^{\pm 0.0017}$ & 2.1138$\,^{\pm 0.0531}$ \\
Gemma-2-9B-it             & 0.6774$\,^{\pm 0.0022}$ & 2.9293$\,^{\pm 0.0899}$ & 0.6492$\,^{\pm 0.0018}$ & 2.0862$\,^{\pm 0.0452}$ \\
Gemma-2-27b-it            & 0.6893$\,^{\pm 0.0019}$ & 3.0455$\,^{\pm 0.0921}$ & 0.6793$\,^{\pm 0.0012}$ & 2.1877$\,^{\pm 0.0517}$ \\
Qwen2.5-7B-Instruct       & 0.6759$\,^{\pm 0.0020}$ & 2.5303$\,^{\pm 0.0793}$ & 0.6584$\,^{\pm 0.0012}$ & 2.0646$\,^{\pm 0.0477}$ \\
Qwen2.5-14B-Instruct      & 0.6651$\,^{\pm 0.0031}$ & 2.8485$\,^{\pm 0.1071}$ & 0.6560$\,^{\pm 0.0013}$ & 2.1662$\,^{\pm 0.0518}$ \\
Qwen2.5-32B-Instruct      & 0.6737$\,^{\pm 0.0023}$ & 3.2020$\,^{\pm 0.0974}$ & 0.6541$\,^{\pm 0.0016}$ & 2.2092$\,^{\pm 0.0566}$ \\
Aya-expanse-8b            & 0.6794$\,^{\pm 0.0018}$ & 2.9343$\,^{\pm 0.0859}$ & 0.6633$\,^{\pm 0.0012}$ & 2.1815$\,^{\pm 0.0512}$ \\
Aya-expanse-32b           & 0.6820$\,^{\pm 0.0019}$ & 3.4646$\,^{\pm 0.0935}$ & 0.6615$\,^{\pm 0.0014}$ & 2.4123$\,^{\pm 0.0595}$ \\
Mistral-7B-Instruct-v0.3  & 0.6695$\,^{\pm 0.0019}$ & 2.1313$\,^{\pm 0.0528}$ & 0.6373$\,^{\pm 0.0019}$ & 1.7415$\,^{\pm 0.0335}$ \\
\hline
Jais-family-6p7b-chat     & 0.6762$\,^{\pm 0.0028}$ & 2.2071$\,^{\pm 0.0703}$ & 0.6226$\,^{\pm 0.0028}$ & 1.5200$\,^{\pm 0.0419}$ \\
Jais-family-13b-chat      & 0.6598$\,^{\pm 0.0033}$ & 2.4495$\,^{\pm 0.0812}$ & 0.6267$\,^{\pm 0.0023}$ & 1.7108$\,^{\pm 0.0430}$ \\
Fanar-1-9B-Instruct       & 0.6799$\,^{\pm 0.0019}$ & 3.3535$\,^{\pm 0.0899}$ & 0.6687$\,^{\pm 0.0011}$ & 2.3200$\,^{\pm 0.0555}$ \\
SILMA-9B-Instruct-v1.0    & 0.6856$\,^{\pm 0.0026}$ & 2.5556$\,^{\pm 0.0824}$ & 0.6581$\,^{\pm 0.0025}$ & 1.6923$\,^{\pm 0.0417}$ \\
ALLaM-7B-Instruct-preview & 0.6893$\,^{\pm 0.0021}$ & 3.2727$\,^{\pm 0.0902}$ & 0.6675$\,^{\pm 0.0011}$ & 2.3077$\,^{\pm 0.0606}$ \\
AceGPT-v2-8B-Chat         & 0.6730$\,^{\pm 0.0017}$ & 3.0606$\,^{\pm 0.0870}$ & 0.6455$\,^{\pm 0.0010}$ & 2.0031$\,^{\pm 0.0450}$ \\
\hline
Claude-Sonnet-4           & 0.6990$\,^{\pm 0.0018}$ & 3.7778$\,^{\pm 0.0848}$ & 0.6795$\,^{\pm 0.0012}$ & 2.6646$\,^{\pm 0.0687}$ \\
Claude-3.5-Sonnet         & \textbf{0.6998}$\,^{\pm 0.0017}$ & \textbf{3.8939}$\,^{\pm 0.0887}$ & \textbf{0.6798}$\,^{\pm 0.0013}$ & \textbf{2.9262}$\,^{\pm 0.0724}$ \\
Gemini-1.5-flash          & 0.6700$\,^{\pm 0.0020}$ & 3.5152$\,^{\pm 0.0911}$ & 0.6360$\,^{\pm 0.0028}$ & 2.4769$\,^{\pm 0.0587}$ \\
Gemini-2.5-flash-lite-preview-06-17 & 0.6827$\,^{\pm 0.0020}$ & 3.6818$\,^{\pm 0.0921}$ & 0.6684$\,^{\pm 0.0013}$ & 2.8400$\,^{\pm 0.0700}$ \\
GPT-4o                    & 0.6811$\,^{\pm 0.0022}$ & 3.6667$\,^{\pm 0.0982}$ & 0.6560$\,^{\pm 0.0011}$ & 2.9046$\,^{\pm 0.0819}$ \\
GPT-4o-mini               & 0.6813$\,^{\pm 0.0019}$ & 3.4394$\,^{\pm 0.1029}$ & 0.6552$\,^{\pm 0.0010}$ & 2.1231$\,^{\pm 0.0682}$ \\
\hline
						
\textbf{Average}          & 0.6759                  & 3.0572                  & 0.6520                  & 2.1901 \\
\hline
\end{tabular}
\caption{Explanation generation scores using BERTScore-F1 ($\uparrow$) and LLM-as-a-judge ($\uparrow$) for \textbf{Jawaher} and \textbf{Kinayat}.}
\label{tab:gen-eval-grouped}
\end{table*}

\begin{table*}[!ht]
\centering
\resizebox{\textwidth}{!}{
\begin{tabular}{l | c | c | c | c}
\hline
\textbf{Model}
    & \textbf{Proverb}
    & \textbf{Proverb Explanation}
    & \textbf{Idiom}
    & \textbf{Idiom Explanation} \\
\hline
Llama-3.1-8B-Instruct     & 0.5577$\,^{\pm 0.0489}$ & 0.7019$\,^{\pm 0.0451}$ & 0.7404$\,^{\pm 0.0432}$ & 0.7692$\,^{\pm 0.0415}$ \\
Llama-3.1-70B-Instruct    & 0.2788$\,^{\pm 0.0442}$ & 0.7019$\,^{\pm 0.0451}$ & 0.2115$\,^{\pm 0.0402}$ & 0.8269$\,^{\pm 0.0373}$ \\
Gemma-2-9B-it             & 0.6058$\,^{\pm 0.0482}$ & 0.7981$\,^{\pm 0.0396}$ & 0.7692$\,^{\pm 0.0415}$ & 0.8654$\,^{\pm 0.0336}$ \\
Gemma-2-27b-it            & 0.6250$\,^{\pm 0.0477}$ & 0.7885$\,^{\pm 0.0402}$ & 0.8269$\,^{\pm 0.0373}$ & 0.8750$\,^{\pm 0.0326}$ \\
Qwen2.5-7B-Instruct       & 0.5288$\,^{\pm 0.0492}$ & 0.7788$\,^{\pm 0.0409}$ & 0.5481$\,^{\pm 0.0490}$ & 0.8365$\,^{\pm 0.0364}$ \\
Qwen2.5-14B-Instruct      & 0.6442$\,^{\pm 0.0472}$ & 0.7788$\,^{\pm 0.0409}$ & 0.7692$\,^{\pm 0.0415}$ & 0.8558$\,^{\pm 0.0346}$ \\
Qwen2.5-32B-Instruct      & 0.5673$\,^{\pm 0.0488}$ & 0.8173$\,^{\pm 0.0381}$ & 0.5769$\,^{\pm 0.0487}$ & 0.8558$\,^{\pm 0.0346}$ \\
Aya-expanse-8b            & 0.3462$\,^{\pm 0.0469}$ & 0.6058$\,^{\pm 0.0482}$ & 0.2596$\,^{\pm 0.0432}$ & 0.5769$\,^{\pm 0.0487}$ \\
Aya-expanse-32b           & 0.5962$\,^{\pm 0.0483}$ & 0.7308$\,^{\pm 0.0437}$ & 0.3462$\,^{\pm 0.0469}$ & 0.7788$\,^{\pm 0.0409}$ \\
Mistral-7B-Instruct-v0.3  & 0.5096$\,^{\pm 0.0493}$ & 0.7404$\,^{\pm 0.0432}$ & 0.4423$\,^{\pm 0.0489}$ & 0.8558$\,^{\pm 0.0346}$ \\
\hline
Jais-family-6p7b-chat     & 0.2019$\,^{\pm 0.0396}$ & 0.2692$\,^{\pm 0.0437}$ & 0.0865$\,^{\pm 0.0277}$ & 0.1442$\,^{\pm 0.0346}$ \\
Jais-family-13b-chat      & 0.1635$\,^{\pm 0.0364}$ & 0.1635$\,^{\pm 0.0364}$ & 0.0865$\,^{\pm 0.0277}$ & 0.0865$\,^{\pm 0.0277}$ \\
Fanar-1-9B-Instruct       & 0.4327$\,^{\pm 0.0488}$ & 0.7115$\,^{\pm 0.0446}$ & 0.3750$\,^{\pm 0.0477}$ & 0.8077$\,^{\pm 0.0388}$ \\
SILMA-9B-Instruct-v1.0    & 0.3654$\,^{\pm 0.0474}$ & 0.7308$\,^{\pm 0.0437}$ & 0.2500$\,^{\pm 0.0427}$ & 0.8654$\,^{\pm 0.0336}$ \\
ALLaM-7B-Instruct-preview & 0.2308$\,^{\pm 0.0415}$ & 0.3173$\,^{\pm 0.0459}$ & 0.1923$\,^{\pm 0.0388}$ & 0.1923$\,^{\pm 0.0388}$ \\
AceGPT-v2-8B-Chat         & 0.4808$\,^{\pm 0.0492}$ & 0.7308$\,^{\pm 0.0437}$ & 0.2596$\,^{\pm 0.0432}$ & 0.7788$\,^{\pm 0.0409}$ \\
\hline
Claude-Sonnet-4           & 0.7115$\,^{\pm 0.0446}$ & \textbf{0.8654}$\,^{\pm 0.0336}$ & 0.8365$\,^{\pm 0.0364}$ & \textbf{0.8942}$\,^{\pm 0.0303}$ \\
Claude-3.5-Sonnet         & \textbf{0.7404}$\,^{\pm 0.0432}$ & 0.8077$\,^{\pm 0.0388}$ & \textbf{0.8558}$\,^{\pm 0.0346}$ & 0.9038$\,^{\pm 0.0290}$ \\
GPT-4o                    & 0.7115$\,^{\pm 0.0446}$ & 0.7981$\,^{\pm 0.0396}$ & 0.8462$\,^{\pm 0.0356}$ & 0.8558$\,^{\pm 0.0346}$ \\
GPT-4o-mini               & 0.6442$\,^{\pm 0.0472}$ & 0.7981$\,^{\pm 0.0396}$ & 0.7404$\,^{\pm 0.0432}$ & 0.8173$\,^{\pm 0.0381}$ \\
\hline
\textbf{Average}          & 0.4971                  & 0.6817                  & 0.5010                  & 0.7221 \\
\hline
\end{tabular}
}
\caption{Accuracy ($\uparrow$) and standard error ($\downarrow$) for the connotations of proverbs, their explanations (Jawaher dataset), idioms, and their explanations (Kinayat dataset).}
\label{tab:sentiment-match}
\end{table*}

\begin{table*}[t]
\centering
\small
\begin{tabular}{l | ccc}
\hline
\textbf{Task} & \textbf{Multilingual} & \textbf{Arabic} & \textbf{Closed-Source} \\
& \textbf{Average} & \textbf{Average} & \textbf{Average} \\
\hline
MAPS MCQ & 0.9129 & 0.8739 & 0.9344 \\
MAPS MCQ Context & 0.9560 & 0.9294 & 0.9797 \\
MAPS Completion & 0.7126 & 0.6366 & 0.9141 \\
\hline
\textbf{English Average Accuracy} & 0.8605 & 0.8133 & 0.9428 \\
\hline
Jawaher MCQ (general) & 0.8356 & 0.8123 & 0.9588 \\
Jawaher MCQ (SRL) & 0.8406 & 0.8199 & 0.9335 \\
Jawaher MCQ Negation & 0.7997 & 0.7811 & 0.9369 \\
Jawaher Completion & 0.0527 & 0.0673 & 0.2323 \\
Kinayat MCQ (general) & 0.7097 & 0.6867 & 0.8990 \\
Kinayat MCQ (SRL) & 0.7504 & 0.7436 & 0.8682 \\
Kinayat MCQ Negation & 0.6844 & 0.6174 & 0.8662 \\
Kinayat Pragmatic Use & 0.6081 & 0.5778 & 0.7678 \\
Kinayat MCQ (150 samples) & 0.7356 & 0.7133 & 0.9144 \\
Kinayat MCQ Context (150 samples) & 0.8674 & 0.8456 & 0.9622 \\
\hline
\textbf{Arabic Average Accuracy} & 0.6884 & 0.6665 & 0.8339 \\
\hline
Jawaher Generation BERTScore-F1 & 0.6679 & 0.6770 & 0.6857 \\
Kinayat Generation BERTScore-F1 & 0.6478 & 0.6530 & 0.6625 \\
\hline
\textbf{Arabic Average BERTScore-F1} & 0.6579 & 0.6650 & 0.6741 \\
\hline
Jawaher Generation LLM-Judge & 2.7997 & 2.8480 & 3.6625 \\
Kinayat Generation LLM-Judge & 2.0643 & 2.0683 & 2.6559 \\
\hline
\textbf{Arabic Average LLM-Judge} & 2.4320 & 2.4581 & 3.1592 \\
\hline
\end{tabular}
\caption{Performance results ($\uparrow$) across different tasks and evaluation metrics (Multilingual Average excludes Llama-3.1 70B Instruct model).}
\label{tab:averages}
\end{table*}

\begin{table*}[!ht]
\centering
\resizebox{\textwidth}{!}{
\begin{tabular}{l|cccccccccccccccccccc}
\hline
\textbf{Model} & \textbf{EGY} & \textbf{MSA} & \textbf{UAE} & \textbf{JOR} & \textbf{MAU} & \textbf{PAL} & \textbf{ALG} & \textbf{SYR} & \textbf{IRQ} & \textbf{LEB} & \textbf{SAU} & \textbf{YEM} & \textbf{MOR} & \textbf{QAT} & \textbf{SUD} & \textbf{KUW} & \textbf{OMA} & \textbf{TUN} & \textbf{BAH} & \textbf{LIB} \\
\hline
Llama-3.1-8B-Instruct & 0.778 & 0.900 & 1.000 & 0.700 & 0.400 & 0.455 & 0.800 & 0.700 & 1.000 & 0.444 & 0.800 & 0.700 & 0.800 & 0.900 & 0.400 & 0.700 & 0.900 & 0.700 & 1.000 & 0.900 \\
Llama-3.1-70B-Instruct & 0.778 & 1.000 & 1.000 & 1.000 & 0.900 & 0.909 & 1.000 & 1.000 & 0.900 & 0.778 & 1.000 & 0.900 & 1.000 & 0.900 & 0.900 & 0.800 & 0.900 & 0.900 & 0.778 & 1.000 \\
Gemma-2-9b-it & 0.889 & 0.900 & 1.000 & 0.900 & 0.900 & 0.909 & 1.000 & 1.000 & 1.000 & 0.889 & 1.000 & 1.000 & 0.800 & 0.700 & 0.900 & 0.800 & 0.900 & 0.900 & 0.889 & 0.900 \\
Gemma-2-27b-it & 1.000 & 0.900 & 1.000 & 0.800 & 0.800 & 0.909 & 0.800 & 1.000 & 0.900 & 0.778 & 1.000 & 0.900 & 0.800 & 0.800 & 0.800 & 0.900 & 1.000 & 1.000 & 1.000 & 1.000 \\
Qwen2.5-7B-Instruct & 1.000 & 1.000 & 0.900 & 0.900 & 0.600 & 0.909 & 0.800 & 0.800 & 0.700 & 0.889 & 0.800 & 1.000 & 0.800 & 0.800 & 0.700 & 0.800 & 1.000 & 0.900 & 1.000 & 0.800 \\
Qwen2.5-14B-Instruct & 0.889 & 0.900 & 1.000 & 0.800 & 0.900 & 0.818 & 0.700 & 0.800 & 0.900 & 1.000 & 0.900 & 1.000 & 0.900 & 0.800 & 1.000 & 0.800 & 1.000 & 0.900 & 1.000 & 1.000 \\
Qwen2.5-32B-Instruct & 1.000 & 1.000 & 1.000 & 0.800 & 0.800 & 0.909 & 0.900 & 1.000 & 0.900 & 1.000 & 1.000 & 0.900 & 0.800 & 0.900 & 0.800 & 0.900 & 1.000 & 0.900 & 1.000 & 0.900 \\
Aya-expanse-8b & 0.889 & 1.000 & 1.000 & 0.700 & 0.600 & 0.727 & 0.700 & 0.800 & 0.900 & 0.556 & 0.500 & 0.900 & 1.000 & 0.600 & 0.600 & 0.700 & 1.000 & 0.800 & 1.000 & 0.900 \\
Aya-expanse-32b & 0.889 & 1.000 & 1.000 & 0.900 & 0.600 & 0.818 & 0.800 & 0.900 & 1.000 & 1.000 & 0.800 & 1.000 & 0.800 & 0.800 & 0.900 & 0.900 & 1.000 & 0.900 & 1.000 & 1.000 \\
Mistral-7B-Instruct & 0.889 & 0.800 & 0.900 & 0.700 & 0.300 & 0.636 & 0.500 & 0.600 & 0.700 & 0.556 & 0.400 & 0.300 & 0.800 & 0.500 & 0.400 & 0.700 & 0.700 & 0.500 & 0.667 & 0.700 \\
Jais-family-6p7b-chat & 0.556 & 1.000 & 0.900 & 0.600 & 0.800 & 0.455 & 0.700 & 0.500 & 0.900 & 0.556 & 0.900 & 0.700 & 0.600 & 0.700 & 0.500 & 0.800 & 0.900 & 1.000 & 1.000 & 0.900 \\
Jais-family-13b-chat & 0.889 & 1.000 & 0.900 & 0.600 & 0.500 & 0.636 & 0.900 & 0.700 & 1.000 & 0.889 & 0.500 & 0.800 & 0.800 & 0.800 & 0.400 & 0.700 & 0.800 & 0.700 & 1.000 & 0.700 \\
Fanar-1-9B-Instruct & 0.778 & 0.900 & 1.000 & 0.900 & 0.700 & 1.000 & 0.800 & 0.800 & 0.900 & 1.000 & 1.000 & 0.900 & 0.800 & 0.800 & 0.600 & 0.800 & 1.000 & 1.000 & 1.000 & 0.900 \\
SILMA-9B-Instruct & 0.778 & 0.900 & 1.000 & 0.900 & 0.600 & 0.818 & 1.000 & 0.800 & 1.000 & 0.889 & 0.800 & 1.000 & 0.800 & 1.000 & 0.800 & 0.700 & 0.900 & 1.000 & 1.000 & 1.000 \\
ALLaM-7B-Instruct & 1.000 & 0.900 & 1.000 & 0.900 & 1.000 & 0.545 & 0.800 & 0.900 & 1.000 & 0.667 & 0.900 & 0.900 & 0.800 & 0.900 & 0.700 & 0.900 & 0.800 & 1.000 & 0.889 & 0.800 \\
AceGPT-v2-8B-Chat & 0.778 & 0.700 & 0.900 & 0.800 & 0.400 & 0.545 & 0.800 & 0.800 & 0.800 & 0.778 & 0.600 & 0.800 & 0.900 & 0.900 & 0.500 & 0.800 & 0.900 & 0.700 & 0.778 & 0.900 \\
Claude-Sonnet-4 & 1.000 & 1.000 & 1.000 & 0.900 & 1.000 & 1.000 & 1.000 & 1.000 & 1.000 & 0.889 & 1.000 & 1.000 & 0.900 & 0.900 & 1.000 & 1.000 & 1.000 & 1.000 & 1.000 & 1.000 \\
Claude-3.5-Sonnet & 1.000 & 1.000 & 1.000 & 1.000 & 1.000 & 1.000 & 1.000 & 1.000 & 1.000 & 1.000 & 1.000 & 1.000 & 0.900 & 1.000 & 0.900 & 0.900 & 1.000 & 1.000 & 1.000 & 1.000 \\
Gemini-1.5-flash & 0.889 & 0.900 & 1.000 & 1.000 & 0.800 & 0.909 & 1.000 & 0.900 & 1.000 & 1.000 & 0.900 & 0.900 & 0.900 & 0.900 & 0.800 & 0.900 & 1.000 & 1.000 & 1.000 & 1.000 \\
Gemini-2.5-flash-lite & 1.000 & 1.000 & 1.000 & 1.000 & 0.900 & 1.000 & 0.900 & 1.000 & 0.900 & 1.000 & 1.000 & 1.000 & 0.800 & 0.900 & 0.900 & 1.000 & 0.900 & 1.000 & 1.000 & 1.000 \\
GPT-4o & 1.000 & 0.900 & 1.000 & 1.000 & 1.000 & 1.000 & 1.000 & 1.000 & 1.000 & 1.000 & 1.000 & 1.000 & 0.900 & 0.800 & 1.000 & 1.000 & 1.000 & 1.000 & 1.000 & 1.000 \\
GPT-4o-mini & 0.889 & 1.000 & 1.000 & 0.800 & 0.600 & 0.909 & 0.900 & 0.900 & 1.000 & 1.000 & 1.000 & 0.900 & 1.000 & 0.800 & 0.800 & 1.000 & 1.000 & 0.800 & 1.000 & 1.000 \\
\hline
\textbf{Average} & 0.798 & 0.936 & 0.932 & 0.909 & 0.791 & 0.893 & 0.891 & 0.882 & 0.841 & 0.869 & 0.845 & 0.777 & 0.941 & 0.836 & 0.755 & 0.886 & 0.864 & 0.868 & 0.833 & 0.918 \\
\hline
\end{tabular}
}
\caption{Accuracy (↑) of models across Arabic dialects and countries (with the general prompt).}
\label{tab:arabic-dialect-accuracy}
\end{table*}

\begin{table*}[!ht]
\centering
\resizebox{\textwidth}{!}{
\begin{tabular}{l|cccccccccccccccccccc}
\hline
\textbf{Model} & \textbf{EGY} & \textbf{MSA} & \textbf{UAE} & \textbf{JOR} & \textbf{MAU} & \textbf{PAL} & \textbf{ALG} & \textbf{SYR} & \textbf{IRQ} & \textbf{LEB} & \textbf{SAU} & \textbf{YEM} & \textbf{MOR} & \textbf{QAT} & \textbf{SUD} & \textbf{KUW} & \textbf{OMA} & \textbf{TUN} & \textbf{BAH} & \textbf{LIB} \\
\hline
Llama-3.1-8B-Instruct & 0.444 & 1.000 & 0.800 & 1.000 & 0.700 & 0.455 & 0.900 & 0.700 & 0.700 & 0.778 & 0.900 & 0.800 & 0.900 & 0.800 & 0.600 & 0.600 & 0.600 & 0.800 & 0.667 & 0.700 \\
Llama-3.1-70B-Instruct & 1.000 & 0.900 & 0.900 & 0.800 & 0.700 & 0.818 & 0.800 & 0.900 & 0.600 & 1.000 & 1.000 & 0.700 & 1.000 & 0.600 & 0.600 & 0.900 & 0.800 & 0.800 & 0.667 & 0.900 \\
Gemma-2-9b-it & 0.778 & 1.000 & 0.900 & 1.000 & 0.900 & 0.818 & 1.000 & 0.800 & 0.900 & 0.889 & 0.800 & 0.900 & 1.000 & 0.800 & 0.700 & 0.900 & 0.900 & 0.900 & 0.778 & 0.900 \\
Gemma-2-27b-it & 0.667 & 1.000 & 1.000 & 1.000 & 0.800 & 1.000 & 1.000 & 1.000 & 0.800 & 0.778 & 0.900 & 0.800 & 1.000 & 0.800 & 0.700 & 0.900 & 0.900 & 1.000 & 0.778 & 1.000 \\
Qwen2.5-7B-Instruct & 0.667 & 1.000 & 1.000 & 0.900 & 0.600 & 1.000 & 0.900 & 0.900 & 0.700 & 1.000 & 0.900 & 0.800 & 0.900 & 0.800 & 0.800 & 0.700 & 0.900 & 1.000 & 0.778 & 1.000 \\
Qwen2.5-14B-Instruct & 0.778 & 1.000 & 1.000 & 1.000 & 0.800 & 1.000 & 0.900 & 1.000 & 0.800 & 0.889 & 1.000 & 0.900 & 1.000 & 1.000 & 0.800 & 1.000 & 0.900 & 0.900 & 1.000 & 1.000 \\
Qwen2.5-32B-Instruct & 0.889 & 1.000 & 1.000 & 1.000 & 1.000 & 1.000 & 0.900 & 0.900 & 1.000 & 1.000 & 1.000 & 0.900 & 1.000 & 0.900 & 0.800 & 1.000 & 1.000 & 1.000 & 0.889 & 1.000 \\
Aya-expanse-8b & 0.667 & 0.800 & 0.800 & 0.800 & 0.800 & 0.818 & 0.700 & 0.900 & 0.700 & 0.889 & 0.500 & 0.700 & 0.900 & 0.600 & 0.700 & 0.900 & 0.800 & 0.700 & 0.778 & 0.900 \\
Aya-expanse-32b & 0.889 & 1.000 & 1.000 & 1.000 & 0.800 & 0.909 & 1.000 & 0.900 & 0.900 & 0.889 & 0.900 & 0.900 & 1.000 & 0.900 & 1.000 & 1.000 & 0.900 & 0.900 & 0.889 & 0.900 \\
Mistral-7B-Instruct & 1.000 & 0.500 & 0.800 & 0.500 & 0.600 & 0.909 & 0.700 & 0.500 & 0.600 & 0.556 & 0.500 & 0.300 & 0.700 & 0.400 & 0.300 & 0.800 & 0.700 & 0.600 & 0.444 & 0.500 \\
Jais-family-6p7b-chat & 0.778 & 0.900 & 0.900 & 0.700 & 0.700 & 0.545 & 0.900 & 0.900 & 0.600 & 0.667 & 0.800 & 0.700 & 0.900 & 0.900 & 0.800 & 0.700 & 0.900 & 0.700 & 0.778 & 1.000 \\
Jais-family-13b-chat & 0.556 & 0.800 & 0.900 & 0.900 & 0.700 & 0.909 & 0.900 & 0.900 & 0.800 & 0.778 & 0.800 & 0.600 & 0.900 & 0.800 & 0.700 & 0.800 & 0.700 & 0.600 & 0.889 & 0.700 \\
Fanar-1-9B-Instruct & 0.778 & 1.000 & 0.900 & 1.000 & 0.700 & 1.000 & 0.800 & 1.000 & 0.900 & 0.889 & 0.700 & 0.600 & 0.900 & 0.800 & 0.800 & 1.000 & 0.900 & 0.900 & 0.889 & 0.900 \\
SILMA-9B-Instruct & 0.889 & 1.000 & 0.800 & 1.000 & 0.800 & 0.818 & 1.000 & 0.900 & 1.000 & 1.000 & 1.000 & 0.700 & 0.900 & 0.800 & 0.700 & 0.900 & 0.800 & 0.900 & 0.667 & 0.900 \\
ALLaM-7B-Instruct & 0.889 & 0.900 & 0.900 & 0.900 & 0.800 & 0.818 & 0.700 & 0.900 & 0.800 & 0.778 & 1.000 & 0.800 & 0.900 & 0.900 & 0.600 & 0.900 & 0.700 & 0.900 & 1.000 & 0.900 \\
AceGPT-v2-8B-Chat & 0.778 & 0.700 & 0.900 & 0.800 & 0.400 & 0.909 & 0.800 & 0.800 & 0.700 & 0.778 & 0.300 & 0.800 & 0.800 & 0.800 & 0.700 & 0.900 & 0.800 & 0.900 & 0.667 & 0.900 \\
Claude-Sonnet-4 & 1.000 & 1.000 & 1.000 & 1.000 & 0.900 & 1.000 & 1.000 & 0.900 & 1.000 & 1.000 & 1.000 & 1.000 & 1.000 & 1.000 & 0.900 & 1.000 & 1.000 & 1.000 & 1.000 & 1.000 \\
Claude-3.5-Sonnet & 1.000 & 1.000 & 1.000 & 1.000 & 1.000 & 1.000 & 1.000 & 1.000 & 1.000 & 1.000 & 1.000 & 1.000 & 0.900 & 1.000 & 0.900 & 0.900 & 1.000 & 1.000 & 1.000 & 1.000 \\
Gemini-1.5-flash & 0.778 & 1.000 & 1.000 & 0.900 & 0.900 & 0.909 & 0.900 & 0.900 & 0.900 & 0.889 & 0.900 & 0.800 & 1.000 & 0.900 & 0.800 & 0.900 & 1.000 & 0.900 & 0.667 & 1.000 \\
Gemini-2.5-flash-lite & 0.889 & 1.000 & 1.000 & 0.800 & 0.800 & 1.000 & 0.900 & 1.000 & 0.900 & 1.000 & 1.000 & 1.000 & 0.800 & 0.900 & 0.900 & 1.000 & 0.900 & 1.000 & 1.000 & 1.000 \\
GPT-4o & 1.000 & 0.900 & 1.000 & 1.000 & 1.000 & 1.000 & 1.000 & 1.000 & 1.000 & 1.000 & 1.000 & 1.000 & 0.900 & 0.800 & 1.000 & 1.000 & 1.000 & 1.000 & 1.000 & 1.000 \\
GPT-4o-mini & 0.889 & 1.000 & 1.000 & 0.800 & 0.900 & 0.909 & 0.800 & 0.800 & 1.000 & 0.889 & 0.800 & 0.600 & 1.000 & 0.800 & 0.800 & 0.800 & 0.800 & 0.700 & 1.000 & 1.000 \\
\hline
\textbf{Average} & 0.808 & 0.932 & 0.927 & 0.900 & 0.782 & 0.888 & 0.882 & 0.877 & 0.827 & 0.879 & 0.845 & 0.768 & 0.941 & 0.823 & 0.741 & 0.891 & 0.859 & 0.864 & 0.828 & 0.914 \\
\hline
\end{tabular}
}
\caption{Accuracy (↑) of models across Arabic dialects and countries (with incorrect explanations generated with SRL).}
\label{tab:arabic-dialect-accuracy-srl}
\end{table*}

\begin{table*}[!ht]
\centering
\resizebox{\textwidth}{!}{
\begin{tabular}{l|cccccccccccccccccccc}
\hline
\textbf{Model} & \textbf{EGY} & \textbf{MSA} & \textbf{UAE} & \textbf{JOR} & \textbf{MAU} & \textbf{PAL} & \textbf{ALG} & \textbf{SYR} & \textbf{IRQ} & \textbf{LEB} & \textbf{SAU} & \textbf{YEM} & \textbf{MOR} & \textbf{QAT} & \textbf{SUD} & \textbf{KUW} & \textbf{OMA} & \textbf{TUN} & \textbf{BAH} & \textbf{LIB} \\
\hline
Llama-3.1-8B-Instruct & 0.611 & 0.950 & 0.900 & 0.850 & 0.550 & 0.455 & 0.850 & 0.700 & 0.850 & 0.611 & 0.850 & 0.750 & 0.850 & 0.850 & 0.500 & 0.650 & 0.750 & 0.750 & 0.833 & 0.800 \\
Llama-3.1-70B-Instruct & 0.833 & 0.950 & 1.000 & 0.900 & 0.900 & 0.955 & 1.000 & 0.800 & 0.950 & 0.778 & 1.000 & 0.950 & 0.900 & 0.900 & 0.950 & 0.750 & 0.950 & 0.950 & 0.889 & 0.900 \\
Gemma-2-9b-it & 0.833 & 0.950 & 0.950 & 0.950 & 0.900 & 0.864 & 1.000 & 0.900 & 0.950 & 0.889 & 0.900 & 0.950 & 0.900 & 0.750 & 0.800 & 0.850 & 0.900 & 0.900 & 0.833 & 0.900 \\
Gemma-2-27b-it & 0.833 & 0.950 & 1.000 & 0.900 & 0.800 & 0.955 & 0.900 & 1.000 & 0.850 & 0.778 & 0.950 & 0.850 & 1.000 & 0.800 & 0.750 & 0.900 & 0.950 & 1.000 & 0.889 & 1.000 \\
Qwen2.5-7B-Instruct & 0.833 & 1.000 & 0.950 & 0.900 & 0.600 & 0.955 & 0.850 & 0.850 & 0.700 & 1.000 & 0.900 & 0.800 & 0.850 & 0.800 & 0.800 & 0.700 & 0.900 & 1.000 & 0.778 & 1.000 \\
Qwen2.5-14B-Instruct & 0.833 & 0.950 & 1.000 & 0.900 & 0.850 & 0.909 & 0.800 & 0.900 & 0.850 & 0.944 & 0.950 & 0.950 & 0.950 & 0.900 & 0.900 & 0.900 & 0.950 & 0.900 & 1.000 & 1.000 \\
Qwen2.5-32B-Instruct & 0.944 & 1.000 & 1.000 & 0.900 & 0.900 & 0.955 & 0.900 & 0.950 & 0.950 & 1.000 & 1.000 & 0.900 & 1.000 & 0.900 & 0.800 & 1.000 & 1.000 & 1.000 & 0.944 & 0.950 \\
Aya-expanse-8b & 0.778 & 0.900 & 0.900 & 0.750 & 0.700 & 0.773 & 0.700 & 0.850 & 0.800 & 0.722 & 0.500 & 0.700 & 0.900 & 0.600 & 0.700 & 0.900 & 0.800 & 0.700 & 0.778 & 0.900 \\
Aya-expanse-32b & 0.889 & 1.000 & 1.000 & 0.950 & 0.700 & 0.864 & 0.900 & 0.900 & 0.950 & 0.944 & 0.850 & 0.950 & 0.900 & 0.850 & 0.950 & 0.950 & 0.950 & 0.900 & 0.944 & 0.950 \\
Mistral-7B-Instruct & 1.000 & 0.500 & 0.800 & 0.600 & 0.450 & 0.773 & 0.600 & 0.550 & 0.650 & 0.556 & 0.450 & 0.300 & 0.750 & 0.450 & 0.350 & 0.750 & 0.700 & 0.550 & 0.556 & 0.600 \\
Jais-family-6p7b-chat & 0.667 & 0.950 & 0.900 & 0.650 & 0.750 & 0.500 & 0.800 & 0.700 & 0.750 & 0.611 & 0.850 & 0.700 & 0.750 & 0.800 & 0.650 & 0.750 & 0.900 & 0.850 & 0.889 & 0.950 \\
Jais-family-13b-chat & 0.722 & 0.900 & 0.900 & 0.750 & 0.600 & 0.773 & 0.900 & 0.800 & 0.900 & 0.833 & 0.650 & 0.700 & 0.850 & 0.800 & 0.550 & 0.750 & 0.750 & 0.650 & 0.944 & 0.700 \\
Fanar-1-9B-Instruct & 0.778 & 1.000 & 0.900 & 1.000 & 0.700 & 1.000 & 0.800 & 1.000 & 0.900 & 0.889 & 0.700 & 0.600 & 0.900 & 0.800 & 0.800 & 1.000 & 0.900 & 0.900 & 0.944 & 0.900 \\
SILMA-9B-Instruct & 0.833 & 1.000 & 0.900 & 0.950 & 0.700 & 0.818 & 1.000 & 0.850 & 1.000 & 1.000 & 1.000 & 0.700 & 0.900 & 0.800 & 0.700 & 0.900 & 0.800 & 0.900 & 0.667 & 0.900 \\
ALLaM-7B-Instruct & 0.944 & 0.900 & 0.950 & 0.900 & 0.900 & 0.682 & 0.750 & 0.900 & 0.900 & 0.722 & 0.950 & 0.850 & 0.850 & 0.900 & 0.650 & 0.900 & 0.750 & 0.950 & 0.944 & 0.850 \\
AceGPT-v2-8B-Chat & 0.778 & 0.700 & 0.900 & 0.800 & 0.400 & 0.727 & 0.800 & 0.800 & 0.750 & 0.778 & 0.450 & 0.800 & 0.800 & 0.800 & 0.600 & 0.850 & 0.800 & 0.900 & 0.722 & 0.900 \\
Claude-Sonnet-4 & 1.000 & 1.000 & 1.000 & 0.950 & 0.950 & 1.000 & 1.000 & 0.950 & 1.000 & 1.000 & 1.000 & 1.000 & 1.000 & 1.000 & 0.900 & 1.000 & 1.000 & 1.000 & 1.000 & 1.000 \\
Claude-3.5-Sonnet & 0.944 & 1.000 & 1.000 & 1.000 & 0.950 & 1.000 & 1.000 & 0.950 & 1.000 & 1.000 & 1.000 & 0.950 & 0.950 & 0.950 & 0.900 & 0.950 & 1.000 & 0.950 & 1.000 & 1.000 \\
Gemini-1.5-flash & 0.833 & 0.950 & 1.000 & 0.950 & 0.850 & 0.909 & 0.950 & 0.900 & 0.950 & 0.944 & 0.900 & 0.850 & 0.950 & 0.900 & 0.800 & 0.900 & 1.000 & 0.950 & 0.833 & 1.000 \\
Gemini-2.5-flash-lite & 0.944 & 1.000 & 1.000 & 0.900 & 0.850 & 1.000 & 0.900 & 1.000 & 0.900 & 1.000 & 1.000 & 1.000 & 0.800 & 0.900 & 0.900 & 1.000 & 0.900 & 1.000 & 1.000 & 1.000 \\
GPT-4o & 0.944 & 0.950 & 1.000 & 1.000 & 1.000 & 1.000 & 1.000 & 1.000 & 1.000 & 1.000 & 1.000 & 1.000 & 0.900 & 0.800 & 1.000 & 1.000 & 1.000 & 1.000 & 1.000 & 1.000 \\
GPT-4o-mini & 0.889 & 1.000 & 1.000 & 0.800 & 0.750 & 0.909 & 0.850 & 0.850 & 1.000 & 0.944 & 0.900 & 0.750 & 1.000 & 0.800 & 0.800 & 0.900 & 0.900 & 0.750 & 1.000 & 1.000 \\
\hline
\textbf{Average} & 0.846 & 0.934 & 0.955 & 0.873 & 0.761 & 0.853 & 0.873 & 0.861 & 0.886 & 0.856 & 0.850 & 0.834 & 0.889 & 0.830 & 0.750 & 0.861 & 0.902 & 0.882 & 0.899 & 0.916 \\
\hline
\end{tabular}
}
\caption{Accuracy (↑) of models across Arabic dialects and countries (average of two runs).}
\label{tab:arabic-dialect-accuracy-avg}
\end{table*}

\clearpage 
\section{Ablation}\label{app:ablation}

\begin{figure}[htb]
    \centering
    \includegraphics[width=\columnwidth]{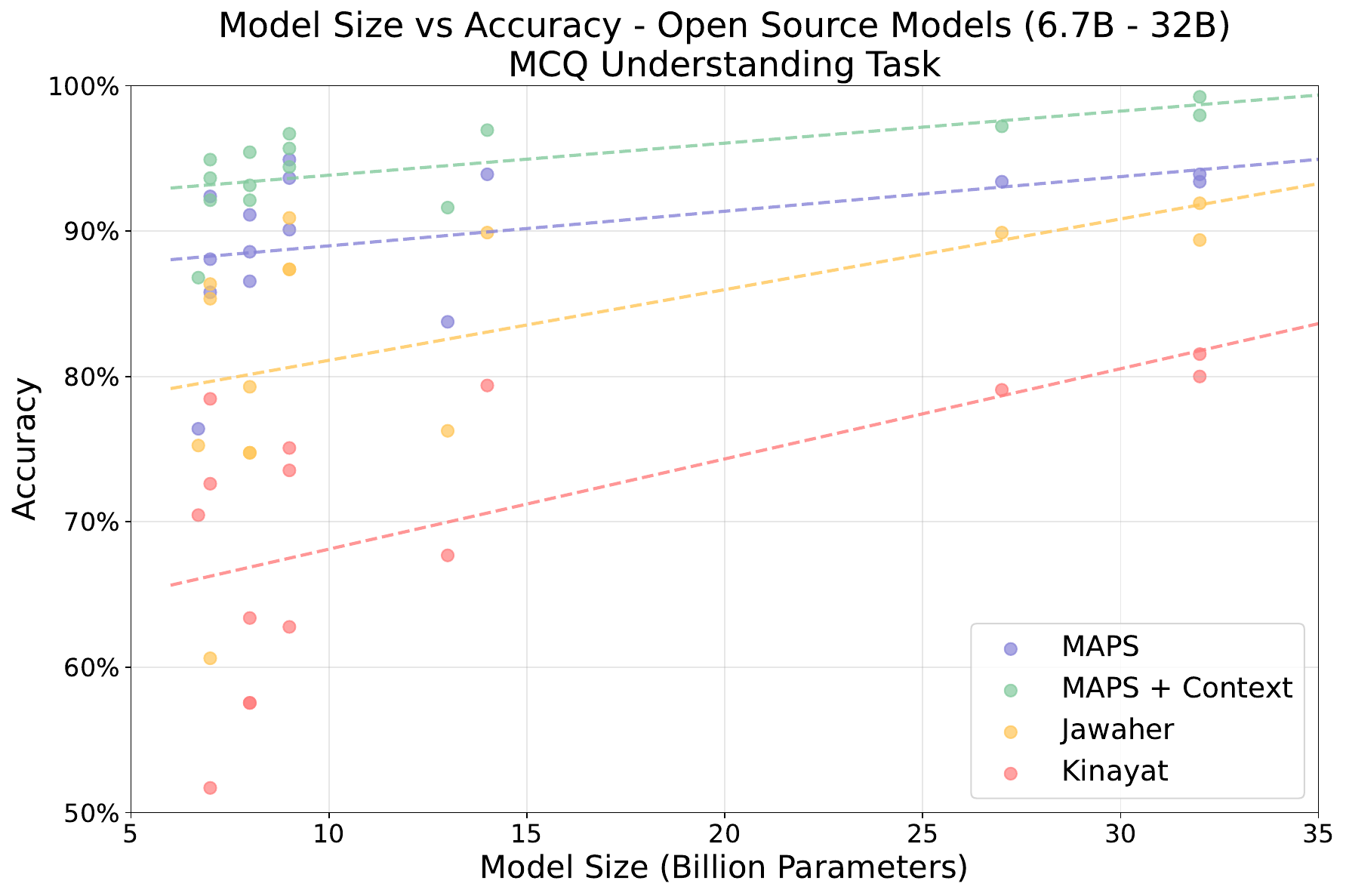}
    \caption{Model size vs accuracy ($\uparrow$) of open-source models (6.7B to 32B) on the MCQ Understanding task.}
    \label{fig:model-size-mcq}
\end{figure}

Table~\ref{tab:mcq_model_size} presents the relationship between model size and performance accuracy for open-source models ranging from 6.7B to 32B parameters on the MCQ Understanding task across different datasets, while Figure~\ref{fig:model-size-mcq} provides the corresponding statistical correlation analysis illustrating the strength of this relationship. Across datasets, correlations were assessed using a significance threshold of $\alpha = 0.05$. For MAPS, the weak positive trend did not reach statistical significance ($p = 0.1053$), whereas adding context (MAPS + Context) resulted in a significant correlation ($p = 0.0086$), suggesting that contextualization enhances the link between model size and performance. Jawaher showed a borderline, yet non-significant, trend ($p = 0.0529$), while Kinayat demonstrated a statistically significant correlation ($p = 0.0178$), indicating that larger models more reliably benefit in this dataset.

\begin{figure}[tb]
    \centering
    \includegraphics[width=\columnwidth]{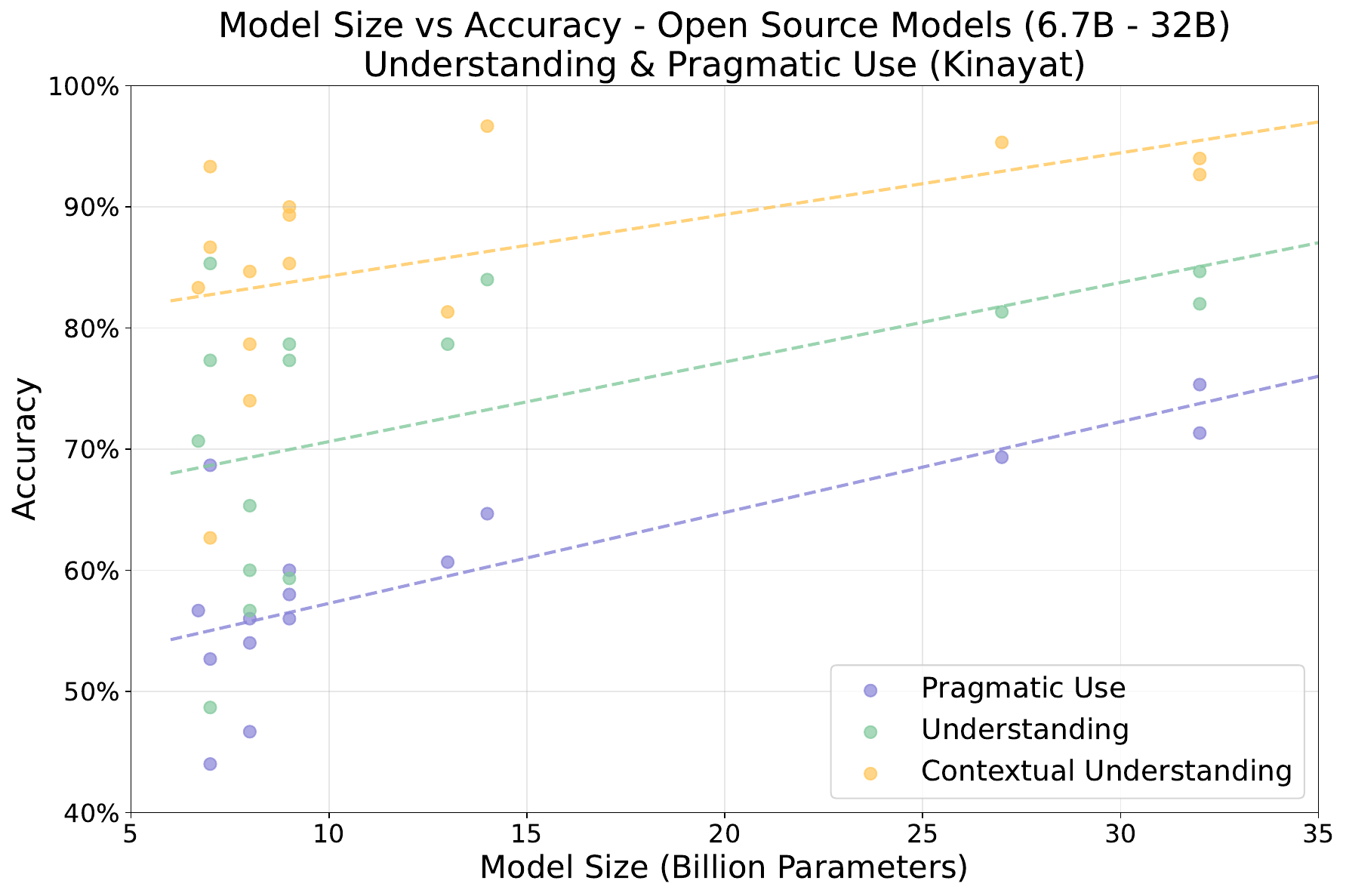}
    \caption{Model size vs accuracy ($\uparrow$) of open-source models (6.7B to 32B) on the Understanding and Pragmatic Use tasks.}
    \label{fig:model-size-pragmatic}
\end{figure}

Similarly, Table~\ref{tab:[prag-use-model-size]} reports the relationship between model size and accuracy for the Pragmatic Use tasks in Kinayat, with Figure~\ref{fig:model-size-pragmatic} visualizing the emerging correlation pattern, further supporting the observation that larger models generally exhibit stronger pragmatic competence in figurative language understanding. Here, model size demonstrates a moderate and statistically significant correlation with pragmatic competence ($R^{2} = 0.600$, $p = 0.0007$), in contrast to a very weak but significant trend observed in MCQ Understanding ($R^{2} = 0.265$, $p = 0.0497$) and Contextual Understanding ($R^{2} = 0.265$, $p = 0.0495$). These findings suggest that larger models not only benefit more clearly from scale in pragmatic reasoning but also gain modestly in interpretive understanding tasks.

\begin{table}[htb]
\centering
\resizebox{\columnwidth}{!}{
\begin{tabular}{lcccl}
\hline
\textbf{Task} & \textbf{R²} & \textbf{Slope} & \textbf{P-value} & \textbf{Interpretation} \\
\hline
MAPS & 0.189 & 0.0024 & 0.1053 & Very weak correlation \\
MAPS + Context & 0.424 & 0.0022 & 0.0086 & Weak correlation \\
Jawaher & 0.259 & 0.0049 & 0.0529 & Very weak correlation  \\
Kinayat & 0.361 & 0.0062 & 0.0178 & Weak correlation \\
\hline
\end{tabular}
}
\caption{Statistical correlation analysis for different datasets: goodness of fit, slope, and significance testing.}
\label{tab:mcq_model_size}
\end{table}

\begin{table}[htb]
\centering
\resizebox{\columnwidth}{!}{
\begin{tabular}{lcccl}
\hline
\textbf{Task} & \textbf{R²} & \textbf{Slope} & \textbf{P-value} & \textbf{Interpretation} \\
\hline
Pragmatic Use & 0.600 & 0.0075 & 0.0007 & Moderate correlation\\
Understanding & 0.265 & 0.0066 & 0.0497 & Very weak correlation \\
Contextual Understanding & 0.265 & 0.0051 & 0.0495 & Very weak correlation \\
\hline
\end{tabular}
}
\caption{Statistical correlation analysis for Kinayat: goodness of git, slope, and significance testing.}
\label{tab:[prag-use-model-size]}
\end{table}

\section{Sample Idioms}\label{app:idioms}
Table~\ref{tab:egyptian-idioms} presents a sample of idioms from the Kinayat dataset along with their corresponding explanations.

\begin{table*}[!htbp]
\centering
\renewcommand{\arraystretch}{0.7}
\resizebox{\textwidth}{!}{
\begin{tabular}{p{4cm} | p{11cm}}
\hline
\textbf{Idiom} & \textbf{Explanation} \\
\hline
{\footnotesize\begin{arabtext}الدُّنْيَا بِتِضْرَبْ وِتِقْلبْ\end{arabtext}} &
{\footnotesize\begin{arabtext}كناية عن كثرة ازدحام الناس في مكان وذهابهم ومجيئهم فيه.\end{arabtext}} \\
{\footnotesize\begin{arabtext}رِجِعْ إيدْ وَرا وإِيدْ قُدَّامْ\end{arabtext}} &
{\footnotesize\begin{arabtext}أي رجع يحرِّك ذراعيه ولا يحمل شيئًا، كناية عن الرجوع بالخيبة.\end{arabtext}} \\
{\footnotesize\begin{arabtext}زَوِّدِ المِبَلَّهْ طِينْ\end{arabtext}} &
{\footnotesize\begin{arabtext}يريدون بالمبلة موضع نقع الكتان، وإذا زادها طينًا فقد زادها فسادًا، كناية عن زيادة الشيء الفاسد فسادًا.\end{arabtext}} \\
{\footnotesize\begin{arabtext}شَم عَلَى ضَهْرْ إيدُهْ\end{arabtext}} &
{\footnotesize\begin{arabtext}يقولون: «ما شمتش على ضهر إيدي» كناية عن عدم معرفة الغيب، أي لم يكن لهذا الخبر رائحة على ظهر يدي فأشمها فمن أين لي معرفته.\end{arabtext}} \\
{\footnotesize\begin{arabtext}عِرِفْهَا وِهيَّ طَايْرَهْ\end{arabtext}} &
{\footnotesize\begin{arabtext}المراد الكلمة يسرع قائلها بها، كناية عن شدة الذكاء، أي عرف ما يقال وأدرك معناه من أوَّل وهلة.\end{arabtext}} \\
{\footnotesize\begin{arabtext}عِينِي عِينِكْ\end{arabtext}} &
{\footnotesize\begin{arabtext}كناية عن الجهر بالشيء بين الناس.\end{arabtext}} \\
{\footnotesize\begin{arabtext}غِرِق فِي شِبْرْ مَيَّهْ\end{arabtext}} &
{\footnotesize\begin{arabtext}كناية عن الارتباك من العجز وقلة الحيلة.\end{arabtext}} \\
{\footnotesize\begin{arabtext}لَتِّ وْعَجْن\end{arabtext}} &
{\footnotesize\begin{arabtext}كناية عن كثرة الكلام وتطويله وإعادة ما قيل، ومعنى اللت والعجن معروف.\end{arabtext}} \\
{\footnotesize\begin{arabtext}لِسَانُه طَوِيلْ\end{arabtext}} &
{\footnotesize\begin{arabtext}كناية عن السفاهة والتطاول بالقذع على الناس وتعود ذلك، والمراد هنا الطول المعنويُّ.\end{arabtext}} \\
{\footnotesize\begin{arabtext}مَا تِنْبَلِّشْ فِي بُقُّهْ فُولهْ\end{arabtext}} &
{\footnotesize\begin{arabtext}البُق الفم، أي لا تبلُّ في فمه فولة، كناية عن عدم كتمان السرِّ والتسرُّع في إفشائه، فالكلمة لا تستقر في فمه كأنها الباقلاءة يسرع في إخراجها قبل أن تبلَّ بريقه.\end{arabtext}} \\
{\footnotesize\begin{arabtext}مَاشِي عَلى قِشْرْ بِيضْ\end{arabtext}} &
{\footnotesize\begin{arabtext}كناية عن المتباطئ الحذر في مشيه، أي كأنه في تباطئه ماشٍ على بيض يخشى أن يكسر قشره بوطئه عليه.\end{arabtext}} \\
{\footnotesize\begin{arabtext}نَشِّفِ الرِّيقْ\end{arabtext}} &
{\footnotesize\begin{arabtext}التنشيف عندهم: التجفيف، كناية عن المضايقة الشديدة بالمماطلة.\end{arabtext}} \\
\hline
\end{tabular}
}
\caption{Examples of Egyptian Arabic idioms from the Kinayat dataset and their explanations.}
\label{tab:egyptian-idioms}
\end{table*}

\clearpage
\section{Error Analysis}\label{app:error}

Tables~\ref{tab:mcq_errors}, \ref{tab:mcq_context_errors}, and \ref{tab:pragmatic_errors} summarize the most frequent error patterns observed across 22 models on the 150-idiom \textsc{Kinayat} subset. Table~\ref{tab:mcq_errors} reports the top eight error types in MCQ idiom understanding before and after the addition of contextual sentences, while Table~\ref{tab:mcq_context_errors} focuses on the most common errors in MCQ contextual understanding. Finally, Table~\ref{tab:pragmatic_errors} contrasts the dominant pragmatic-use errors with those observed in MCQ understanding, highlighting differences in error distributions across evaluation tasks.

\begin{table}[htbp]
\centering
\small
\begin{tabular}{l | cc}
\hline
\textbf{Idiom} & \textbf{MCQ} & \textbf{MCQ w/ Context} \\
\hline
\<خَبَرَ أبْيَضْ> & 10 & 10 \\
\<سَكْرِةْ يَنِّي> & 10 & 10 \\
\<الصَّبَاحْ رَبَاحْ> & 10 & 11 \\
\<إيدْ مِنْ وَرَا وِإيدْ مِنْ قُدَّامْ> & 9 & 3 \\
\<وَلَّعْ> & 9 & 1 \\
\<عَنْدُه الدُّنْيَا بِالْخُلْخَالْ> & 9 & 4 \\
\<إلي حيثْ> & 8 & 7 \\
\<حَطِّ صْبَاعُهْ فِي الشَّق> & 8 & 4 \\
\hline
\end{tabular}
\caption{Most frequent errors in MCQ Understanding on the 150-idiom Kinayat subset, before and after adding contextual sentences (22 models).}
\label{tab:mcq_errors}
\end{table}

\begin{table}[htbp]
\centering
\small
\begin{tabular}{l | c}
\hline
\textbf{Idiom} & \textbf{MCQ w/ Context} \\
\hline
\<الصَّبَاحْ رَبَاحْ> & 11 \\
\<خَبَرَ أبْيَضْ> & 10 \\
\<سَكْرِةْ يَنِّي> & 10 \\
\<دَمُّهْ يُلْطُشْ> & 9 \\
\<عِينِي عِينِكْ> & 8 \\
\<عَمَلِ البَحْرِ طْحِينَه> & 8 \\
\<عَلَى قَفَاهْ> & 7 \\
\<إلي حيثْ> & 7 \\
\hline
\end{tabular}
\caption{Most frequent errors in MCQ Contextual Understanding on the 150-idiom Kinayat subset (22 models).}
\label{tab:mcq_context_errors}
\end{table}

\begin{table}[htbp]
\centering
\small
\begin{tabular}{l | cc}
\hline
\textbf{Idiom} & \textbf{Pragmatic Use} & \textbf{MCQ} \\
\hline
\<اِبْنْ حَرَامْ> & 15 & 0 \\
\<أَشْكرَهْ خَبَرْ> & 15 & 3 \\
\<جَوَازِةْ نَصَارَى> & 15 & 4 \\
\<إتَّاوِب عَ النَّامُوسْ> & 14 & 3 \\
\<يَا مَوْلَايْ كَمَا خَلَقْتِنِي> & 14 & 6 \\
\<وَلَّعْ لهْ قَنْدِيلْ> & 13 & 3 \\
\<جَابْهَا فِي قُبِّتُهْ> & 13 & 4 \\
\<عَمَلِ البَحْرِ طْحِينَه> & 13 & 6 \\
\hline
\end{tabular}
\caption{Comparison of the most frequent pragmatic-use errors against MCQ Understanding errors on the 150-idiom Kinayat subset (22 models).}
\label{tab:pragmatic_errors}
\end{table}

\end{document}